\documentclass{article}
\usepackage{arxiv}
\usepackage[utf8]{inputenc} % allow utf-8 input
\usepackage[T1]{fontenc}    % use 8-bit T1 fonts
\usepackage{hyperref}       % hyperlinks
\usepackage{url}            % simple URL typesetting
\usepackage{booktabs}       % professional-quality tables
\usepackage{amsfonts}       % blackboard math symbols
\usepackage{nicefrac}       % compact symbols for 1/2, etc.
\usepackage{microtype}      % microtypography
\usepackage{lipsum}
\usepackage{graphicx}
\graphicspath{ {./images/} }
\usepackage[numbers]{natbib}
\usepackage{amsmath,amssymb,amsfonts}
\usepackage{textcomp}
\usepackage{enumitem}
\usepackage{bm}
\usepackage{pifont} 
\usepackage{subcaption}
\usepackage{algorithm}%
\usepackage{algorithmicx}%
\usepackage{algpseudocode}%
\usepackage{array}
\usepackage{float}
\usepackage{adjustbox}
\usepackage{xcolor}
\usepackage[preprint]{}
\usepackage{wrapfig}

\newcommand{\mydata}{\texttt{VidSum‑Reason}}

\title{\textbf{Prompts to Summaries: Zero‑Shot Language‑Guided Video Summarization with Large Language and Video Models}}
\author{
  \color{violet}{\textbf{Mario Barbara\quad \quad  \quad \quad\quad
  \href{https://alaamaalouf.github.io/}{\color{violet}Alaa Maalouf}}} \\\\
  \color{magenta}{Department of Computer Science, the University of Haifa} \\
  \color{blue}{\texttt{\{mario.bar98, alaamalouf12\}@gmail.com}} \\
}

\begin{document}
\maketitle

\begin{center}
\color{red}{\texttt{\url{https://github.com/mario998-hash/ZeroShotVideoSummary.git}}}
\end{center}

\begin{abstract}
The explosive growth of video data intensified the need for flexible user‑controllable summarization tools that operate without training data. Existing methods either rely on domain‑specific datasets, limiting generalization, or cannot incorporate user intent expressed in natural language. 
We introduce Prompts-to-Summaries: the first zero-shot, text-queryable video-summarizer that converts off-the-shelf video–language models (VidLMs) captions into user-guided skims via large-language-models (LLMs) judging, without the use of training data, beating unsupervised and matching supervised methods. 
Our pipeline (i) segments video into scenes, (ii) produces scene descriptions with a memory-efficient batch prompting scheme that scales to hours on a single GPU, (iii) scores scene importance with an LLM via tailored prompts, and (iv) propagates scores to frames using new consistency (temporal coherence) and uniqueness (novelty) metrics for fine-grained frame importance.  
On SumMe and TVSum, our approach surpasses all prior data-hungry unsupervised methods and performs competitively on the Query‑Focused Video Summarization benchmark, where the competing methods require supervised frame-level importance.
We release \mydata{}, a query-driven dataset featuring long-tailed concepts and multi-step reasoning, where our framework serves as the first challenging baseline.
Overall, we demonstrate that pretrained multi-modal models, when orchestrated with principled prompting and score propagation, provide a powerful foundation for universal, text‑queryable video summarization.
\end{abstract}

\keywords{Large language models, text guided, training free, unsupervised learning, video language models, video summarization, zero-shot.}

\begin{figure*}[h]
    \centering
    \includegraphics[width=0.9\linewidth]{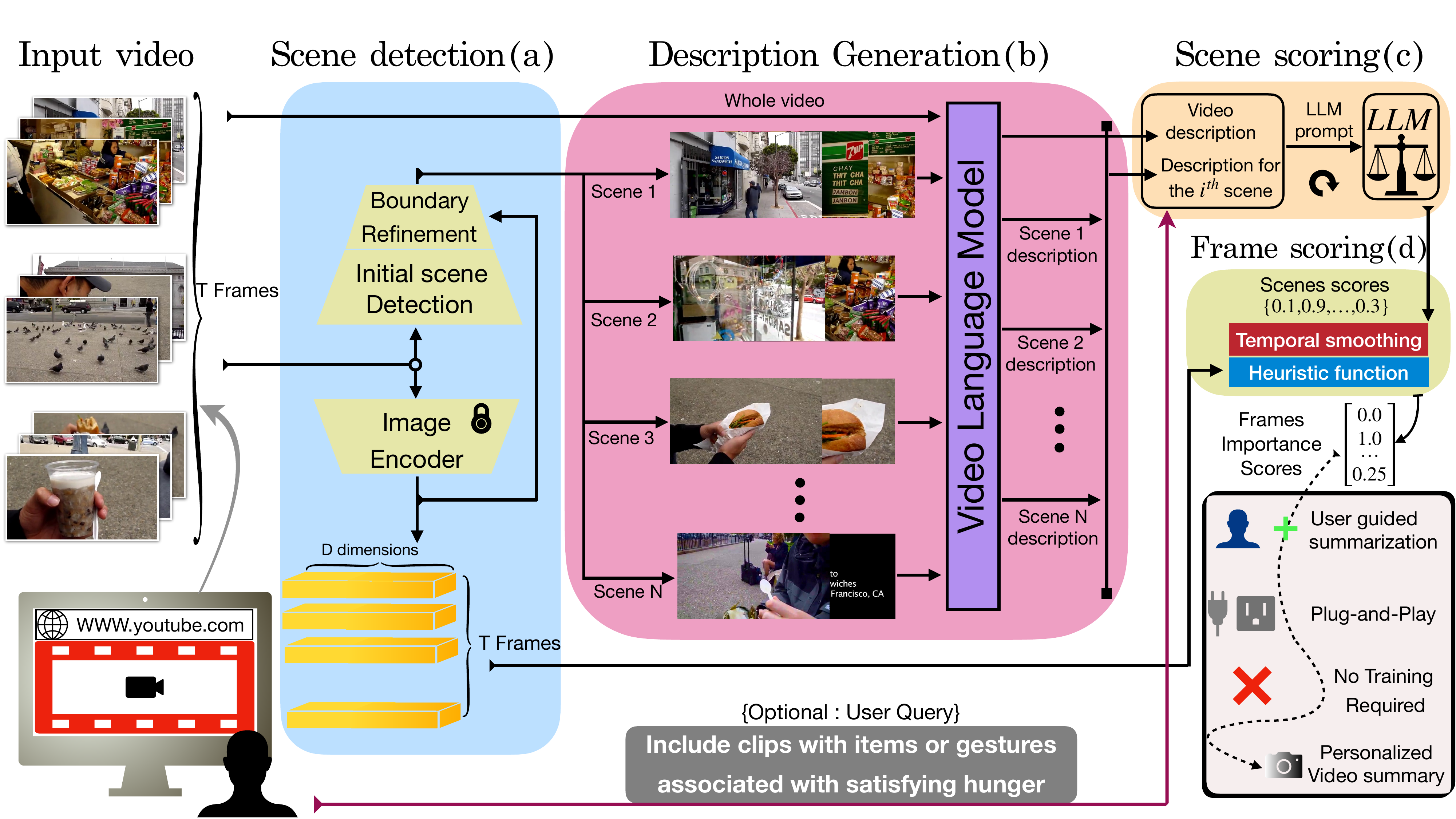}
    \caption{\textbf{Prompts to summaries overview.} Our zero-shot pipeline turns any video + text query into a tailored highlight reel. We (a) detect scenes and snap boundaries with visual embeddings, (b) caption each scene via a video-language model, and (c) let an LLM rank captions against the query alongside its importance as part of the whole video. (d) Smoothed relevance scores weight every frame, yielding a query-aligned summary. No extra training, ready for personal highlights,  generalizes across domains and handles long-tailed, reasoning-heavy queries.}
    \label{fig:pipeline_teaser}
\end{figure*}

\section{Introduction}\label{sec:introduction}

The exponential rise in video content has introduced major challenges in how videos are managed, processed, and consumed~\cite{ma2002user,money2008video,truong2007video}. Driven by the ubiquity of smartphones and the explosion of social media, videos are being produced at an unprecedented scale~\cite{apostolidis2021video}. This growth underscores the urgent need for solutions to information overload, escalating storage demands, and the increasing difficulty of extracting insights from raw footage.

\paragraph{Video summarization.} In response to these challenges, \textit{video summarization} has become a key solution in computer vision research~\cite{rochan2018video,zhang2016video}, aiming to condense lengthy videos into concise, informative representations; allowing users to quickly access essential content without viewing entire sequences~\cite{hussain2021comprehensive,otani2017video}. Thus, enabling broad applications—from surveillance review~\cite{ antani2002survey,zhang2016context}, to educational retrieval and entertainment highlights~\cite{mahapatra2018videoken,alrumiah2022educational}—while reducing storage, bandwidth, and improving indexing~\cite{almeida2012vison,muhammad2019deepres}.

\paragraph{Limitations of exiting video summarization methods.} Despite numerous research efforts~\cite{vo2025integrate,abbasi2023adopting,apostolidis2020ac,he2023align,jiang2022joint,narasimhan2021clip,ghauri2021supervised,rochan2018video,yuan2019cycle}, 
current video summarization approaches face two major drawbacks. First, most solutions—whether supervised~\cite{he2023align,jiang2022joint,narasimhan2021clip,ghauri2021supervised} or unsupervised~\cite{vo2025integrate,abbasi2023adopting,apostolidis2020ac,rochan2018video,yuan2019cycle}—depend heavily on the training data, limiting their ability to generalize beyond the distributions they were trained on and thus reducing their real-world applicability.
Second, they are constrained to the generic, pre-defined summarization task and cannot accommodate user-specified text queries to guide the summarization process toward specific content or themes. 
Although there was an effort to generate query-focused video summarization employing a supervised approach~\cite{narasimhan2021clip,sharghi2016query,gong2014diverse}, the suggested methods rely on dense query-focused user annotations for every frame for training, and they don't provide support for long-tailed queries, or queries requiring reasoning beyond e.g., simple object detection.

To be as broadly applicable as possible across diverse video contexts and user needs, summarization algorithms should fulfill two key requirements: 
(i) they must operate effectively in a zero-shot setting, i.e., without retraining for each new video domain/type, thereby providing a more flexible and generalizable solution; and (ii) they must be \emph{text-queryable}, giving users the ability to influence the summarization process via natural language prompts, such as ``highlight violent scenes'', ``remove inappropriate frames,'', and ``include expressive emotional moments''.

\subsection{Problem Statement}
\newcommand{\ALG}{\textsc{\texttt{ALG}}}
Motivated by the previous discussion, we now define the problem as follows. 
Given a sequence of video frames  $F = \{f_t\}_{t=1}^{T},$ the goal of the \textit{traditional} video summarization task is to develop an algorithm, denoted as \ALG{}, which receives $F$ as input and produces a video skim composed of selected video segments $ S = \{s_n\}_{n=1}^{N},$ ordered chronologically to create a shorter version of the original video. Each segment $s_n$ is a contiguous set of consecutive frames, that is, $s_n = \{f_{t_n}, f_{t_n+1}, \ldots, f_{t_n+\Delta_n}\},$ for some starting frame \( t_n \) and segment length \( \Delta_n \). This form aims to maintain temporal continuity, motion, and audio, thereby offering a richer and more natural viewing experience.

In our formulation, we introduce additional complexity and broaden the applicability of the task: the algorithm \ALG{} \textbf{can also accept an optional user-provided text query $T$ to guide the summarization process}. In this case, the selected segments $S = \{s_n\}_{n=1}^{N}$ are chosen according to the guidance provided by $T$, for example, by focusing on specific parts of the video or ignoring less desired content. Additionally, \ALG{} \textbf{must operate without relying on any training data}.

\newcommand{\vlm}{VideoLM}
\subsection{Main Idea}

Elsewhere, video language models (\vlm{}s) have emerged as a promising tool, enabling language-guided question answering on videos. These models possess some of the desired traits needed to achieve an open-set, text-queryable method for video summarization, as they are already trained on a huge amount of internet-scale data. This extensive training allows them to be highly adaptable, making them not confined to a specific data distribution.

Simply put, while there are no \vlm{}s specialized for video summarization, \vlm{}s can answer textual queries about videos by generating text descriptions (e.g., "What do you see in this scene?"). These text outputs can then be processed by large language models (LLMs), which, using carefully crafted prompts (e.g., "Given this scene description $<$Here$>$ and the full video summary $<$Here$>$, rate the importance of this scene from 1 to 10"), can rank or evaluate the scenes. 
 This flexibility enables this approach to perform summarization tasks across various domains without being limited to one specific application. In addition one can add guidance by text to the prompt, e.g. "Given this scene description $<$Here$>$, the full video description $<$Here$>$, and the user query: $<$Prioritize scenes with negative emotions$>$, rate the importance of this scene from 1 to 10".

\paragraph{Challenges.} Notably, it's not as easy as described as LLM and \vlm{}s where not designed for this purpose. For instance, it is still unclear how to: (i) formulate effective queries for ranking scenes or generating informative text descriptors; (ii) segment videos into semantically meaningful and coherent scenes; (iii) ensure that each scene is both unique and complete; 
(iv) preserving the temporal context in long videos, especially since video-language models are unable to process the entire video at once due to input length limitations.
(v) propagate scene-level importance scores to individual frames—especially since certain frames may better capture the essence of a scene than others; and, most critically, (vi) develop an integrated framework that combines all these stages while leveraging the capabilities of LLMs and \vlm{}s in a principled and effective manner. Thus, these challenges and gaps in current knowledge require a careful and critical examination of how to effectively apply \vlm{}s and LLMs toward a zero-shot text-queryable video summarization.

\subsection{Our Contributions}

To this end, we investigate in depth the usage of large-scale pre-trained video-language models with large language models for zero-shot and data free, text-queryable video summarization. 
Specifically, we explore how \vlm{}s and LLMs, can work together as team to semantically align text queries with video content and assess the relevance of individual scenes within the context of the entire video based on textual information; LLMs can be utilized to assign relevance scores to each video segment based on the textual descriptions generated by the \vlm{}s and user-provided queries, as demonstrated in Figure~\ref{fig:pipeline_teaser}. 
Finally, we delve into methods to propagate scene scoring to frame scores. 
With our approach, we provide the following contributions:

\begin{enumerate}

    \item \textbf{A novel framework. }We provide the first zero-shot text guided video summarization framework that enables generating video summaries conditioned on natural language queries without the need for training data, annotations, or fine-tuning. %Our framework (see Figure~\ref{fig:pipeline_teaser}) can be split into the following stages: (i) Scene detection (ii) Scene description generation (iii) Scene importance within the video, (iv) Frame weight within the scene (v) Frame importance withing the video by aggregating frame weight and scene importance. 

     \item\textbf{State-of-the-art (SOTA) results:} Our approach achieves SOTA performance among all unsupervised methods and performs comparably to some leading supervised approaches, both of which require domain-related training data. 
    Specifically, our method yields the highest F1-scores on well-established standard video summarization datasets of SumMe and TVSum, among all unsupervised approaches, and comparable F1 scores to supervised (per-frame) annotations hungry methods on the query-focused video summarization (QFVS) dataset. All of that without relying on any training data. 
    
    \item\textbf{Experimental study.} To determine the optimal design of each component and assess its contribution to our pipeline, we conducted a comprehensive study by systematically omitting and modifying key elements. This analysis quantifies the individual design and impact of each part on the overall performance and reveals insightful findings about the behavior of \vlm{}s and LLMs in the context of video summarization. \label{cont:2}

     \item \textbf{\mydata{}: A new public dataset for text-guided video summarization requiring reasoning.} Since the QFVS dataset does not capture long-tailed concepts or reasoning-based queries, we curated a diverse dataset by selecting videos from the web and annotating each frame according to goal-driven, text-based queries. This allows for evaluating methods beyond standard benchmark datasets. To enable more nuanced analysis, the queries are structured across multiple levels of reasoning, offering a richer testbed for current and future approaches. We test our method on this dataset, achieving high F1 scores. 
\end{enumerate}

\noindent\textbf{Novelty of our work.} To enable our \emph{training-free}, \emph{text-guided} video summarization framework, we introduce four main novelties: (i) formulating video summarization as LLM-based importance scoring over structured caption units, (ii) scaling video description generation of \vlm{} to long videos via memory-aware batched description generation, (iii) designing a reproducible, rubric-driven prompting protocol for stable importance judgments, and (iv) proposing lightweight score propagation to derive frame-level saliency---all in a fully zero-shot, domain-agnostic setting (see Appendix~Section~\ref{sec:novelty} for more details).

\section{Related work}

\paragraph{Standard video summarization.} In recent years, a wide range of efforts have focused on generating video skims using deep learning techniques~\cite{he2023align,narasimhan2021clip}. These approaches can be broadly categorized on the basis of their training paradigms into the following groups. \textit{Supervised methods} rely on human-annotated summaries to train models that predict frame- or segment-level importance scores. These approaches often use deep neural networks such as recurrent models or reinforcement learning frameworks to capture temporal dependencies and semantic relevance ~\cite{feng2018extractive,ghauri2021supervised,apostolidis2021combining,he2023align,narasimhan2021clip}.
While they achieve strong performance, their reliance on large-scale annotated datasets limits their scalability and generalization across domains. \textit{Weakly supervised methods} attempt to reduce this dependency by learning from coarser supervision, such as video-level tags, summary-level annotations, or user engagement signals~\cite{yuan2019cycle,rochan2018video}.
These methods offer a compromise between label efficiency and learning effectiveness.  \textit{Unsupervised approaches.} In contrast, unsupervised approaches aim to generate summaries without any manual annotations. They typically utilize visual features, motion analysis, or statistical cues like diversity and representativeness. Techniques such as clustering, graph modeling, or temporal segmentation (e.g., KTS~\cite{potapov2014category}) are commonly employed to identify informative segments ~\cite{apostolidis2020ac},~\cite{abbasi2023adopting},~\cite{vo2025integrate}.
Although unsupervised methods are more scalable, they may struggle to understand and relate temporal relationships between scenes and full videos, high-level semantics, or align with user-specific goals. Notably, they still require training data as they leverage a powerful self-supervised learning method  adopted for training language models ~\cite{apostolidis2020ac} and incorporating temporal video segmentation with the summarization process ~\cite{vo2025integrate}.

Recent zero- and few-shot approaches have explored video summarization with minimal supervision. Contrastive Zero-Shot~\cite{pang2024unleashing} ranks frames or segments based on pre-trained video representations learned through a contrastive objective, without relying on labeled summaries. Few-shot methods, such as STeMI~\cite{li2025spatial}, rely on a limited set of annotated examples to learn segment importance. Both approaches, however, do not support text-queryable summarization and cannot directly generate summaries conditioned on arbitrary user queries.

\paragraph{Query-focused video summarization.}\cite{sharghi2016query} introduced the Query-Focused Video Summarization (QFVS) dataset, based on UT Egocentric videos~\cite{lee2012discovering}, which includes user-defined summaries for a fixed set of concepts. They later proposed a memory network~\cite{sharghi2017query} that uses a recurrent attention mechanism to attend to frames and shots based on a user query. However, this approach suffers from limited parallelism and struggles with long-range dependencies—issues our \vlm{}-LLM based model addresses. Additionally, their method is restricted to predefined, keyword-based queries from the QFVS dataset.
More recently, CLIP-It~\cite{narasimhan2021clip} enabled freeform query input by training on dense video descriptions using language encoders, however it relies on clip, thus limiting its reasoning capabilities to explicitly defined objects rather than long tailed concepts and higher reasoning tasks.  
Other works~\cite{kanehira2018aware,wei2018video} also condition summarization on keyword-based queries, however,~\cite{kanehira2018aware} requires detailed annotations and~\cite{kanehira2018aware} performance found to be not robust, compared to modern techniques. 

\paragraph{Limited Supervision and Temporal Reasoning.} Several recent methods tackle video–text alignment under limited supervision. ~\cite{liu2022few} leverages memory modules to improve semantic generalization with scarce annotations, illustrating one approach to limited supervision in multi‑modal settings. ~\cite{liu2023hypotheses} constructs a hierarchical segment tree from a single labeled frame and introduces self‑supervised training strategies to score candidate segments under extreme label scarcity. While both methods focus on localizing query‑relevant temporal segments, our approach differs by leveraging pretrained ~\vlm{} and LLM prompting to score semantic importance without any training data. ~\cite{liu2023conditional} highlights another recent direction in video–text reasoning that could potentially be explored for future fine‑grained scoring in summarization. In contrast, our framework emphasizes zero‑shot, text‑guided summarization by combining scene descriptions and LLM‑driven scoring to produce semantically relevant skims across diverse domains.

%Oftentimes users browsing videos on YouTube are looking
%for something specific so a generic summary might not suffice. 
%In this case, there should be an option
%to customize the generated summary using a natural language query. 
%Sharghi et al.~\cite{sharghi2016query} introduced
%the Query-Focused Video Summarization (QFVS) dataset for UT Egocentric ~\cite{lee2012discovering} videos containing user-defined video summaries for a set of pre-defined concepts. Later,~\cite{sharghi2017query} proposed a memory
%network to attend over different video frames and shots with the user query as input. However, this recurrent attention mechanism precludes parallelization and limits modeling long-range dependencies, which is overcome by our Transformer architecture. Moreover, their method only works with the
%pre-defined set of keyword based queries in QFVS dataset. 
%CLIP-It~\cite{narasimhan2021clip} introduced a model that encodes language input and trains it on dense video descriptions, allowing users to define
%freeform queries at test time. Other works ~\cite{kanehira2018aware,wei2018video} similarly condition the summary generation on keyword-based queries but have not released their data.

\section{Method}
We now define our framework, which is designed to effectively capture and represent the essential content of videos without relying on annotated training data. 

\paragraph{Introduction to the framework.} Our proposed pipeline (see Figure~\ref{fig:pipeline_teaser}), composed of the following key components, each of which must be carefully evaluated to determine the optimal design. %The pipeline consists of the following stages:

\begin{enumerate}
    \item \textbf{Scene detection:} The video is segmented into temporally coherent and semantically meaningful scenes (Figure~\ref{fig:pipeline_teaser}a).
    
    \item \textbf{Scene description:} For each detected scene, a textual descriptor is generated to summarize its salient content as part of the video in natural language (Figure~\ref{fig:pipeline_teaser}b).
    
    \item \textbf{Scene-level importance estimation:} Using the generated textual descriptors, an importance score is assigned to each scene, incorporating both semantic relevance and temporal context within the video (Figure~\ref{fig:pipeline_teaser}c).
    
    \item \textbf{Frame-level weight estimation within the scene} Within each scene, the representativeness of individual frames is evaluated to determine their relative importance to the scene (Figure~\ref{fig:pipeline_teaser}d).
    
    \item \textbf{Fusion of scene and frame importance:} For each frame in the full video, the scene-level (Step 3: scene importance within the video) and frame-level (Step 4: representativeness within the scene) scores are combined/fused to produce a final frame (importance) score (Figure~\ref{fig:pipeline_teaser}d).% for each frame in the full video.
\end{enumerate}

%This structured approach enables a principled, modular evaluation of the zero-shot video summarization process, leveraging both temporal structure and semantic alignment with language models.

\subsection{Scene Detection}\label{sec:scene_detection}
To segment the input video into coherent scenes, we apply a two stage process: initial scene detection with dynamic thresholding and scene boundary refinement. 

\subsubsection{Initial scene detection with dynamic thresholding}\label{sec:init_scene_detect}

\paragraph{Goal:} Segment the video into non-overlapping, chronologically ordered scenes, where each scene captures a coherent event action.
%\paragraph{Goal:} Segmenting the video to non-overlapping chronologically ordered scenes, each representing a single event or action.

%\paragraph{Approaches to investigate:} In this phase we utilize a frame-level content detection algorithm to compute the mean intensity difference between consecutive frames. Instead of using a fixed threshold, we dynamically select the threshold values by iterating over a predefined range. 
%For each candidate threshold, we compute the number of detected scenes the select the threshold that maximize the rate of change (slope) in the number of scenes with respect to the threshold values (add equation); See Figure~\ref{} for example.  \alaa{explain why this is helpfull to use the slope}
%This approach, allows our algorithm to adapt to various video content, ensuring robust initial video segmentation.

%\paragraph{Approach:} In this phase, we employ a frame-level content detection algorithm based on the mean intensity difference (of raw pixels) between consecutive frames. Namely, we leverage the \texttt{contentDetector} function from PySceneDetect~\cite{castellano2024pyscenedetect}. 
%Rather than relying on a fixed difference threshold for splitting the scenes, we dynamically determine the optimal threshold by evaluating a predefined range of candidate values.
\paragraph{Approach:} For each consecutive frame pair $(F_t,F_{t+1})$, we 
use PySceneDetect’s \texttt{contentDetector}~\cite{castellano2024pyscenedetect} to compute the mean absolute pixel–intensity difference \(D_t\) as:
%\begin{equation}
%\label{eq:pySceneDetect}
$D_t = \frac{1}{H \times W} \sum_{i=1}^{H} \sum_{j=1}^{W} \left| F_t(i, j) - F_{t+1}(i, j) \right|$,
%\end{equation}
where $H$ and $W$ are the height and width of the frame, respectively, and $F_t(h, w)$ denotes the value at pixel location $(h, w)$ in frame $t$. A cut is declared when \(D_t \ge \tau\), otherwise $F_{t+1}$ is merged with the current shot.  Then, scenes shorter than 2\,s are merged to avoid over-fragmentation. Finally, scanning the entire video thus produces \(N(\tau)\) scenes (see Figure ~\ref{fig:scene_detection_flow_diagram} in the appendix).

Rather than relying on a fixed threshold to detect the boundaries of the scene, we dynamically determine the optimal threshold $\tau^*$ by evaluating a predefined range of candidate values. 
For each candidate threshold $\tau_i \in \{\tau_{\min},\tau_{min+\Delta\tau} , \dots, \tau_{\max}\}$, we compute the number of detected scenes $N(\tau_i)$ and select the threshold that maximizes the rate of change in the number of scenes with respect to $\tau$:
$\tau^* = \arg\max_{\tau_i} -\frac{N(\tau_i + \Delta \tau) - N(\tau_i)}{\Delta \tau}$, When applying this method, we observe an initial increase at low threshold values, reaching a maximum, and then a sharp decrease with the number of detected scenes. We select the optimal threshold which yields the maximum negative slope after this peak, as shown in Figure~\ref{fig:scene_detection_flow_diagram} in the appendix, capturing the point with the most significant reduction in scene count.  This slope captures the moment when noise-driven fragments disappear yet true boundaries remain (see Figure~\ref{fig:scene_bearpark_pipeline_grid} in the appendix), yielding a segmentation that balances  between over- and under-segmentation while ensuring temporal coherence in the detected scenes (Figure~\ref{fig:general_pipeline_example}(a)).

\subsubsection{Scene boundary refinement}\label{sec:scene_boundary_refinement}

\paragraph{Goal:} To improve coherence and semantic consistency, we refine the initial scene segmentation by merging excessively short scenes (not containing sufficient information), often caused by noise or minor visual changes, with their most similar neighbors, reducing unnecessary fragmentation.

%Merging temporally short (detected) scenes not containing sufficient information. 

\paragraph{Approach:} 
A scene is defined as \textit{short} if it contains fewer than $M = 150$ frames (approximately 5 seconds, we observe that defining short segments as those under 5 seconds is not sufficient to fully prevent over-fragmentation). 

The previous stage marked scene boundaries by detecting sharp intensity changes. To avoid spurious cuts, we now refine them: any short segment whose visual embedding closely matches its neighbors is merged back, ensuring temporally coherent and visually consistent boundaries. Given a short scene $S_i$, we compute its mean embedding: $
\mathbf{e}_i = \frac{1}{|S_i|} \sum_{\mathbf{f} \in S_i} \mathbf{f}
$, 
where $\mathbf{f}$ denotes the frame-level embedding (e.g., based on DINO~\cite{caron2021emerging} or CLIP~\cite{radford2021learning}). We then calculate cosine similarity between $\mathbf{e}_i$ and the mean embeddings of its adjacent scenes, $\mathbf{e}_{i-1}$ and $\mathbf{e}_{i+1}$:
$\text{sim}_{\text{prev}} = \frac{\mathbf{e}_i \cdot \mathbf{e}_{i-1}}{\|\mathbf{e}_i\| \, \|\mathbf{e}_{i-1}\|}, \; \text{sim}_{\text{next}} = \frac{\mathbf{e}_i \cdot \mathbf{e}_{i+1}}{\|\mathbf{e}_i\| \, \|\mathbf{e}_{i+1}\|}$

The scene $S_i$ is merged with the neighboring scene yielding the higher similarity; $\max(\text{sim}_{\text{prev}}, \text{sim}_{\text{next}})$. 
This refinement reduces over-segmentation and yields more temporally coherent scenes that better capture complete events or actions, as shown in Figure~\ref{fig:general_pipeline_example}b.

%\begin{equation}
%\text{merge}(S_i) = 
%\begin{cases}
%S_{i-1}, & \text{if } \text{sim}_{\text{prev}} > \text{sim}_{\text{next}} \\
%S_{i+1}, & \text{otherwise}.
%\end{cases}
%\end{equation}

\subsection{Scene Description Generation}\label{sec:desc}
\paragraph{Goal:} Generating textual description summarizing the scene's content and the overall video.

\paragraph{Approach:} 
After establishing scene boundaries (Figure~\ref{fig:general_pipeline_example}b), we generate textual descriptions for each detected scene and the full video using a pre-trained video-language model. The prompt template used for the \vlm{} is defined as:\\ 
\label{Vlm prompt}
$P_{\;\vlm{}}$: \texttt{"Describe this video in detail"}\\
which produces detailed meaningful descriptions for the corresponding frame sets.

\paragraph{Handling long videos and capturing larger context.} 
To address the memory constraints of \vlm{}s when processing long frame sequences, we adopt a batch-level processing strategy: sampled frames are divided into manageable batches, a description is generated for each batch, and the resulting texts are concatenated to form the full description of a scene or an entire video. 
To ensure smooth transitions between batches, we enforce consistent phrasing, e.g., modifying intermediate batch outputs to begin with “The video continues” and end with “The scene concludes,” except for the first and last batches—thereby preserving temporal and semantic coherence.
Furthermore, to provide the \vlm{} with sufficiently broad temporal context within each batch, we uniformly sample frames at a rate of 1 frame per second, selecting the middle frame of each second to avoid transition or black frames that may occur near shot boundaries. 
This approach allows each batch to represent an extended yet coherent temporal window, effectively capturing high-level semantics while maintaining computational feasibility.

This procedure is consistently applied to generate both scene-level descriptions and a holistic description for the full video.

\subsection{Scene level scoring}
\label{sec:scene_scoring}
\paragraph{Goal:} For every scene, we wish to generate a score which represent the importance of this scene as part of the whole video, based on the text descriptions as shown in Figure~\ref{fig:general_pipeline_example}(c). 

\paragraph{Approach:} 
We utilize the capabilities of LLMs to estimate scene-level importance scores based on the textual descriptions of both the individual scenes and the overall video, which were generated earlier. Specifically, we employ LLMs as judges to assess the contribution of each scene via a single carefully designed prompt template (shown in Section ~\ref{sec:appendix_LLM_prompt} in the) appendix; For each scene, we construct an LLM query by plugging the  scene description, and the overall video description, into the prompt template, as detailed in Algorithm ~\ref{alg:scene-level-scoring} in the appendix. %by reply on LLMs ability for reasoning, we assess the importance of each scene within the video, 

\paragraph{The carefully designed prompt:} The prompt in section~\ref{sec:appendix_LLM_prompt} in the appendix was meticulously engineered to elicit consistent, fine‑grained importance scores from LLMs. First, we cast the LLM in the explicit role of an “objective evaluator”, which anchors the model’s behavior and reduces response variance. 
Next, we supply both the global (video‑level) and local (scene‑level) descriptions, ensuring the model reasons in context rather than in isolation. 
A five‑band rubric—complete with numeric intervals, textual labels, and concrete definitions—calibrates the 1‑to‑100 scale, so identical semantic cues map to comparable scores across videos. 
Crucially, the prompt enforces a sparse‑high prior: it instructs the model to reserve scores above 80 for scenes that directly advance the core narrative, while encouraging low scores for ancillary content. This bias guards against score inflation and aligns the LLM's output with human summarization objectives, where only a handful of pivotal moments should survive into a concise synopsis.
Finally, explicit reminders to “prioritize” essential scenes and “omit” secondary ones further steer the model toward selective, summary‑oriented judgments—yielding reproducible importance estimates that integrate seamlessly with our downstream ranking pipeline.

\paragraph{User guided scoring}
\paragraph{Goal:} Guiding the summary process based on textual user query.

%\paragraph{Approach:}
%To incorporate user guidance into the video summarization process, we use the same prompt used during the scene scoring stage of our pipeline, shown in the Appendix(section~\ref{sec:appendix_implementation_details}), with minimal changes. The key addition involves incorporating explicit instructions to assess the importance of a scene relative to a user-provided query, as shown in (Appendix section~\ref{sec:appendix_implementation_details}). When a user query is provided, the prompt includes generic guidance for the language model to evaluate how well the scene aligns with the user's content preference. 

%Specifically, scenes that represents a key events or highly align to the query are assigned a higher importance score, e.g., scenes demonstrating emotions. while unrelated segments or strongly contradictory scenes are deprioritized, meaning such scenes are assigned low importance scores, e.g. scenes demonstrating violence.  

%This lightweight adjustment enables zero-shot, query-focused summarization through prompt design alone, without altering the underlying model or scoring logic.

\paragraph{Approach.}
To enable user-guided video summarization, we adapt the same prompt used in the standard scene scoring stage (see Section ~\ref{sec:appendix_LLM_prompt} in the appendix) with one key modification: we add instructions that explicitly consider the user’s query when evaluating scene importance. When a query is given, the prompt guides the language model to assess how well each scene aligns with the user’s content preference. 
Scenes that match key events or strongly relate to the query are given higher importance scores. In contrast, scenes that are irrelevant or conflict with the user’s intent are down-weighted.

This lightweight modification enables zero-shot, query-aware summarization via prompt engineering alone, without changing the model or core scoring logic.  A more detailed explanation, along with user-guided summarize examples, can be found in  Appendix (Section ~\ref{sec:appendix_LLM_prompt} in the appendix)).

%For implementation details and examples of query-based summaries, refer to Appendix~\ref{sec:appendix_implementation_details}.
\paragraph{User-guided extension to the scene scoring prompt.} We now give the extension of  the base scene scoring prompt (a full version of the prompt can be found in Section ~\ref{sec:appendix_LLM_prompt} in the appendix):% with the following addition when a user preference is provided:

\begin{quote}
\small % or \footnotesize, \scriptsize for smaller
\texttt{The user has provided the following content preference to guide the summarization: \\
\textbf{User Query: <user query>} \\
When assigning a score, consider how well the scene aligns with this preference. \\
Scenes that closely match or contradict the user’s intent should be scored accordingly,\\ reflecting their relevance or irrelevance to the desired summary focus. 
If the scene is not clearly related to this preference, assign a score based on the default scale and criteria below.}
\end{quote}

\subsection{Frame level scoring}
To obtain frame‑level importance scores, we follow a three‑stage procedure. First, each frame initially inherits the importance score of its parent scene, as shown in Figure~\ref{fig:general_pipeline_example}c. Then, a temporal smoothing function is applied around scene boundaries to avoid abrupt changes and create a gradual transition between adjacent scenes. After that, within every scene, we compute a representativeness weight for each frame (how well it reflects that scene). Finally, this weight is then fused with the smoothed scene score, yielding the final per‑frame importance value for the entire video.

\subsubsection{Scene to frames : scene score normalization and temporal smoothing function}\label{sec:scene_smoothing}

\begin{figure}[t !]
    \centering
    \includegraphics[width=1.0\linewidth]{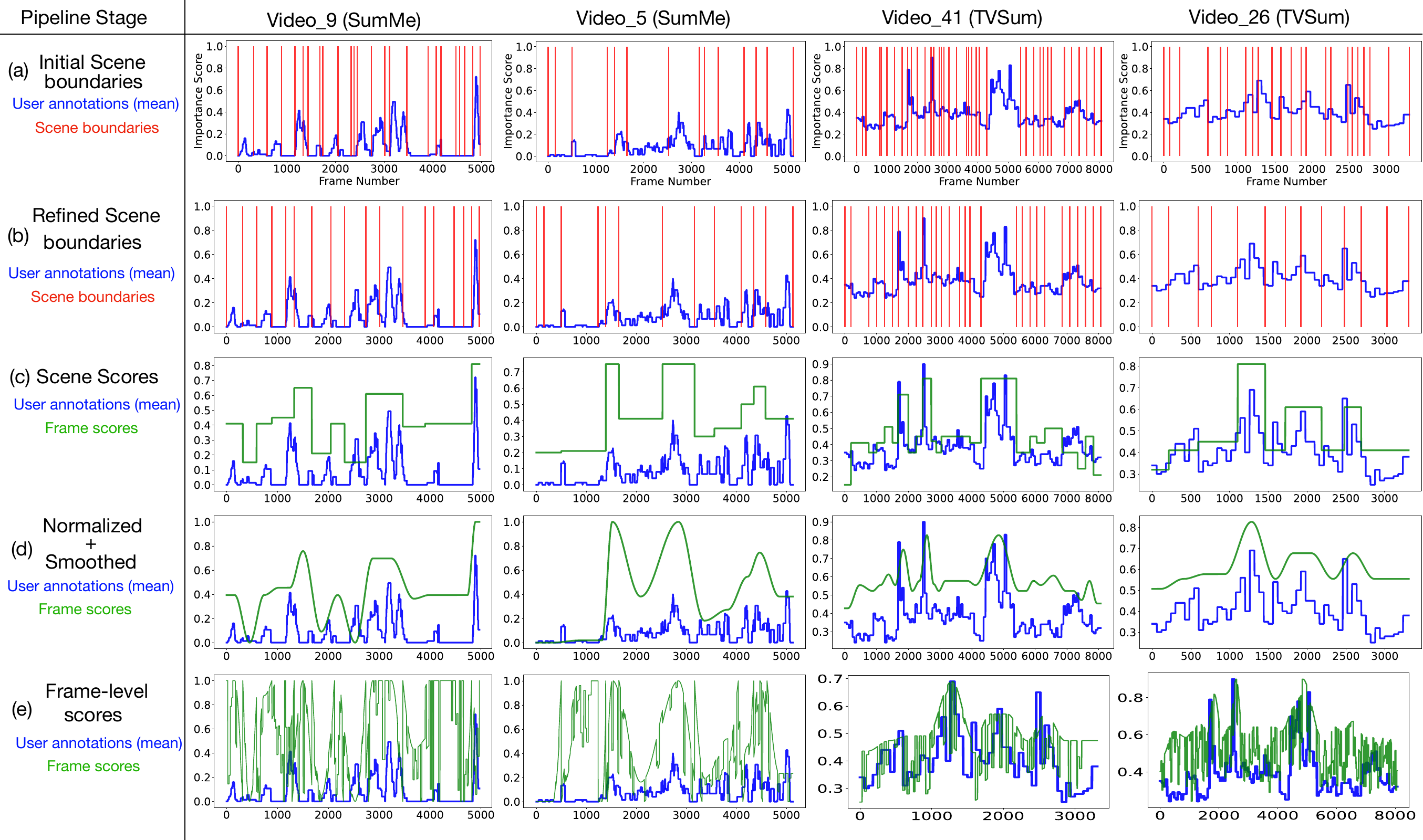}
    \caption{\textbf{Illustrative example of our video summarization pipeline.} The figure presents the step-by-step transformation for two sample videos from the SumMe dataset (Video\_5, Video\_9) and two from the TVSum dataset (Video\_26, Video\_41). (a) Initial scene boundaries are detected from raw frames. (b) Scene boundaries are refined based on embedding similarity. (c) Scene-level scores are computed using a language model conditioned on a user query (based on descriptions of each scene generate from a \vlm{}). (d) Scores are normalized and temporally smoothed to produce frame-level importance scores. (e) The predicted frame-level scores are compared with averaged user annotations, highlighting alignment between the model’s predictions and user intent.}
    \label{fig:general_pipeline_example}
\end{figure}

\paragraph{Goal and Approach:} 
After establishing scene boundaries (Figure~\ref{fig:general_pipeline_example}b) and computing the score of each scene (Figure~\ref{fig:general_pipeline_example}c), a normalization step is applied to the scores of the scenes to emphasize relative differences in importance as perceived by the LLMs. 
Since the scores are not derived from a binary scale (important or not important), but rather fall along a spectrum of relevance (1-100), normalization helps highlight subtle distinctions and amplifying the contrast between different scenes, ensuring that highly important scenes are more prominently reflected while less relevant scenes are de-prioritized. See Section~\ref{sec:standard_video_sum} for the exact used normalization procedure. Now, each frame within the scene is initially assigned the same normalized score as the scene to which it belongs (Figure~\ref{fig:general_pipeline_example}d).

In order to gradually blend a scene score to the score of its neighboring, we employ a temporal smoothing function, for each two consecutive scenes, starting from the midpoint of the first scene and ending in the midpoint of the second, a cosine function is used to interpolate between the score of the two scenes based on their relative positions (Figure~\ref{fig:general_pipeline_example}d). The transition helps eliminate unnatural jumps in scores values, resulting in more coherent and gradual shift across scene boundaries. 
This smoothing is crucial for generating a natural and visually consistent video summary, where the importance of each frame reflects not only the content of its own scene but also the surrounding temporal context.

\subsubsection{Frame weighting with in a scene}\label{sec:frame_weighting}
\paragraph{Goal:} To go beyond our scene-level importance scores—which already summarize contextual, semantic, and temporal cues—we now assign frame-level weights. Each weight quantifies how well a given frame describes its entire scene, letting us pinpoint the most representative and informative moments within that scene; i.e., assessing every frame a representativeness score within the scene.

\paragraph{Approach:}
First, we note that Algorithm ~\ref{alg:frame-level-weights} (in the appendix) provides the full details for this step. We first segment each scene into non-overlapping segments each of size $W$. For each segment, we define and compute two key properties to assess its contribution to the overall summary: \textbf{consistency} and \textbf{uniqueness}. These two scores aim to balance stability and novelty in the selected content. To compute these  properties, we begin by employing KMeans~\cite{lloyd1982least,pedregosa2011scikit} for clustering the frame embeddings of each scene, to reveal the underlying structure and visual patterns. The optimal number of cluster for each scene, donated as $K^*$, is selected dynamically by evaluating a predefined range of candidate values; For each candidate value $K_i \in \{K_{min}, K_{min} + \Delta{k}, \dots, K_{max} \}$, we compute Within-Cluster Sum of Squares (WCSS) ~\cite{bishop2006pattern}, donated, we then apply the Elbow method~\cite{thorndike1953belongs} to the WCSS curve to select $K^*$, the value at which the rate of decrease in WCSS significantly slows down—identified as the point with the greatest change in the slope of the WCSS curve.

\paragraph{Consistency.} Reflects the internal coherence of a segment, quantifying how uniformly its frames belong to the same visual pattern based on their cluster labels. A segment with high consistency suggests a stable and semantically meaningful portion of the video. Formally, given a segment $S$ of $W$ consecutive frames with cluster labels $\{l_{t+1}, l_{t+2}, \dots, l_{t+W}\}$, the consistency score is defined as:

\begin{figure}[h!]
\centering
\begin{minipage}{0.48\linewidth}
\begin{equation}
\text{Consistency}(S) = 
\frac{\max\limits_{c} \left| \{ l_{t+i} \mid l_{t+i} = c \} \right|_{i=1}^{W}}{W}
\label{eq:consistency}
\end{equation}
\end{minipage}%
\hfill
\begin{minipage}{0.51\linewidth}
\begin{equation}
\text{Uniqueness}(S) = 
\frac{1}{W} \sum_{i=1}^{W} \left\| \mathbf{f}_{t+i} - \bar{\mathbf{f}} \right\|_2\;;\; \bar{\mathbf{f}} = \frac{1}{W} \sum_{i=1}^{W} \mathbf{f}_{t+i}
\label{eq:uniqueness}
\end{equation}
\end{minipage}
\end{figure}

\paragraph{Uniqueness.} Measures how distinct the segment is with respect to internal variation in visual features. It captures how much the frame embeddings within the segment deviate from their mean, emphasizing visually diverse or rare content. 

\begin{figure}[h!]
\centering
\begin{minipage}{0.48\linewidth}
\begin{equation}
\label{eq:segment_weight}   
\adjustbox{scale=1.0}{$
\sigma \cdot \text{Consistency}(S) + (1 - \sigma) \cdot \text{Uniqueness}(S)
$}
\end{equation}
\end{minipage}%
\hfill
\begin{minipage}{0.48\linewidth}
\begin{equation}
F_S(\text{Video}) =
\begin{cases}
\{\sigma{=}0.1,\; W{=}1\,\text{s} \} & \text{if } T > 5S \\
\{\sigma{=}1.0,\; W{=}1\,\text{s} \} & \text{if }  5S \ge T > S \\
\{\sigma{=}0.3,\; W{=}3\,\text{s} \} & T < S
\end{cases}
\label{eq:Dynamic_sigma_segment}
\end{equation}
\end{minipage}
\end{figure}

\paragraph{Combined Segment Weight.}  
We combine consistency and uniqueness into a unified score to assign an overall weight to each segment. This weighting balances stable (consistent) regions and distinctive (unique) content, and is governed by the parameter $\sigma$, which adjusts their relative importance. All frames within a segment inherit the computed segment weight. Furthermore, the frame-consistency weight $\sigma$ and segment duration $W$ are dynamically determined based on the video duration $T$, as shown in Equation~\ref{eq:Dynamic_sigma_segment}.  Here, $S$ denotes a short-video duration threshold used to categorize videos into three classes:  
(i) \textit{Long videos} ($T > 5S$),  
(ii) \textit{Intermediate videos} ($S < T \leq 5S$), and  
(iii) \textit{Short videos} ($T \leq S$).  
We set $S = 1.8$ minutes (approximately 100 seconds), so a video is considered long if its duration exceeds 9 minutes. Additional insights behind these design choices are discussed in Section~\ref{sec:ablation}.

\subsubsection{Frame-level importance scoring}\label{sec:final_frame_scores}
\paragraph{Goal:} Fusing scene importance and frame weights to create per-frame importance. 

%First, we begin by normalizing the scene scores, utilizing $F_{norm}$, and activating the transition (smoothing) function (algorithm~\ref{alg:scene-scores-smoothing}) to ensure smooth transitions between consecutive scene, donated as $F_{smooth}$. %Next, we proceed to compute the frame weights, donated as $F_{weights}$ , employing algorithm~\ref{alg:frame-level-weights},  with $\sigma$ acting as a parameter to control the balance between the consistency and uniqueness contribution. 
\paragraph{Approach:}
The final step calculates frame-level importance scores, which are the output of our algorithm; providing the importance each frame within the video.
Given the smoothed per-frame score (from Section~\ref{sec:scene_smoothing}) and per-frame wights (as in Section~\ref{sec:frame_weighting}), for each frame within a scene, we compute the final frame scores as the product of the smoothed scores, $F_{smooth}$ and the frame weights, $F_{weights}$. These frame-level scores are then normalized, by applying the same normalization function $F_{norm}$, within each scene. We apply $F_{norm}$ of frame scores within each scene-rather than the entire video, to ensure that the frame's importance is evaluated based on the context of their own scene and not based on the context of the overall video.% ensuring that every scene contributes fairly to the final summary.

The result is a set of frame-level importance scores (Figure~\ref{fig:general_pipeline_example}e) $\tilde{S}$, which represent the contribution of each frame to the overall video summary. This process, as shown in Algorithm~\ref{alg:frame-level-scoring} in the appendix, ensures that the frame-level importance scores take into account both the scene’s intrinsic importance, the temporal and contextual relationships between frames.

\section{Experimental results}
In this section, we conduct extensive experiments to compare our approach with state-of-the-art methods on both standard and text-guided video summarization tasks, followed by a detailed ablation study of each component in our pipeline.

\subsection{Standard video summarization}\label{sec:standard_video_sum}

\paragraph{Datasets.}
We evaluated our approach on two benchmark datasets: \textit{SumMe}\cite{gygli2014creating} and \textit{TVSum}\cite{song2015tvsum}.
SumMe contains 25 user videos (1–6 minutes) of everyday activities, while TVSum includes 50 YouTube videos from 10 activity categories.
Both datasets provide multiple user annotations of video importance, enabling evaluation of summary relevance and coverage, additional datasets details are provided in Section~\ref{sec:appendix_datasets} in the appendix.

\paragraph{Evaluation protocol.} 
We follow the classic key-shot-based evaluation protocol introduced by~\cite{zhang2016video}, which is widely adopted by state-of-the-art video summarization methods. 
This metric evaluates the similarity between the generated summary ($A$), constrained to be less than 15\% of the original video duration, and the user-provided summary ($B$). 
The comparison is based on their temporal overlap, measured using (P)recision and (R)ecall as follows:

\begin{figure}[ht!]
    \centering
    \begin{minipage}{0.33\linewidth}
    \begin{equation}
    \label{eq:P}
    \begin{aligned}
    \text{(P)recision} &= \frac{|A \cap B|}{|A|} 
    \end{aligned}
    \end{equation}
    \end{minipage}
    \begin{minipage}{0.33\linewidth}
        \begin{equation}
        \label{eq:R}
        \begin{aligned}
        \text{(R)ecall} &= \frac{|A \cap B|}{|B|}        
        \end{aligned}
        \end{equation}
    \end{minipage}
    \hfill
    \begin{minipage}{0.33\linewidth}
    \begin{equation}
    \label{eq:f1}
    \begin{aligned}
    \text{F} &= 2 \cdot \frac{P\cdot R}{P+R} 
    \end{aligned}
    \end{equation}
    \end{minipage}
\end{figure}

\noindent
Here, $|A|$ denotes the duration of the set $A$, measured in number of frames. 
Finally, we compute the F1-score ($F$), which represents the harmonic mean of (P)recision and (R)ecall, as shown in Equation~\ref{eq:f1}.

\begin{table}[t !]
\centering
\caption{\textbf{Performance on TVSum and SumMe.} Every baseline—supervised or "unsupervised"—relies on a training stage. Our pipeline is truly zero-shot: it needs no tuning, no access to ground-truth labels, and no prior knowledge of the datasets. Despite that handicap, it sets a new state-of-the-art among unsupervised methods and matches the best supervised models.}
\label{tab:SOTA_performance_TVSum_sumMe}
\begin{adjustbox}{width=1.0\textwidth}
\begin{tabular}{|c|c|c|c|c|}
\hline
\textbf{Method name} & \textbf{Doesn't require training data} & \textbf{Text-queryable} & \multicolumn{2}{c|}{\textbf{Dataset}} \\ \hline
                                    \multicolumn{3}{|c|}{\textbf{Supervised methods}}  & SumMe & TVSum \\ \hline
re-SEQ2SEQ~\cite{ghauri2021supervised}                                  & \textcolor{red}{\ding{55}} & \textcolor{red}{\ding{55}}   & -     & 63.9 \\
CLIP-It~\cite{narasimhan2021clip}                 & \textcolor{red}{\ding{55}} & \textcolor{green}{\ding{51}} & 54.2     & 66.3 \\
MAVS ~\cite{feng2018extractive}                                       & \textcolor{red}{\ding{55}} & \textcolor{red}{\ding{55}}   & -     & 67.5 \\
iPTNet~\cite{jiang2022joint}                                  & \textcolor{red}{\ding{55}} & \textcolor{red}{\ding{55}}   & 54.5  & 63.4    \\
A2Summ~\cite{he2023align}                                  & \textcolor{red}{\ding{55}} & \textcolor{red}{\ding{55}}   & 55.0  & 63.4    \\
PGL-SUM~\cite{apostolidis2021combining}                                     & \textcolor{red}{\ding{55}} & \textcolor{red}{\ding{55}}   & 57.1  & -    \\ \hline \hline
\multicolumn{5}{|c|}{\textbf{Unsupervised methods}} \\ \hline
AC-SUM-GAN~\cite{apostolidis2020ac}                              & \textcolor{red}{\ding{55}} & \textcolor{red}{\ding{55}}   & 50.8  & 60.6 \\
RS-SUM~\cite{abbasi2023adopting}                                  & \textcolor{red}{\ding{55}} & \textcolor{red}{\ding{55}}   & 52.0  & 61.4 \\
SegSum~\cite{vo2025integrate}                  & \textcolor{red}{\ding{55}} & \textcolor{red}{\ding{55}}   & 54.07 & 62.02 \\ \hline \hline

\multicolumn{5}{|c|}{\textbf{Zero-shot methods}} \\ \hline

\textbf{Random}           & \textcolor{green}{\ding{51}}   & \textcolor{red}{\ding{55}} & 44.89 & 56.43 \\

Contrastive (Zero-Shot)~\cite{pang2024unleashing}  & \textcolor{green}{\ding{51}} & \textcolor{red}{\ding{55}} & 47.2 & 58.4 \\

STeMI (Few-shot)~\cite{li2025spatial}    & \textcolor{red}{\ding{55}} & \textcolor{red}{\ding{55}} & 54.65 & 60.89 \\

\textbf{Ours}           & \textcolor{green}{\ding{51}}   & \textcolor{green}{\ding{51}} & \textbf{56.84} & \textbf{62.22} \\ \hline
\end{tabular}
\end{adjustbox}
\end{table}

In SumMe, annotations are provided as key-fragments, this formula is directly applied. However, in the TVSum, annotations are provided at the frame level, with translation from the frame level to the key fragment using methodologies presented in~\cite{zhang2016video}. After obtaining key-fragment annotations for a given test video, we calculate the F1-score for the generated summary against each user summary for that video. The F1 scores for the TVSum dataset are averages as proposed in~\cite{song2015tvsum}, while for the SumMe dataset, the F1 score for a given test video is the maximum F1 score calculated for that video, as proposed in its paper~\cite{gygli2015video}. To obtain the final F1 score for each data set, we calculate the mean F1 score in the test videos, and take the mean results in 5 different splits of generated video tests, following the methodology in ~\cite{zhang2016video}. This score serves as a performance metric.

\paragraph{Design choices.}
For each dataset, one of three normalization functions: Min-Max, Exponential, or a hybrid Min-Max~+Exponential, is selected based on dataset characteristics, as described in Section~\ref{sec:scene_smoothing}.

\paragraph{Normalization in SumMe.}
Min-Max normalization linearly scales values between 0 and 1, preserving relative distances but remaining sensitive to outliers. This property makes it well-suited to the \textit{SumMe} dataset, where user-selected key-frames often misaligned due to subjective preferences. As reported in~\cite{zhang2016video}, the evaluation protocol for SumMe selects the closest user summary by taking the maximum F1-score across all references. Accordingly, we adopt Min-Max normalization to better capture the variability inherent in human annotations.

\paragraph{Normalization in TVSum.}
In contrast, Exponential normalization amplifies differences among higher values, emphasizing salient segments. This makes it particularly effective for the \textit{TVSum} dataset, where user annotations are more consistent and evaluation is based on the average F1-score across all users. Therefore we employ Exponential normalization to align importance scoring with the dataset’s lower inter-user variability.

\paragraph{Combined normalization.}
The hybrid Min-Max+Exponential approach balances local contrast with global consistency in frame-level scores. To confirm our choices, we conducted comparative experiments across all normalization methods and parameter settings on both datasets; results are presented in Section~\ref{sec:ablation}.

From this point forward, we employ Min-Max normalization for the SumMe dataset and Exponential normalization for the TVSum dataset.

\subsubsection{Results}\label{sec:sumMe_TVSum_QUN}

\begin{wraptable}{r}{0.3\textwidth} % r = right, width ~ 35% of text width
\centering
\caption{\textbf{Performance on TVSum and SumMe.}}
\label{tab:Our_Precision_Recall_F1-Score}
\begin{tabular}{|c|c|c|}
\hline 
\textbf{Metric} & \textbf{SumMe} & \textbf{TVSum} \\ \hline
(P) & 55.77 & 62.22 \\
(R) & 57.96 & 62.22 \\
F1  & 56.84 & 62.22 \\
\hline
\end{tabular}
\end{wraptable}

\textbf{Precision, recall, and F1 score.} We evaluated our model’s precision, recall, and F1 score on the test sets of SumMe and TVSum to provide a fare comparison to other methods (no training is needed in our case).  First, Table~\ref{tab:Our_Precision_Recall_F1-Score} reports the results of our method; the close alignment between precision, recall, and F1 across both datasets suggests that the model reliably selects relevant segments while maintaining strong coverage of human annotations. Notably, SumMe exhibits slightly higher precision and recall margins than TVSum, likely due to greater variability in its human-labeled summaries as mentioned before. With our conclusive scores of 56.84\% on the SumMe dataset and 62.22\% on the TVSum dataset in Table~\ref{tab:Our_Precision_Recall_F1-Score}, these outcomes will be used for comparative analysis with other methods.

\paragraph{Performance Comparison.}
Although our approach performs competitively with some supervised methods, our main objective is to evaluate its performance against prior \textit{unsupervised} (see Section~\ref{sec:sup_discussion_gap_supervised} in the appendix for details on why supervised methods has advantage on our work). Notably, our model is entirely \textbf{data-free} and supports \textbf{text-based user guidance}.
Table~\ref{tab:SOTA_performance_TVSum_sumMe} presents a comparative analysis showing that our zero-shot unsupervised model outperforms all existing unsupervised , zero and few-shot methods. 

Specifically, our approach achieves a \textbf{2.8\%} improvement on \textit{SumMe} and a \textbf{0.2\%} improvement on \textit{TVSum} compared to the state-of-the-art method of~\cite{vo2025integrate}, which combines KTS-based temporal segmentation~\cite{potapov2014category} with the CA-SUM architecture~\cite{abbasi2023adopting}. Furthermore, we observe gains of \textbf{4.8\%} on \textit{SumMe} and \textbf{0.82\%} on \textit{TVSum} over~\cite{abbasi2023adopting}, which leverages a self-supervised language model training approach.

In addition, our approach consistently improves over recent zero-shot and few-shot baselines that do not support text-queryable video summarization. In particular, it outperforms the contrastive(zero-shot)~\cite{pang2024unleashing} method. We also surpass the few-shot STeMI method~\cite{li2025spatial}, achieving gains of \textbf{2.19\%} on \textit{SumMe} and \textbf{1.33\%} on \textit{TVSum}. Overall, these results demonstrate that our zero-shot, language-guided framework remains competitive with state-of-the-art methods—while uniquely enabling flexible, query-conditioned summarization.

Importantly, our method achieves these improvements \textbf{without any training data}, demonstrating the potential of re-purposing \vlm{}s and LLMs for video summarization without prior knowledge of user annotations or data distributions.

\begin{figure}[t!]
    \centering
    \resizebox{1.0\linewidth}{!}{% ← scales everything inside to 90% of text width
    \begin{minipage}{1.0\linewidth}
        \begin{subfigure}[b]{0.5\linewidth}
            \centering
            \includegraphics[width=\linewidth]{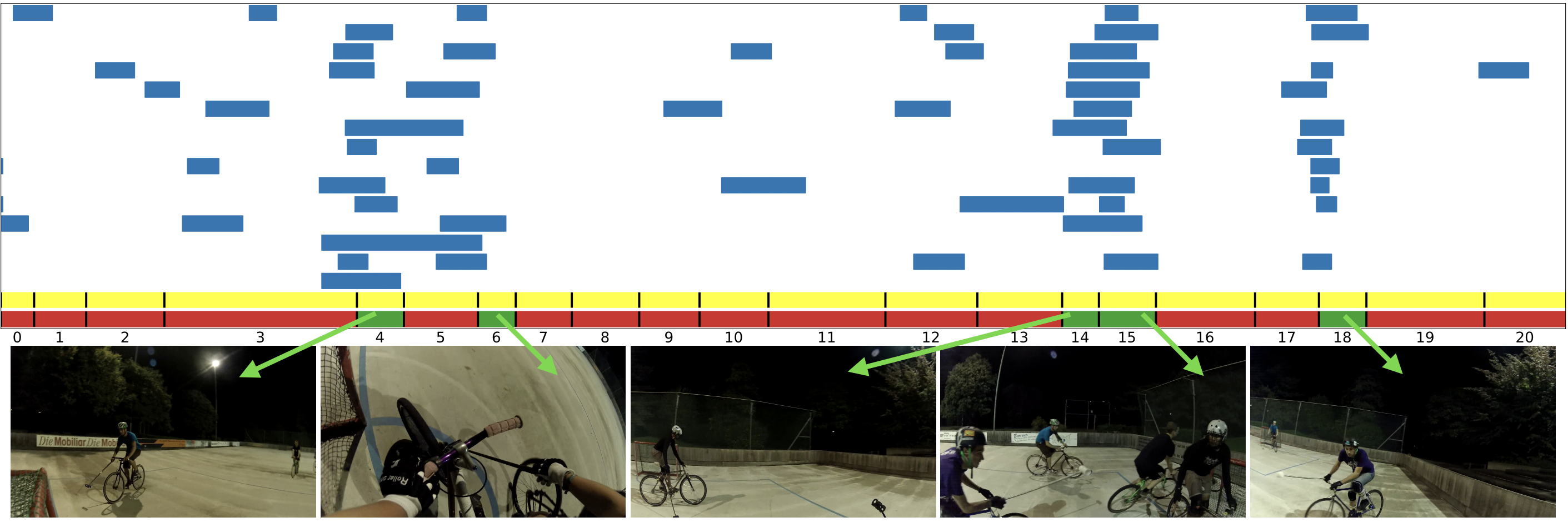}
            \caption{Video summary for "Bike Polo" (SumMe)}
            \label{fig:Bike_Polo_QUALITATIVE}
        \end{subfigure}
        \hfill
        \begin{subfigure}[b]{0.49\linewidth}
            \centering
            \includegraphics[width=0.95\linewidth]{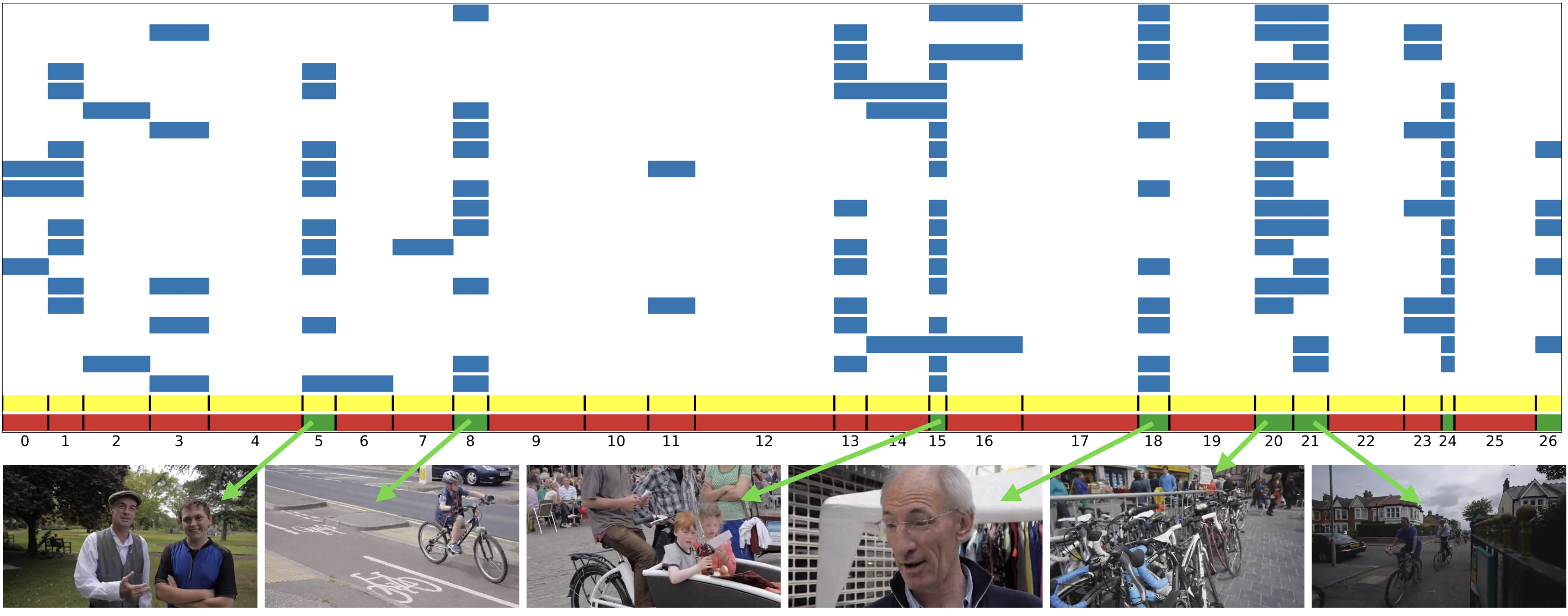}
            \caption{Video summary for "4wU\_LUjG5Ic" (TVSum)}
            \label{fig:4wU_LUjG5Ic_QUALITATIVE}
        \end{subfigure}
    \end{minipage}
    } % end resizebox
    \caption{\textbf{Visualization of selections made by our method, user summaries, and KTS segments.} 
    Blue bars: human summaries; yellow bars: final summary fragments; green bars: our model’s selected frames; red bars: frames not selected. 
    Video frames from each selected shot are included for better visualization.}
    \label{fig:TVSum_SumMe_QUALITATIVE}
\end{figure}

\paragraph{Qualitative Examples.}
To visualize our model’s temporal selection behavior, we first introduce Kernel Temporal Segmentation (KTS)\cite{potapov2014category}, a common method for partitioning videos into temporally coherent segments and often used to generate training labels in supervised settings.
We then present qualitative comparisons (see Figure~\ref{fig:TVSum_SumMe_QUALITATIVE}). of our model’s selected segments (in red) against user summaries (blue) and ground truth KTS segments (yellow) for videos from the \textit{SumMe} and \textit{TVSum} datasets.

For instance, in the “Bike Polo” video from \textit{SumMe} (Figure~\ref{fig:Bike_Polo_QUALITATIVE}), user summaries do not align well with KTS segments, making it difficult to identify optimal summary regions. Nevertheless, our model successfully identifies key segments (14, 15 and 16) that best reflect user preferences.
Conversely, in the “4wU\_LUjG5Ic” video from \textit{TVSum} (Figure~\ref{fig:4wU_LUjG5Ic_QUALITATIVE}), user summaries align well with both KTS segmentation and our predictions. However, our model’s final selected segment (26) slightly diverges from user choices—likely due to summary length constraints or under-representation in the generated textual description, which led the LLM to assign lower relevance scores.

\subsection{Query-Focused video summarization}
\paragraph{Datasets.}
For evaluating our query-focused video summarization approach, we use the \textit{QFVS} dataset~\cite{sharghi2017query}, which extends the UT Egocentric (UTE) dataset~\cite{lee2012discovering}.
It consists of four long egocentric videos (3–5 hours each), where each 5-second shot is tagged with multiple visual concepts such as \textit{CAR}, \textit{STREET}, and \textit{SKY}.
These semantic tags are paired with user queries and annotated summaries, enabling query-conditioned video summarization, additional datasets details are provided in Section~\ref{sec:appendix_datasets} in the appendix.

 \paragraph{Evaluation protocol.} Although prior work predicts a single score per shot, we predict score for each frame. To obtain a single score per shot we follow the same methodologies of ~\cite{narasimhan2021clip}, utilizing the mean strategy-taking the average score of frames per shot-then the generated summary is obtained by taking the shots with the highest scores, the summary budget is set to be exactly as long as video-query Oracle summary as in ~\cite{sharghi2017query}.

\paragraph{Quantitative results.}
In Table~\ref{tab:GFVS_performance}, we compare the F1 scores of our method against five supervised baselines: Seq-DPP~\cite{gong2014diverse}, SH-DPP~\cite{sharghi2016query}, QFVS~\cite{sharghi2017query}, CLIP-It: ResNet~\cite{narasimhan2021clip}, and CLIP-It: CLIP Image~\cite{narasimhan2021clip}. Following the evaluation protocol of~\cite{sharghi2017query}, which performs four rounds of experiments by leaving out one video for testing and one for validation while training on the remaining two, we evaluate our model—despite being entirely training-free—on four independent splits, each containing a single test video. We then average the F1 scores across these splits to obtain the final reported performance.

As shown in Table~\ref{tab:GFVS_performance}, our \textbf{training-free, zero-shot} approach consistently outperforms the four classical supervised baselines on the QFVS dataset. The only method surpassing us slightly is the data-hungry CLIP-It~\cite{narasimhan2021clip}, which requires exhaustive per-frame annotation and is highly sensitive to the fine-tuned image backbone. When CLIP-It relies on ResNet-50~\cite{he2016deep}, our method leads by \textbf{+1 pp} in F\textsubscript{1}, despite using \emph{no} task-specific data. For a like-for-like comparison, we also adopt CLIP embeddings~\cite{radford2021learning}; in this configuration, our zero-shot model achieves an F\textsubscript{1} of \textbf{53.42\%}, virtually matching the \textbf{54.44\%} obtained by the fully supervised CLIP-It.
A per-video analysis further underlines our robustness: we surpass CLIP-It by +2 pp on Video 2
and equal its performance on the notoriously challenging Video 4.

\paragraph{Further analysis.}
To assess the influence of different normalization techniques on performance, we evaluated Min–Max normalization, Exponential normalization, and a combined approach. The combined method was motivated by its strong performance on the SumMe and TVSum datasets, aiming to capture the benefits of both scale consistency and enhanced contrast. As anticipated, the combined normalization yielded the best results across all four videos,as shown in Table~\ref{tab:GFVS_performance}, 
confirming its ability to balance sensitivity and robustness in representing importance scores.

\paragraph{Qualitative example.}
To demonstrate our model’s ability to guide the summarization process based on different user queries expressed in natural language, we present qualitative results in Figure~\ref{fig:QFVS_QUALITATIVE}. Specifically, we apply our model to an input video (Video~3) from the QFVS dataset—a 3-hour egocentric video recorded using a head-mounted camera. The video captures a wide range of daily activities, including driving to various locations, shopping at a supermarket, and preparing and eating food.

\noindent Summary1 and Summary2 illustrate sampled frames from the summaries generated by our model in response to two distinct user queries:
(i) “Focus on scenes containing \textbf{Chair} and/or \textbf{Tree},” and
(ii) “Focus on scenes containing \textbf{Food} and/or \textbf{Hands}.”

These examples highlight the model’s capability to assign high importance scores to segments, scenes, or events that align with user-defined intent. Notably, although the input video remains identical, the generated summaries differ significantly depending on the provided query—demonstrating the model’s adaptability to user focus.

\begin{table*}[t !]
\centering
\caption{\textbf{Performance on Query Focused Video Summarization (QFVS).} All competing methods train end-to-end on shot-level annotations. Our pipeline, in contrast, is \emph{annotation-free}: it assigns frame-level importance scores with zero training or dataset knowledge. Even in this strict zero-shot setting, we finish a close second on QFVS—just one F\textsubscript{1} point behind the fully supervised, data-hungry CLIP-It.}

\begin{adjustbox}{width=1.0\linewidth}
\label{tab:GFVS_performance}
\begin{tabular}{|c|c|c|c|c|c|c|c|}
\hline
\textbf{Method name}          & \textbf{Property}       & \textbf{Doesn't require training data}                             & Video 1    & Video 2     & Video 3     & Video 4    & \textbf{Average} \\ \hline
SeqDPP~\cite{gong2014diverse}                & \textcolor{red}{Supervised}           & \textcolor{red}{\ding{55}}                           & 36.59    & 43.67     & 25.26     & 18.15    & 30.92\\ \hline
SH-DPP~\cite{sharghi2016query}               & \textcolor{red}{Supervised}           & \textcolor{red}{\ding{55}}                           & 35.67    & 42.74     & 36.51     & 18.62    & 33.38\\ \hline
QFVS~\cite{sharghi2017query}                 & \textcolor{red}{Supervised}           & \textcolor{red}{\ding{55}}                           & 48.68    & 41.66     & 56.47     & 29.96    & 44.19\\ \hline
QSAN~\cite{xiao2020query}                    & \textcolor{red}{Supervised}           & \textcolor{red}{\ding{55}}                           & 48.52   & 46.64     & 56.93     & 34.25   & 46.59\\ \hline
IntentVizor~\cite{wu2022intentvizor}                & \textcolor{red}{Supervised}           & \textcolor{red}{\ding{55}}                               & 51.27   & 53.48     & 61.58     & 37.25   & 50.9\\ \hline
CLIP-It : ResNet~\cite{narasimhan2021clip}     & \textcolor{red}{Supervised}    & \textcolor{red}{\ding{55}}                           & 55.19    & 51.03     & 64.26     & 39.47    & 52.49 \\ \hline
CLIP-It : CLIP-Image~\cite{narasimhan2021clip} & \textcolor{red}{Supervised}    & \textcolor{red}{\ding{55}}                           & \textbf{57.13}    & \underline{53.60}     & \textbf{66.08}     & \textbf{41.41}    & \textbf{54.44}\\ \hline
\textbf{Ours} : MinMax norm          & \textcolor{green}{Zero-Shot}     & \textcolor{green}{\ding{51}}         & 51.47    & 50.86     & 61.29     & 37.51    & 50.28 \\ \hline
\textbf{Ours} : Exp norm             & \textcolor{green}{Zero-Shot}    & \textcolor{green}{\ding{51}}         & 48.66    & 50.00     & 55.26     & 42.63    & 49.14\\ \hline
\textbf{Ours} : (MinMax + Exp) norm  & \textcolor{green}{Zero-Shot}      & \textcolor{green}{\ding{51}}         & \underline{53.57}    & \textbf{55.66}     & \underline{63.25}     & \textbf{41.20}    & \underline{53.42} \\ \hline
\end{tabular}                   
\end{adjustbox}
\end{table*}

\subsection{Fine-grained video summarization}
Although QFVS is a useful text-guided benchmark, its queries are mostly are limited to short, literal keywords (e.g.\ \textsc{car}, \textsc{street}, \textsc{pets}) and therefore do not probe temporal reasoning or long-tail semantics. To evaluate fine-grained, intent-aware summarization, we introduce a new dataset, \mydata.

\paragraph{\mydata{}.} Our new benchmark for fine-grained, reasoning-driven video summarization from natural language queries, 
 consists of 9 videos ranging from 2–6 minutes in duration, covering a diverse range of content including sports, DIY projects, movie trailers, award-winning CGI videos and selected videos from Youtube-8M dataset ~\cite{abu2016youtube}.  Each video is paired with one to three queries, yielding 20 video–query pairs in total (full dataset, more statistical data and annotation reliability are provided in Section~\ref{sec:appendix_TGVS_details} in the appendix).  Unlike QFVS—where queries which merely refer to obvious visual entities—\mydata\ poses richer, more nuanced prompts that test intent interpretation, temporal reasoning, and the use of external knowledge. \mydata{} queries not only prioritize certain content but also serve as filters to exclude scenes, e.g., filtering violence or Excluding PG-13+ material. We group the queries into four classes:\\
(i) \textbf{Standard: } Queries that describe a specific object or action observable in the video, similar to QFVS but potentially multi-word. \\
(ii) \textbf{Standard with Special Attribute: } Queries that reference visual content along with a descriptive attribute, requiring models to focus on both presence and characteristics.\\
(iii) \textbf{Reasoning: } Queries that require some level of temporal or contextual reasoning beyond simple object recognition.\\
(iv) \textbf{Reasoning with General Knowledge:} Queries that rely on external knowledge or a broader understanding of context to identify relevant segments.
This design  provides a challenging testbed for models that aim to summarize videos in a genuinely query-focused, knowledge-aware manner.

\paragraph{Ground truth annotation.}
To construct reliable supervision for each video–query pair, we divided each video into non-overlapping 2-second segments, following the segmentation protocol used in the TVSum dataset~\cite{song2015tvsum}. Each segment was rated on a 1–5 scale according to its relevance to the given query, where 1 indicates “not relevant” and 5 indicates “highly relevant.”  
If a segment was not relevant to the query, it was then evaluated for its general importance to the video’s main theme, narrative, or key events—also using the 1–5 scale.  
This dual-scoring approach ensures that both query-relevant and contextually important segments are represented in the ground truth. Initial annotations were created by one annotator (the author) and subsequently reviewed by two additional individuals to ensure consistency, accuracy, and overall quality. The resulting annotations provide a robust basis for evaluating both query-focused and generic summarization performance.

\paragraph{Evaluation protocol.}
We follow the standard key-shot-based evaluation method~\cite{zhang2016video}, widely used in video summarization research. This approach measures the similarity between the generated summary (A), constrained by a duration budget, and the user-provided ground truth summary (B) through temporal overlap, quantified via precision (P), recall (R), and their F1-score, as defined in Equations~\ref{eq:P}, ~\ref{eq:R} and~\ref{eq:f1}.

\begin{figure}[t]
    \centering
    % ------------------- Image 1 -------------------
    \begin{minipage}[t]{0.53\linewidth}
        \centering
        \includegraphics[width=0.95\linewidth]{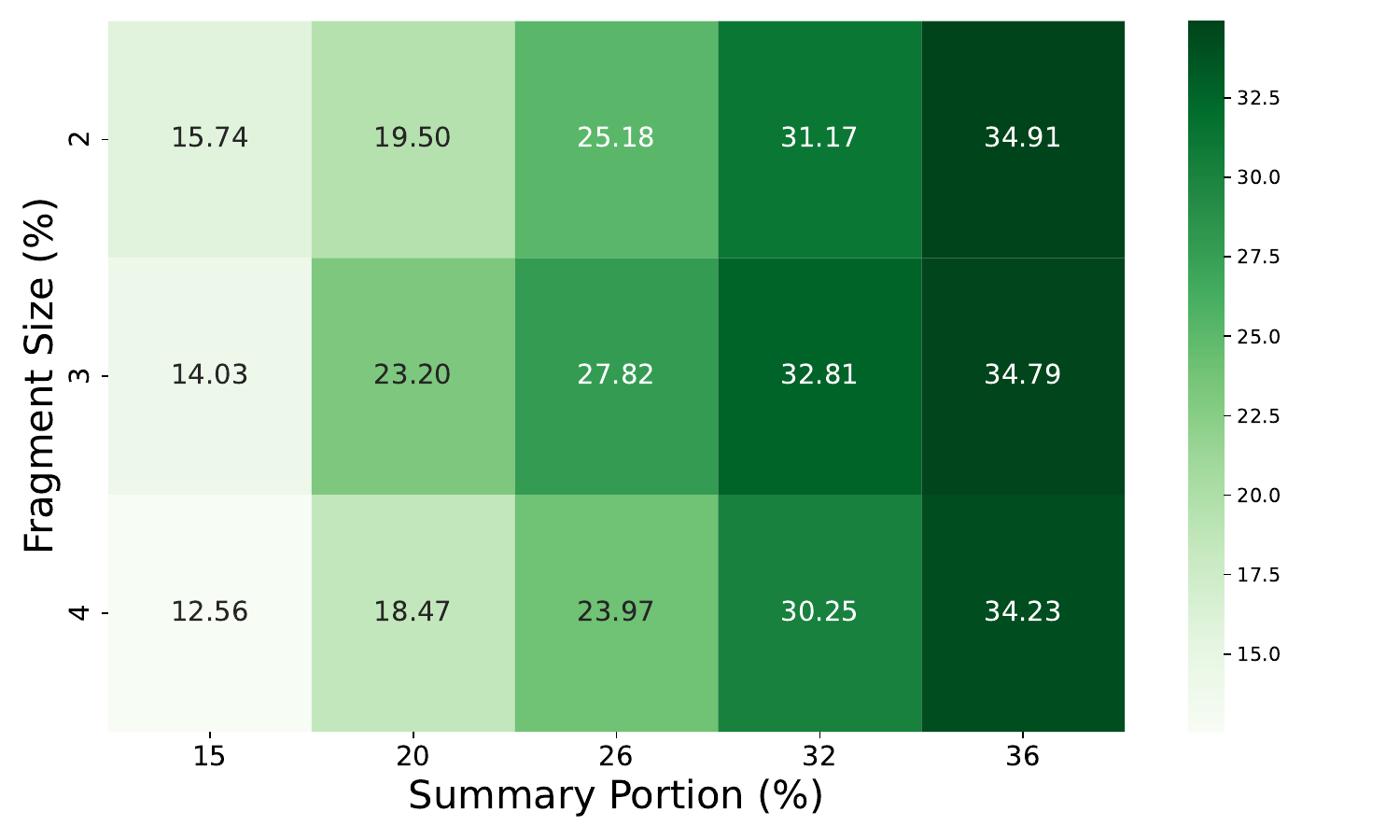}
        \caption{\textbf{Precision-over-Random (PoR) heatmap for different fragment sizes and summary portions.} 
        The figure illustrates the impact of varying fragment sizes and summary budgets on the performance of randomly generated frame-level importance scores. Higher PoR scores are observed for smaller fragment sizes and larger summary portions, indicating better alignment between randomly sampled summaries and user intent.}
        \label{fig:TGVS_PoR_HeatMap}
    \end{minipage}%
    \hfill
    % ------------------- Image 2 -------------------
    \begin{minipage}[t]{0.45\linewidth}
        \centering
        \hspace{-0.5cm}
        \includegraphics[width=0.95\linewidth]{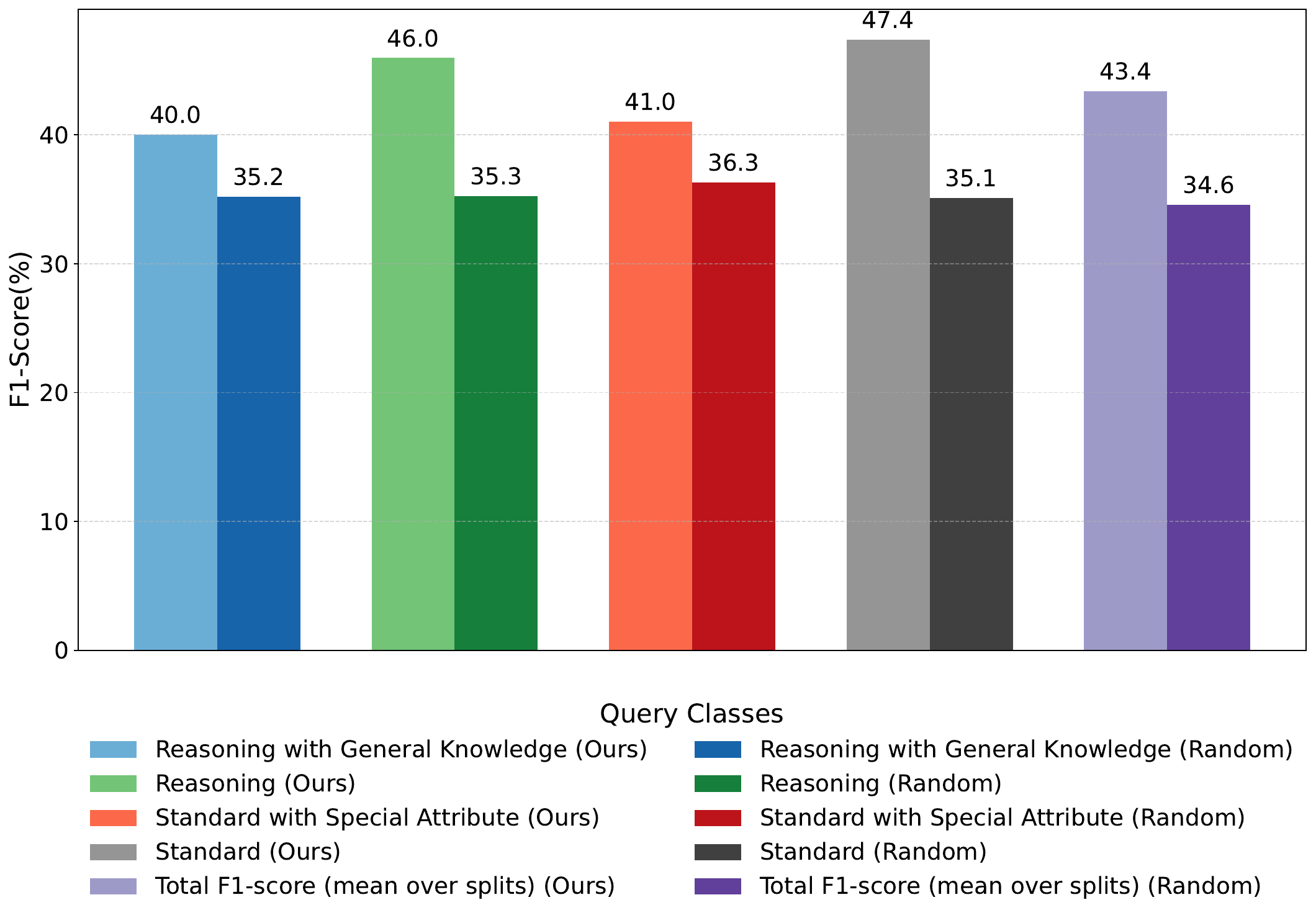}
        \caption{\textbf{Comparison of F1-scores between our approach and the random baseline across the four query classes in the ~\mydata{} dataset.} 
        Our model consistently outperforms the random baseline across all classes. The average F1-score is 43.4 for our model and 34.56 for the random baseline, demonstrating the effectiveness of query-guided summarization.}
        \label{fig:TGVS_categry_ours_vs_random}
    \end{minipage}
\end{figure}

To assess the robustness of our framework, we also analyze the effect of two key parameters: the \textit{fragment size} (i.e., the granularity of video segmentation) and the \textit{summary budget} (i.e., the allowed percentage of the video duration in the summary). Following~\cite{zhang2016video}, we compute the Precision-over-Random (PoR) score by comparing our model’s performance against 100 randomly generated importance score vectors (Figure~\ref{fig:TGVS_PoR_HeatMap}).  

Results indicate that smaller fragments generally yield higher PoR scores, while larger budgets produce more comprehensive summaries. Specifically, a fragment size of 2\% with a budget of 36\% achieves the highest accuracy over the random baseline. To balance summary coverage, query relevance, and fairness against random performance, we adopt a fragment size of 3\% and a summary budget of 36\% in all subsequent experiments, ensuring consistent evaluation across videos and queries in \mydata{}.

The final f1-score, as shown in Figure~\ref{fig:TGVS_categry_ours_vs_random}, was obtained by evaluating our model on 5 randomly generated non-overlapping splits, each containing 4 video-query pairs, with varying queries and different levels of summarization difficulty. 

\begin{figure*}[t]
    \centering
    % --- First figure ---
    \begin{minipage}[t]{0.48\textwidth}
        \centering
        \hspace{-1em}
        \includegraphics[width=0.97\linewidth]{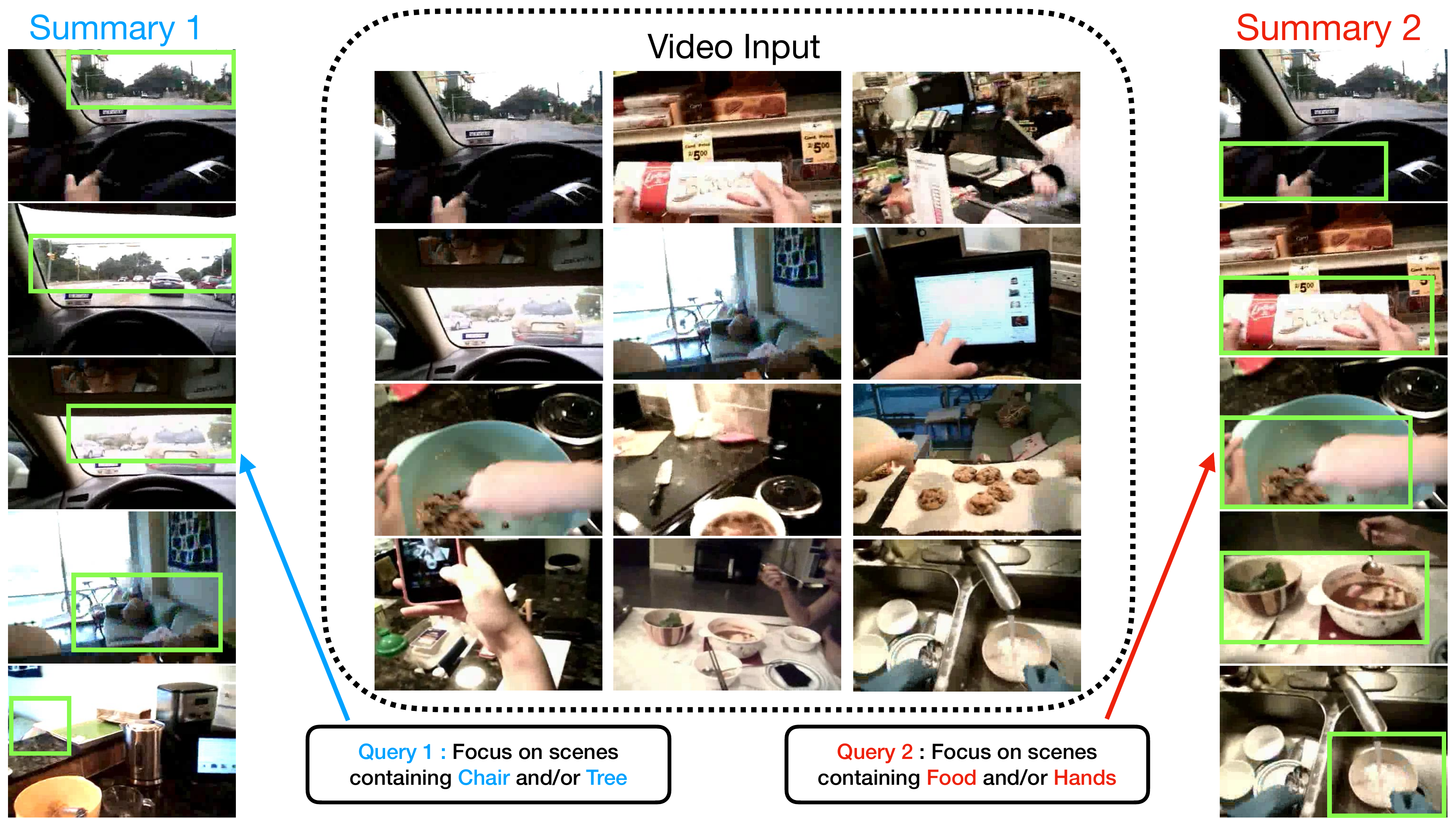}
        \caption{\textbf{Result of our method on the QFVS dataset.} 
        The middle columns framed by black outlines showcase frames sampled from Video 3 (3 hours long) in the QFVS dataset. 
        Summary 1 shows sampled frames for the query ``Focus on scenes containing \textbf{Chair} and/or \textbf{Tree}.'' 
        Summary 2 shows sampled frames for the query ``Focus on scenes containing \textbf{Food} and/or \textbf{Hands}.'' 
        Green boxes highlight the entities specified in each query.}
        \label{fig:QFVS_QUALITATIVE}
    \end{minipage}
    \hfill
    % --- Second figure ---
    \begin{minipage}[t]{0.49\textwidth}
        \centering
        \includegraphics[width=1\linewidth]{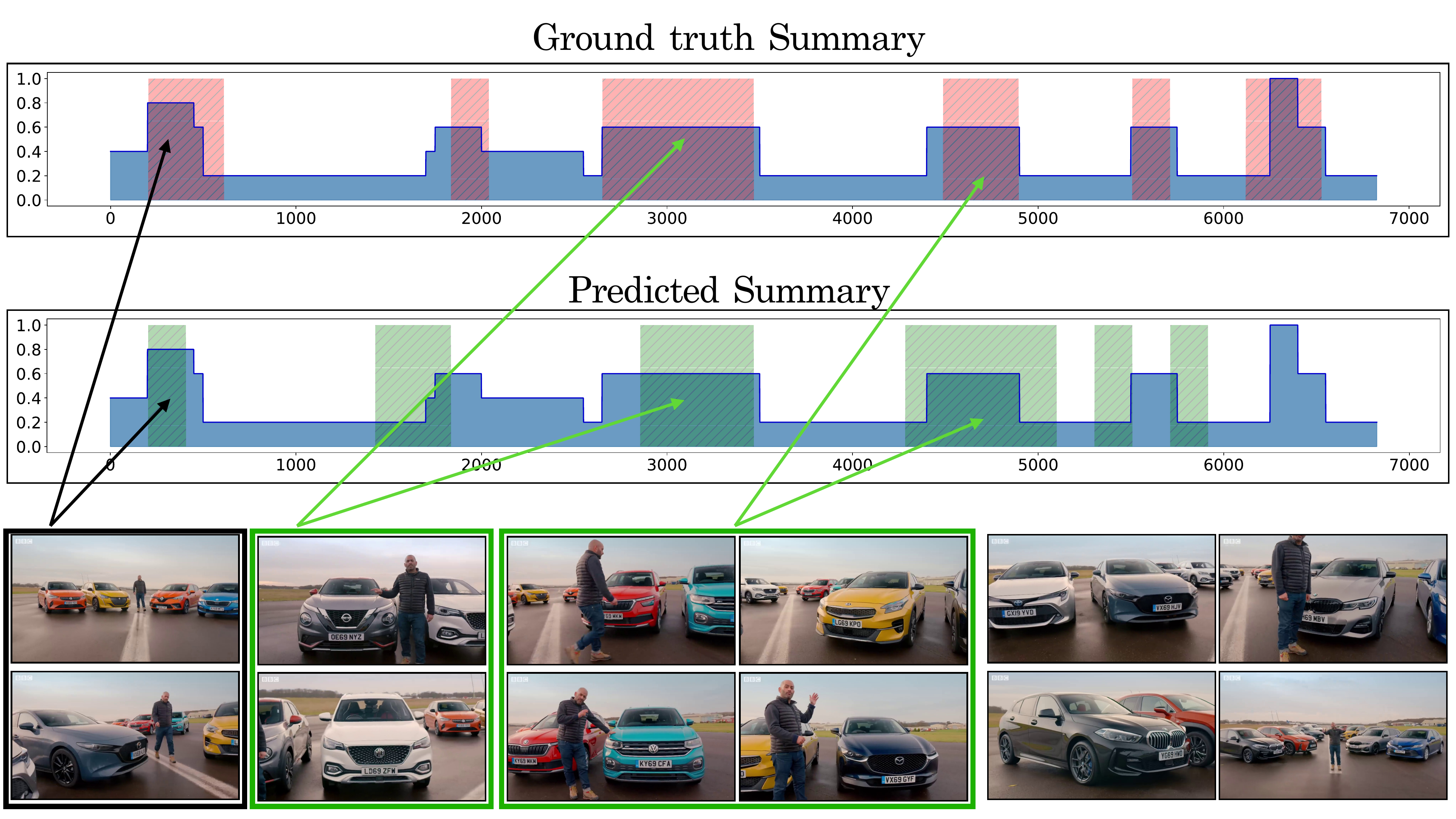}
        \caption{\textbf{Visualization of query-guided video summarization.} 
        The top plot shows ground-truth importance scores (blue) with user summary in red boxes; the bottom plot shows predicted scores with summaries in green boxes. 
        Sample frames illustrate alignment with the query ``Highlight crossover vehicles.'' 
        Frames common to both summaries are highlighted in green; key events in black.}
        \label{fig:TGVS_GT_MACHINE_QUALITATIVE}
    \end{minipage}
\end{figure*}

\paragraph{Quantitative results.} To assess the effectiveness of our fine-grained video summarization model, we conduct a quantitative comparison against a random baseline. As illustrated in Figure~\ref{fig:TGVS_categry_ours_vs_random}, we report the total F1 score (obtained by averaging over the splits accuracy) and the F1 scores across four different query categories defined in our ~\mydata{} dataset: Standard, Standard with Special Attribute, Reasoning, and Reasoning with General Knowledge. Notably, the performance gap is prominent in almost every query type, such as with Reasoning with General Knowledge, where understanding of temporal dependencies and commonsense knowledge is required.
Unfortunately, the accuracy of "Stranded with special attributive" query class wasn't as high as the rest, we suspect, this due to the difficulty level of the query, such as 'focus on bright colors' which might be hard to interpret. 
The random baseline was computed by averaging the F1-scores of 100 randomized tests per query category~\cite{zhang2016video}, ensuring a fair and statistically stable comparison.

On average, our model achieves a total F1-score of 43.4, while the random baseline yields 34.56, demonstrating the advantage of leveraging text queries for fine-grained summarization. These results validate our model’s ability to adapt to diverse query types and produce contextually relevant summaries that align with user intent.

\paragraph{Qualitative example.} To further illustrate the alignment between our model’s predictions and user intent, Figure~\ref{fig:TGVS_GT_MACHINE_QUALITATIVE} provides a detailed visual comparison for the query: “Highlight crossover vehicles in the video.” The top plot displays the ground-truth importance scores (in blue) along with the ground-truth summary, while the bottom plot overlays the predicted summary (green boxes) on the same importance scores. For additional insight, representative video frames are shown. Frames that appear in both summaries and align with the user’s query—focusing on crossover vehicles—are highlighted in green. Meanwhile, frames that correspond to a key event in the video—such as the introduction of the video’s goal (e.g., selecting the best car of the year)—and are also shared across both summaries are highlighted in black. This overlap indicates that the model not only captures query-specific segments but also preserves key content identified by human annotators.

\subsection{Ablation}\label{sec:ablation}
\noindent
We conducted ablation experiments to assess the contribution of each component: (i) image encoder for frame weighting (Section~\ref{sec:frame_weighting}), (ii) global vs. local visual inputs for description generation (Section~\ref{sec:desc}), (iii) 
hyperparameters (segment duration $W$ and frame consistency weight $\sigma$), and (iv) LLMs as scene-importance judges (Section~\ref{sec:scene_scoring}).

\begin{figure*}[t !]
    \centering
    \begin{subfigure}[t]{0.49\linewidth}
        \centering
        \includegraphics[width=\linewidth]{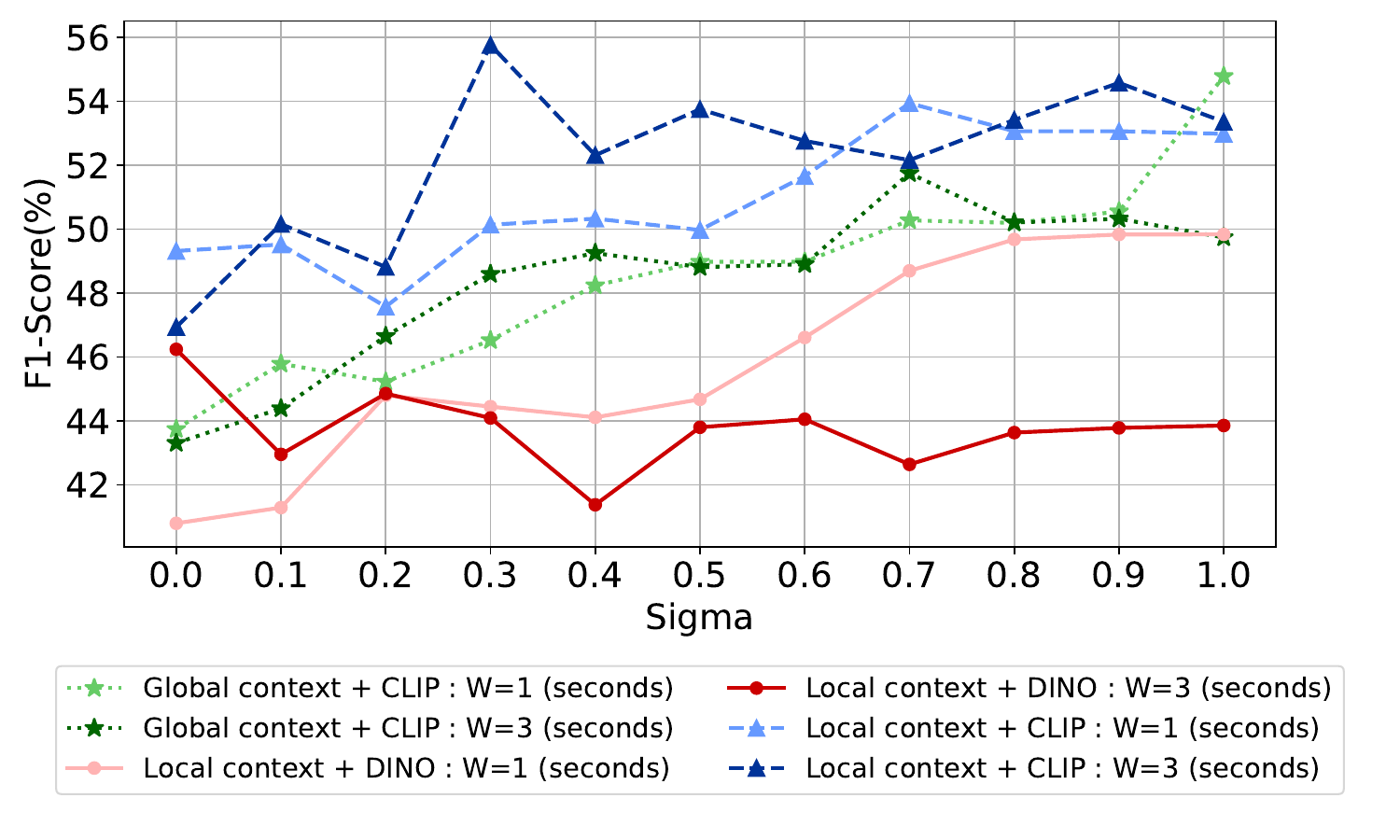}
        \caption{SumMe}
        \label{fig:sumMe_QUANTITATIVE}
    \end{subfigure}
    \hfill
    \begin{subfigure}[t]{0.49\linewidth}
        \centering
        \includegraphics[width=\linewidth]{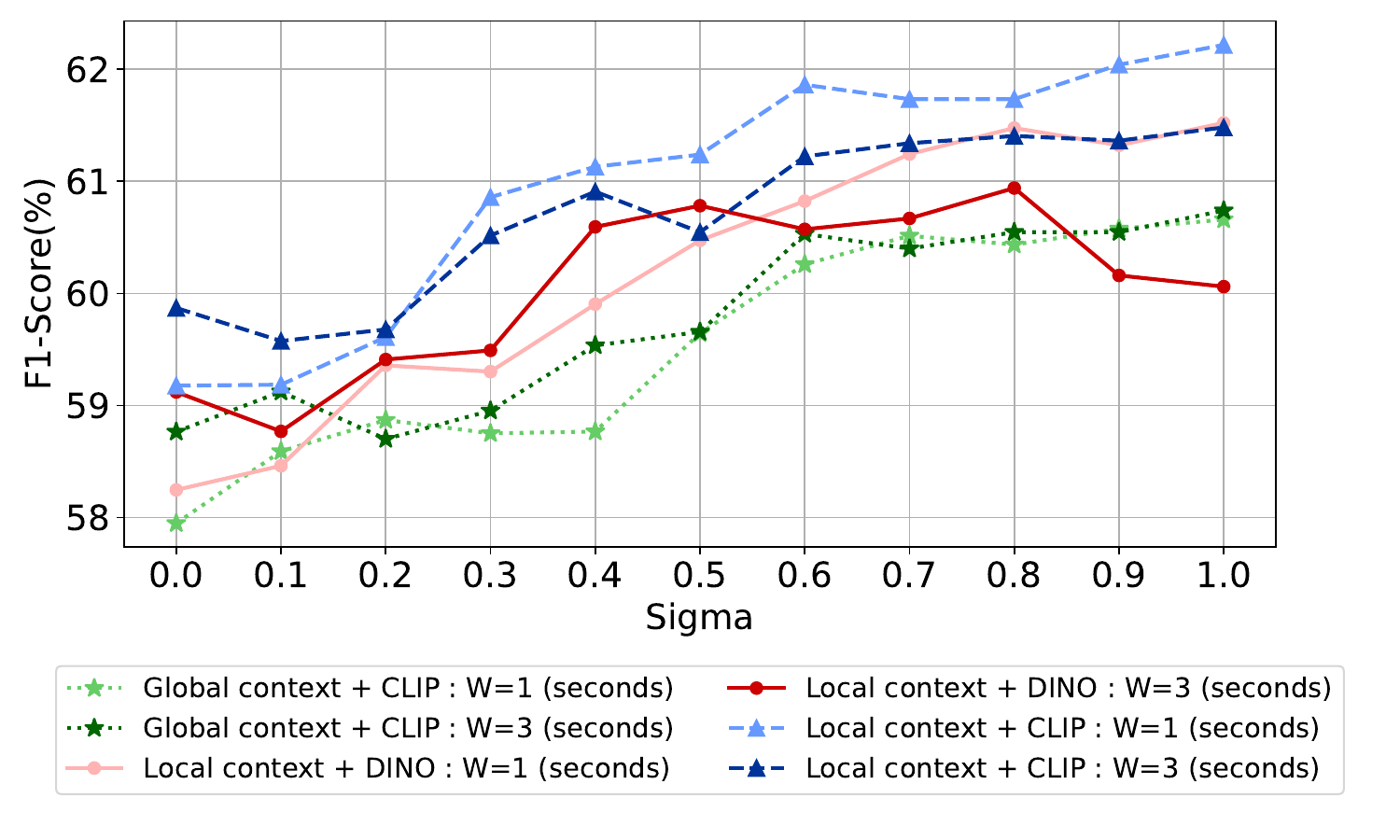}
        \caption{TVSum}
        \label{fig:TVSum_QUANTITATIVE}
    \end{subfigure}
    \caption{\textbf{Ablating different image encoders and visual context inputs for description generation.} F1-score results on the test sets of SumMe (MinMax normalization) and TVSum (Exponential normalization) across different segment durations and $\sigma$ values ($\sigma \in [0.0, 1.0]$).}
    \label{fig:QUANTITATIVE}
\end{figure*}

\paragraph{Image encoder.} The image encoder is a key component of our pipeline, as it directly influences how well frame weights are estimated within each scene and how accurately detected scene boundaries can be refined. To assess its impact on summarization quality, we compared the DINO model~\cite{caron2021emerging} with CLIP across different frame consistency weights ($\sigma$) and segment durations ($W$) on the TVSum and SumMe datasets. As shown in Figure~\ref{fig:QUANTITATIVE}, while some DINO variations achieved competitive results—reaching up to 61.5\% F1 on TVSum (see Figure ~\ref{fig:TVSum_QUANTITATIVE})—they consistently underperformed compared to CLIP. The CLIP encoder yielded the highest F1-scores across both datasets, highlighting its stronger semantic representation and robustness in identifying meaningful video segments.

\paragraph{Global vs. Local Visual Inputs for Description Generation.} Visual context plays a critical role in assessing scene importance: using global or local inputs can directly affect the granularity and accuracy of scene-level descriptions in our pipeline. To study this, we compared two approaches. In the \emph{global} approach, a target scene was masked by replacing its frames with black frames containing the text ``MASKED SCENE,'' while keeping the rest of the video intact. The masked video was then processed with ~\vlm{} using a modified prompt instructing the model to describe both the video and any masked segments. Similarly, LLMs were prompted to assign importance scores based on the degradation of the global narrative caused by the missing scene. In contrast, the \emph{local} approach relied solely on the frames of the scene in question, ignoring broader temporal context.  

Contrary to expectations, the local-only input consistently outperformed the global masked method on both TVSum and SumMe (Figure~\ref{fig:sumMe_QUANTITATIVE} and Figure~\ref{fig:TVSum_QUANTITATIVE}). On SumMe, the global approach reached a competitive 54.76\% F1 score, likely due to abstract scenes (e.g., burning matchsticks video) that are difficult to assess locally. Nevertheless, local descriptions were generally more accurate, focused, and relevant, while global masking sometimes introduced missing or misleading information, leading to noisier importance estimates. These results indicate that, for description-driven video summarization, local visual input provides stronger grounding for semantically rich and precise scene-level assessments. For further analysis of our pipeline's performance on \emph{abstract} and \emph{concrete} videos, see Section~\ref{sec:appendix_experiments} in the appendix (specifically Table~\ref{tab:abstract_concrete}).

\begin{figure}[t !]
    \centering
    \begin{subfigure}[t]{0.49\linewidth}
        \centering
        \includegraphics[width=1.0\linewidth]{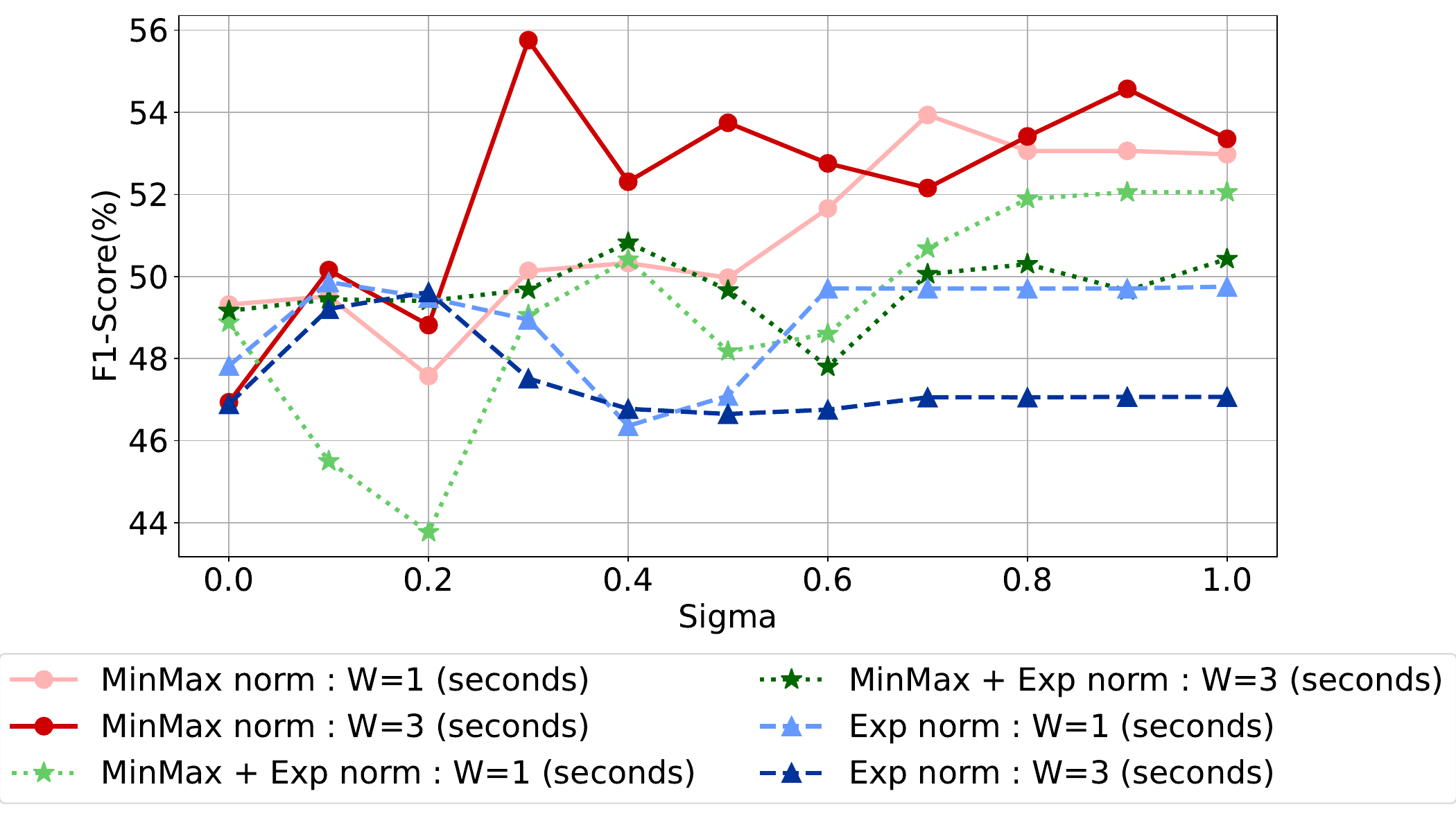}
        \caption{SumMe}
        \label{fig:sumMe_QUANTITATIVE_norm}
    \end{subfigure}
    \hfill
    \begin{subfigure}[t]{0.49\linewidth}
        \centering
        \includegraphics[width=1.0\linewidth]{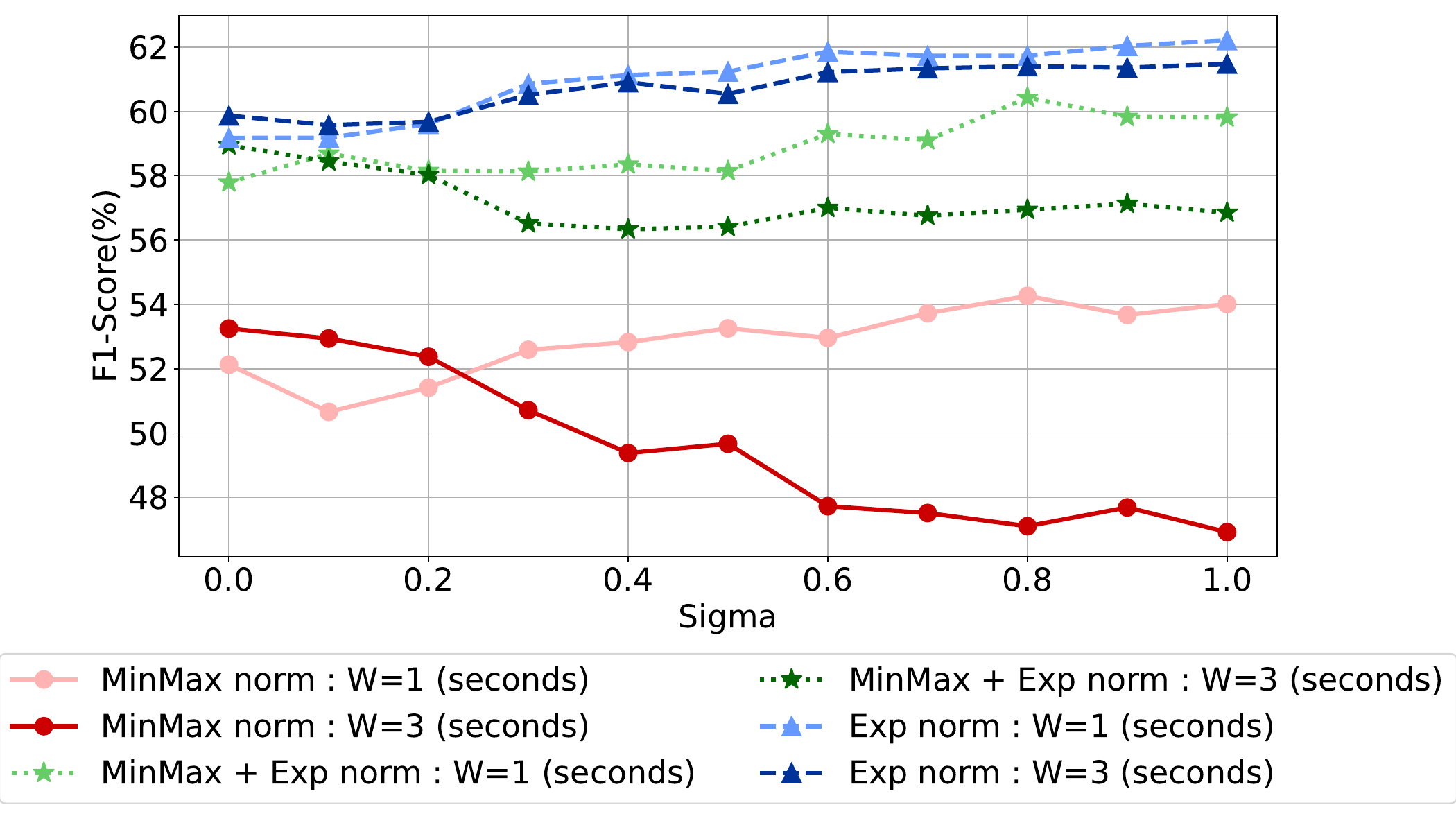}
        \caption{TVSum}
        \label{fig:TVSum_QUANTITATIVE_norm}
    \end{subfigure}
    \caption{\textbf{Ablating different normalization strategies and segment durations.} F1-score results on the test sets of SumMe (MinMax normalization) and TVSum (Exponential normalization) across different segment durations and $\sigma$ values ($\sigma \in [0.0, 1.0]$).}
    \label{fig:QUANTITATIVE_norm}
\end{figure}

\paragraph{Normalization Function.} Proper normalization of frame-level importance scores is essential to balance uniqueness and consistency across segments, directly affecting the quality of the final summary. To evaluate its impact, we tested three normalization strategies—MinMax, Exponential (Exp), and a combined MinMax+Exp—across different frame consistency weights ($\sigma \in [0.0,1.0]$) and segment durations ($W \in \{1,3\}$ seconds) on SumMe (Figure~\ref{fig:sumMe_QUANTITATIVE_norm}) and TVSum (Figure~\ref{fig:TVSum_QUANTITATIVE_norm}).

MinMax normalization consistently outperformed the other methods across both segment durations. Shorter segments benefited from higher $\sigma$ values, emphasizing consistency, while longer segments performed better with lower $\sigma$, emphasizing uniqueness. On TVSum, Exp normalization achieved relatively higher performance, though differences across segment durations were smaller. The combined MinMax+Exp approach yielded competitive results on both datasets, reaching up to 52\% F1 on SumMe and 60\% on TVSum, suggesting that integrating multiple normalization strategies can further enhance summarization performance.

\begin{figure}[t!]
    \centering
    \begin{subfigure}[t]{0.49\linewidth}
        \centering
        \includegraphics[width=1.0\linewidth]{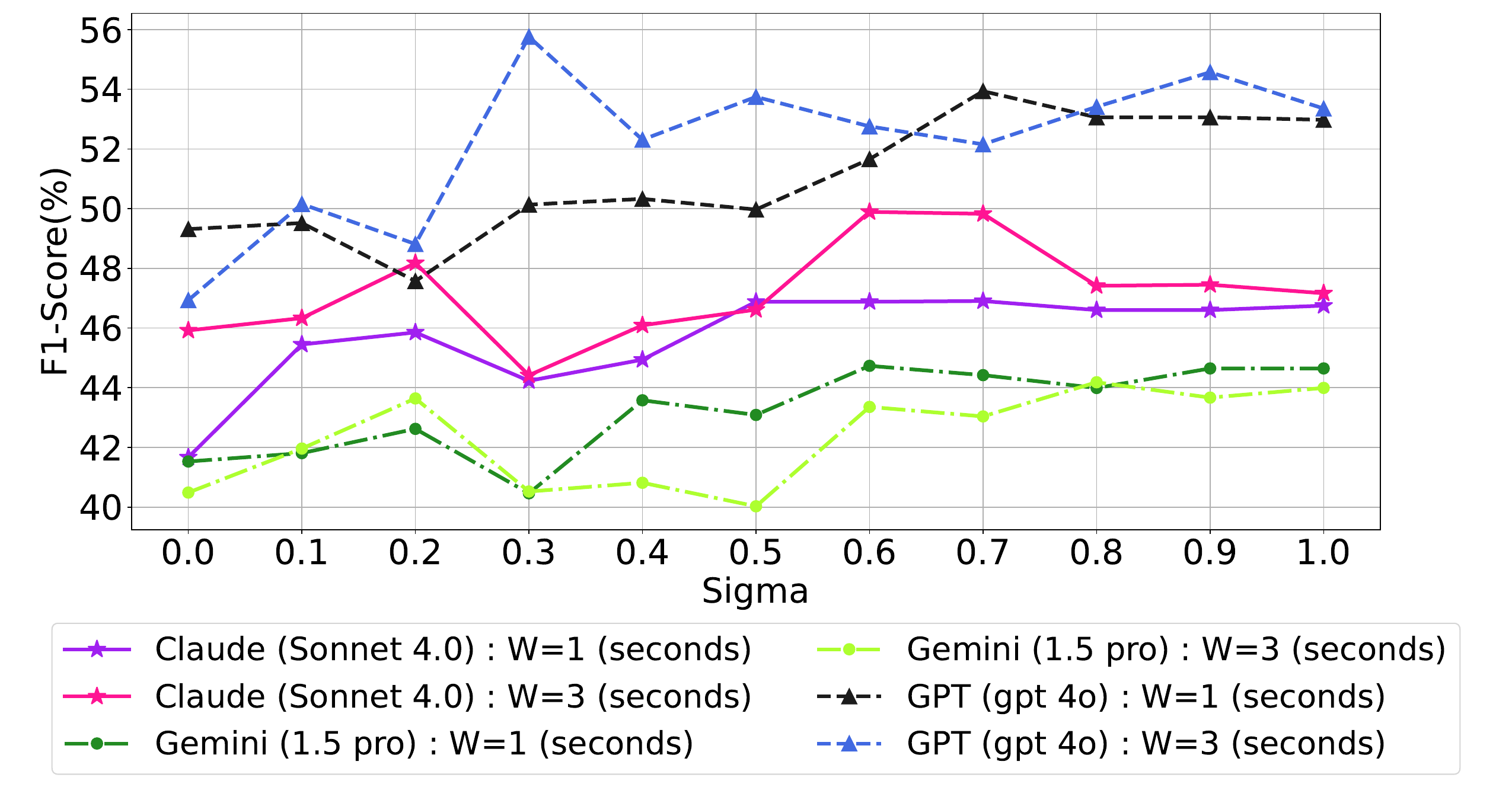}
        \caption{SumMe}
        \label{fig:sumMe_QUANTITATIVE_LLM}
    \end{subfigure}
    \hfill
    \begin{subfigure}[t]{0.49\linewidth}
        \centering
        \includegraphics[width=1.0\linewidth]{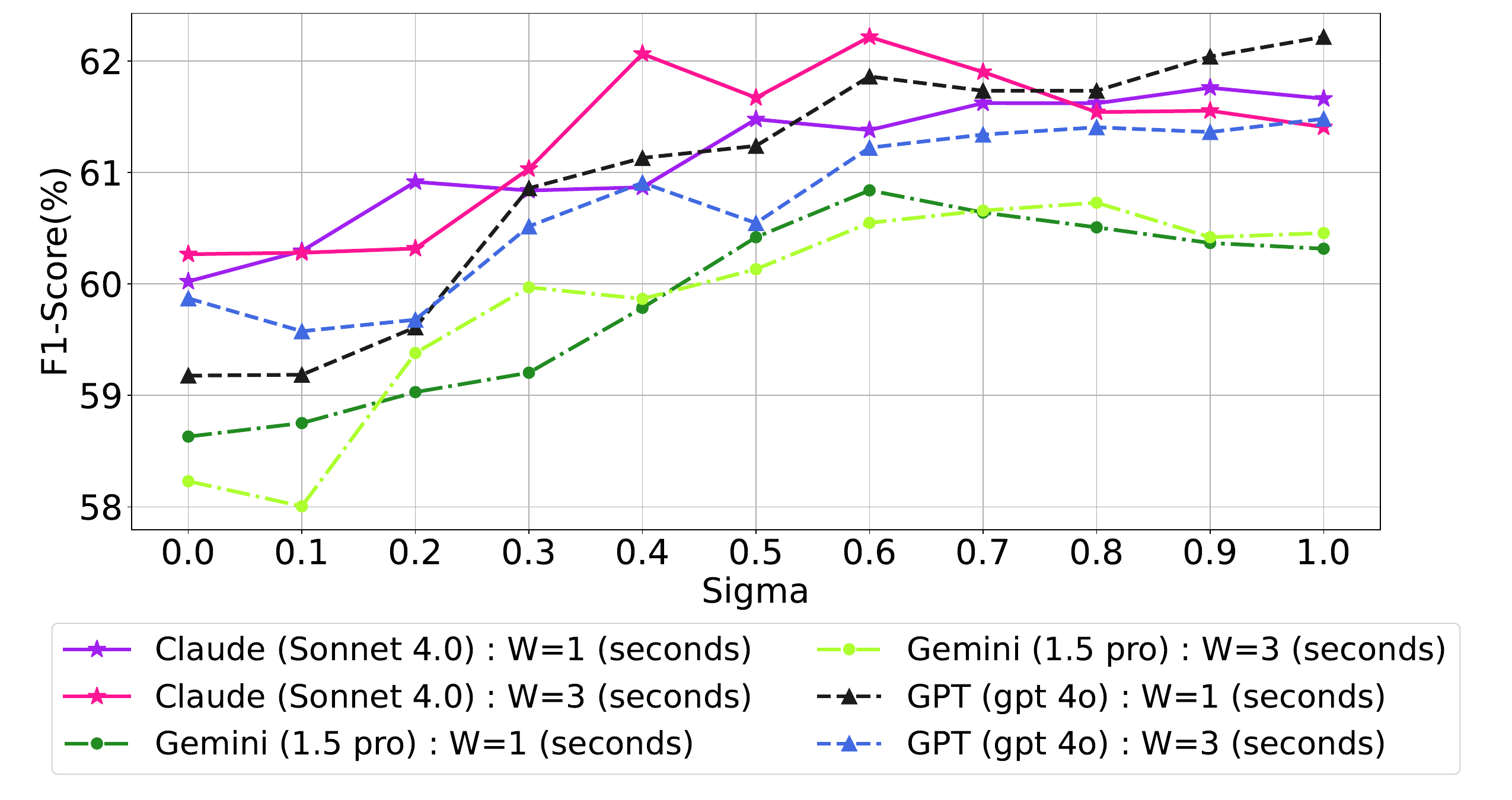}
        \caption{TVSum}
        \label{fig:TVSum_QUANTITATIVE_LLM}
    \end{subfigure}
    \caption{\textbf{Ablating different LLMs as scene-importance judges.} F1-score results on the test sets of SumMe (MinMax normalization) and TVSum (Exponential normalization) across different segment durations and $\sigma$ values ($\sigma \in [0.0, 1.0]$).}
    \label{fig:QUANTITATIVE_LLM}
\end{figure}

\paragraph{LLM as a judge.} Multiple Large Language Models (LLMs) were evaluated for their ability to assign importance scores to scenes based on textual descriptions. We tested GPT-4o~\cite{openai2024gpt4o}, Gemini-1.5 Pro~\cite{gemini15pro}, and Claude-Sonnet4~\cite{claudeSonnet4}, using different frame consistency weights ($\sigma$), segment durations ($W$), and normalization strategies (MinMax for SumMe, Exponential for TVSum). On the TVSum dataset (Figure~\ref{fig:TVSum_QUANTITATIVE_LLM}), all LLMs performed similarly, with some models, e.g., Claude-Sonnet4, approaching GPT-level performance (62.2\% F1-score). In contrast, results on SumMe (Figure~\ref{fig:sumMe_QUANTITATIVE_LLM}) revealed larger differences: non-GPT LLMs were more prone to false judgments, often assigning lower scores to scenes that should be rated highly, which reduced overall precision, highlighting the importance of selecting an LLM that reliably interprets textual descriptions for accurate importance scoring.

\subsection{Additional ablations}
Further high-resolution experiments and configuration validations are provided in Appendix~\ref{sec:appendix_experiments}. 
Prompt sensitivity analyses, including ablations over different prompt designs and their effect on LLM-based scene importance estimation, are presented in Section~\ref{sec:prompt_sensitivity}. 
Scalability experiments addressing ultra-long videos are reported in Section~\ref{sec:appendix_timing}. 
Additional analysis of our pipeline’s performance on \emph{abstract} versus \emph{concrete} videos is also included in Appendix~\ref{sec:appendix_experiments} (see Table~\ref{tab:abstract_concrete}).

\subsection{Implementation details}
All the experiments detailed in this paper were conducted on a PC equipped with one NVIDIA A100-SXM4-40GB GPU and AMD EPYC 7742 64-Core CPUs, additional implementation details are provided in Section ~\ref{sec:appendix_implementation} in the appendix.

\section{Conclusion}
This work introduces \emph{Prompts to Summaries}, the first {\bf zero-shot, text-queryable} video–summarization framework that orchestrates large video–language models (\vlm{}s) and large language models (LLMs) \emph{without any task-specific training data}. Extensive experiments, a new benchmark, and detailed ablations yield the following key findings: 

(i) \textbf{Training-free state of the art.}  
          Our plug-and-play pipeline surpasses all prior \emph{unsupervised} methods on SumMe and TVSum and rivals some \emph{supervised} systems—despite requiring no frame-level annotations, data, or domain tuning. 
          
          (ii) \textbf{~\vlm{} captions are “summary-ready.”}  
          Off-the-shelf \vlm{}s already produce rich, coherent scene descriptions that can double as draft summaries.  
          By simply ranking these captions with an LLM judge and propagating scores, we convert them into competitive, user-controllable video skims with no additional training.
          
          (iii)  \textbf{SOTA LLMs are great text-based “scene judges”}  
          With careful prompt design, off-the-shelf LLMs ranked the generated captions in a zero shot manner, leading to SOTA results.

(iv) \textbf{Effective query guidance.}  
          A single prompt extension lets users steer the summary with natural-language queries; on QFVS we match the fully-supervised CLIP-It while outperforming four classical baselines by up to 20 pp F\textsubscript{1}.

(v) \textbf{New fine-grained benchmark.}  
          We release ~\mydata{}, a new fine-grained text-guided video summarization benchmark whose queries demand temporal reasoning and external knowledge, opening a new testbed for more sophisticated video summarization; our method attains a mean 43.4 F\textsubscript{1}, beating a random baseline by nearly 10 pp.
          
(vi) \textbf{Local trumps global masking.}  
          For the video summarization task, measuring the importance of adding a scene to the summary performs better than computing the loss of excluding it from it, this may be explained as descriptions generated from the scene itself outperform those from globally masked scenes withing video videos, confirming the value of focused visual context.

(vii) \textbf{CLIP embeddings remain strongest.}  
          CLIP/ViT-L14 leads DINO variants by up to 2 pp F\textsubscript{1} across datasets in the context of frame embedding for summarization.

(vii) \textbf{Match normalization to annotation style.}  
          Min–Max works best for SumMe, exponential for TVSum; a hybrid scheme is robust for unseen domains.

(ix) \textbf{Scalable.}  
          Our batch split technique for generating video summaries enables end-to-end processing for a 10-minutes video in real-time and proved to be scalable and robust for a 5-hours video on a single A100 GPU, paving the way for practical deployment.

\section{Limitations and future work}
Our work suffers from the following two limitations: 
       \textbf{(i) Prompt sensitivity.}  
          Small wording changes in the LLM judge prompt can modulate scene scores by several points; a formal stability study is still missing. 
    \textbf{ (ii) Compute overhead for very long videos.}  
          Although processing is real-time for \textasciitilde10-minute clips, multi-hour footage still requires batched \vlm{} inference, and thus the run time is proportional to the number of baches we wish to generate a description for.

Future work induces: \textbf{(i) Audio–visual fusion.}  Integrate ASR transcripts and learned audio embeddings so that sound events and speech can influence scene scoring and user queries can reference audio content. \textbf{(ii) Faster summarization.}  Compress the entire stack to make it faster and enabling it on mobile device, enabling privacy-preserving, real-time highlights on wearable and drones.

\section*{Author contributions statement}
Mario Barbara developed the code, collected and curated the dataset, conducted the experiments, and performed the analyses. Alaa Maalouf contributed substantially to the development of the methodology, supervised the dataset collection and experiments, and provided guidance throughout the study. Both authors contributed to writing and revising the manuscript
All authors reviewed the manuscript.

\section*{Data availability statement}
Our new dataset \mydata{} can be found \href{https://github.com/mario998-hash/ZeroShotVideoSummary/tree/main/VidSum-Reason}{here} and \href{https://doi.org/10.21227/mxpw-gd11}{DOI}. We also encourage the reader to view our code at the \href{https://github.com/mario998-hash/ZeroShotVideoSummary.git}{Github page} and review our explanatory \href{https://youtu.be/cvoavENyB10}{video}.

\section*{Acknowledgment}
We acknowledge support from the Neubauer
Family Foundation and from the MAOF Fellowship of the Council for Higher Education.
We acknowledge the use of ~\cite{openai2024gpt4o} (GPT-4o) to assist with grammar correction, typo fixing, and rephrasing of some sentences for improved clarity and readability.

\newcommand{\JournalTitle}[1]{#1}

\clearpage
\appendix
\section{Additional implementation details}
%\addcontentsline{toc}{subsection}{Additional implementation details}
\label{sec:appendix_implementation_details}

In this section, we provide additional implementation details that were omitted from the main paper due to space constraints.

\definecolor{mygreen}{RGB}{0,102,0} % Option 1: RGB approximation
\subsection{LLM prompt}\label{sec:appendix_LLM_prompt}

\newenvironment{quote*}
  {\begin{quote}\setlength{\leftskip}{0pt} \setlength{\rightskip}{0pt}}
  {\end{quote}}
First we give the exact used LLM prompt used to generate scene scores:
\begin{quote*}
\label{prm:scene-importance}
\noindent\textbf{LLM prompt.}"You are tasked with evaluating the importance of a specific scene within a larger video, 
\textcolor{mygreen}{considering its role in the overall narrative and message of the video.} 

I've provided two descriptions below: one for the entire video and 
one for the specific scene (part) within that video. 
\textcolor{blue}{Your goal is to assess how critical this particular segment is to the
understanding or development of the video's main themes, messages, or emotional impact.}\\ \\
\textbf{If user query is available:}\\
\hspace*{2em}\texttt{The user has provided the following content preference to guide}\\ 
\hspace*{2em}\texttt{the summarization:} \\
\hspace*{2em}\texttt{User Query: <user query>} \\
\hspace*{2em}\texttt{When assigning a score, consider how well the scene aligns with }\\
\hspace*{2em}\texttt{this preference. Scenes that closely match or contradict the user’s }\\
\hspace*{2em}\texttt{intent should be scored accordingly, reflecting their relevance or} \\
\hspace*{2em}\texttt{irrelevance to the desired summary focus.} \\
\hspace*{2em}\texttt{If the scene is not clearly related to this preference, assign a }\\
\hspace*{2em}\texttt{score based on the default scale and criteria below.}\\ \\
Assign an importance score on a scale of 1 to 100, 
\textcolor{mygreen}{based on how crucial it is to the overall video.}
\textcolor{mygreen}{The scale is defined as follows:} \\
\textcolor{mygreen}{* 1-20: Minimally important} 
\\  \textcolor{blue}{\hspace{1em}(contributes very little to the overall theme or message)}\\ 
\textcolor{mygreen}{* 21-40: Somewhat important} 
\\ \textcolor{blue}{\hspace{1em}(offers limited context or details that support the main theme)}\\ 
\textcolor{mygreen}{* 41-60: Moderately important}
\\ \textcolor{blue}{\hspace{1em}(provides useful context or details that support the main theme)} \\
\textcolor{mygreen}{* 61-80: Quite important} 
\\ \textcolor{blue}{\hspace{1em}(adds significant context or detail that enhances understanding of the main theme)} \\
\textcolor{mygreen}{* 81-100: Highly important} 
\\ \textcolor{blue}{\hspace{1em}(crucial to understanding or conveying the main message of the video)}\\ \\
\textcolor{red}{    
When evaluating, focus on the core narrative or emotional impact of the video.
Only assign high scores (80+) to the segments that directly drive the main theme or message forward. Be critical and biased towards giving 
low scores to segments that do not add significant value to the overall narrative or theme. The distribution 
of high scores should be low and reserved for only the most crucial moments in the video. 
The video should be summarized briefly, so please evaluate whether the scene is critical to include i
n the summary of the video, based on its contribution to the core message.
Prioritize scenes that are essential for a concise summary and omit secondary or supporting moments unless they provide meaningful context.}\\
Video Description: $<$video description$>$\\
Scene Description: $<$scene description$>$"

\end{quote*}

\subsection{Our algorithms}

\textbf{Algorithm~\ref{alg:scene-level-scoring}} outlines the process for computing scene-level importance scores using a large language model (LLM). Each scene description is paired with the full video description and inserted into a prompt template. The resulting query (Section~\ref{sec:appendix_LLM_prompt}) is then evaluated by the LLM to estimate how critical that scene is to the overall video, as described in details in Section ~\ref{sec:scene_scoring}

\textbf{Algorithm~\ref{alg:scene-scores-smoothing}} describes the scene score smoothing procedure. The method first normalizes scene-level scores and assigns each frame within a scene the scene’s score. To ensure smooth transitions between consecutive scenes, the algorithm linearly interpolates frame scores over the transition region defined by the midpoints between scenes, using a cosine-based weighting function. This smoothing step helps reduce abrupt changes in frame-level importance and enhances temporal coherence, as discussed in Section~\ref{sec:scene_smoothing}.

\textbf{Algorithm~\ref{alg:frame-level-weights}} details the frame weighting procedure within each scene. The algorithm extracts frame embeddings and determines the optimal number of clusters using the Elbow method based on within-cluster sum of squares (WCSS). Frames are grouped into segments of fixed size $W$, and each segment is scored by computing its consistency and uniqueness measures, weighted by parameter $\sigma$. These segment-level scores are then assigned uniformly to the frames within each segment, providing refined frame-level weights that capture both local coherence and distinctiveness, as detailed in Section~\ref{sec:frame_weighting}.

% ======================================================
% Algorithm 1
% ======================================================
\begin{algorithm}[H]
\caption{Scene-Level Importance Scores (Section~\ref{sec:scene_scoring})}
\label{alg:scene-level-scoring}
\begin{algorithmic}[1]
\State \textbf{Input:} Scene descriptions $\mathcal{D}_{scenes}$, full video description $\mathcal{D}_{video}$, LLM prompt template $\mathcal{P}$
\State \textbf{Output:} Scene-level importance scores $\mathcal{S}$
\State Initialize scene importance scores list $\mathcal{S} \gets \emptyset$

\For{$d_i \in \mathcal{D}_{scenes}$}
    \State Construct query $q_i \gets \text{Format}(\mathcal{P}, d_i, \mathcal{D}_{video})$
    \State Send $q_i$ to LLM and receive scene score $s_i$
    \State Append $s_i$ to $\mathcal{S}$
\EndFor

\State \Return $\mathcal{S}$
\end{algorithmic}
\end{algorithm}

% ======================================================
% Algorithm 2
% ======================================================
\begin{algorithm}[H]
\caption{Scene Scores Smoothing (Section~\ref{sec:scene_smoothing})}
\label{alg:scene-scores-smoothing}
\begin{algorithmic}[1]
\State \textbf{Input:} Scene scores $\{s_1, \dots, s_N\}$, scene boundaries $\{\mathcal{B}_1, \dots, \mathcal{B}_N\}$, normalization $F_{norm}$
\State \textbf{Output:} Frame-level scores $\{f_1, \dots, f_T\}$
\State Initialize scene scores $\mathcal{S} \gets F_{norm}(S)$

\For{$i = 1, \dots, N$}
    \State Get scene score $s_i$
    \State Get scene boundaries $\mathcal{B}_i = [start_i, end_i]$
    \For{$t \in \mathcal{B}_i$}
        \State Assign $f_t = s_i$
    \EndFor
\EndFor

\For{$i = 1, \dots, N-1$}
    \State $mid_i = \frac{start_i + end_i}{2}, \quad mid_{i+1} = \frac{start_{i+1} + end_{i+1}}{2}$
    \State $L = mid_{i+1} - mid_i$
    \For{$t = mid_i, \dots, mid_{i+1}$}
        \State $p = \frac{t - mid_i}{L}$
        \State $w = \frac{1 - \cos(\pi \cdot p)}{2}$
        \State $f_t = (1 - w) \cdot s_i + w \cdot s_{i+1}$
    \EndFor
\EndFor

\State \Return $\{f_1, \dots, f_T\}$
\end{algorithmic}
\end{algorithm}

% ======================================================
% Algorithm 3
% ======================================================
\begin{algorithm}[H]
\caption{Frame Weighting (Section~\ref{sec:frame_weighting})}
\label{alg:frame-level-weights}
\begin{algorithmic}[1]
\State \textbf{Input:} Frames $\{f_1, \dots, f_T\}$, scene boundaries $\{\mathcal{B}_1, \dots, \mathcal{B}_N\}$, segment size $W$, parameter $\sigma$
\State \textbf{Output:} Frame weights $\{w_1, \dots, w_T\}$

\For{$i = 1, \dots, N$}
    \State $E_{\mathcal{B}_i} = \{e_t \mid t \in \mathcal{B}_i\}$
    \State $K^* = \arg\min_{K_i \in [K_{\min}, K_{\max}]} WCSS(E_{\mathcal{B}_i}, K_i)$
    \State Cluster $E_{\mathcal{B}_i}$ into $K^*$ clusters
    \For{$s \in \mathcal{B}_i$}
        \State $c_s = \text{Consistency}(s)$
        \State $u_s = \text{Uniqueness}(s)$
        \State $w_s = \sigma \cdot c_s + (1 - \sigma) \cdot u_s$
        \For{$f_t \in s$}
            \State $w_t = w_s$
        \EndFor
    \EndFor
\EndFor

\State \Return $\{w_1, \dots, w_T\}$
\end{algorithmic}
\end{algorithm}

\begin{algorithm}[H]
\caption{Frame-Level Scoring}
\label{alg:frame-level-scoring}
\begin{algorithmic}[1]
\State \textbf{Input:} Video frames $F_{\text{Video}} = \{f_1, \dots, f_T\}$; \textbf{Optional} user query $Q_{\text{user}}$
\State \textbf{Output:} Frame-level importance scores
\State Initialize $\tilde{\mathcal{S}} \gets \emptyset$

\State // Detect scene boundaries (Section~\ref{sec:scene_detection})
\State $\mathcal{B} \gets \text{Scene\_Boundary\_Detection}(F_{\text{Video}})$

\State // Generate descriptions (Section~\ref{sec:desc})
\State $Scenes_{\text{Desc}}, Video_{\text{Desc}} \gets \text{GenDesc}(\mathcal{B}, F_{\text{Video}})$

\State // Score scenes using LLM (Section~\ref{sec:scene_scoring})
\State $\mathcal{S} \gets \text{Scene\_Scores}(Scenes_{\text{Desc}}, Video_{\text{Desc}}, Q_{\text{user}})$

\State // Normalize scene scores and smooth frame-level scores (Section~\ref{sec:scene_smoothing})
\State $F_{\text{smooth}} \gets \text{Scene\_Scores\_Smoothing}(\mathcal{S}, \mathcal{B}, F_{\text{norm}})$

\State // Dynamically select segment duration and frame consistency weight
\State $W, \sigma \gets F_s(F_{\text{Video}})$

\State // Compute frame weights (Section~\ref{sec:frame_weighting})
\State $F_{\text{weights}} \gets \text{Frame\_Weighting}(F_{\text{Video}}, \mathcal{B}, W, \sigma)$

\For{$i = 1, \dots, N$}
    \State Get scene boundaries $\mathcal{B}_i = [\text{start}_i, \text{end}_i]$
    \State $\tilde{\mathcal{S}}_i \gets \emptyset$
    \For{$t \in \mathcal{B}_i$}
        \State // Compute final frame score
        \State $\tilde{f_t} \gets F_{\text{smooth}}[t] \times F_{\text{weights}}[t]$
        \State Append $\tilde{f_t}$ to $\tilde{\mathcal{S}}_i$
    \EndFor
    \State Normalize within scene: $\tilde{\mathcal{S}}_i \gets F_{\text{norm}}(\tilde{\mathcal{S}}_i)$
    \State Append $\tilde{\mathcal{S}}_i$ to $\tilde{\mathcal{S}}$
\EndFor

\State \Return Frame-level importance scores $\tilde{\mathcal{S}}$
\end{algorithmic}
\end{algorithm}

\subsection{Scene detection full flow illustration and example}
\begin{figure}[t!]
    \centering
    % --- Top-left ---
    \begin{subfigure}[t]{0.48\linewidth}
        \centering
        \includegraphics[width=\linewidth]{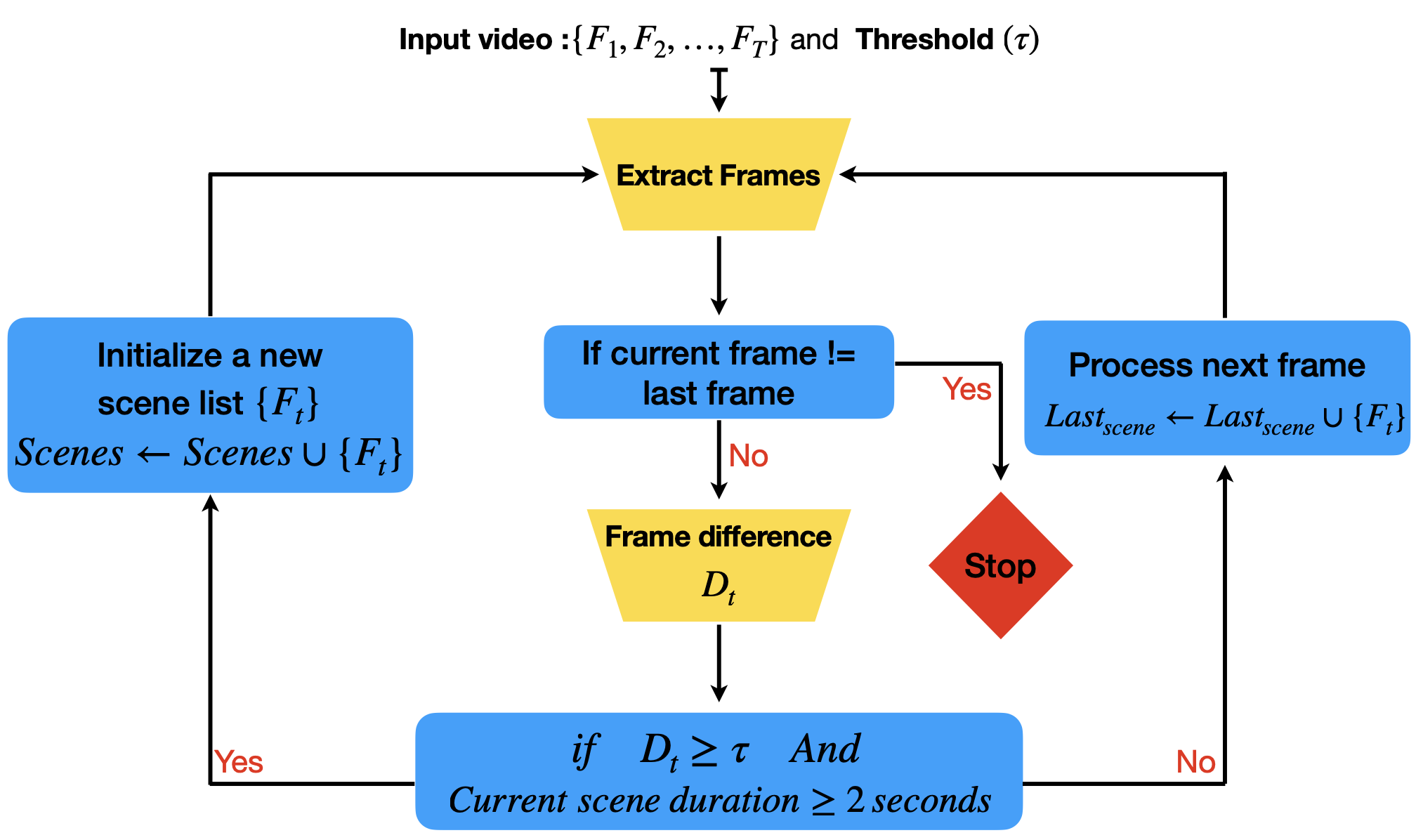}
        \caption[Scene boundary detection flow diagram]{Scene boundary detection flow diagram.}
        \label{fig:scene_detection_flow_diagram}
    \end{subfigure}
    \hfill
    % --- Top-right ---
    \begin{subfigure}[t]{0.48\linewidth}
        \centering
        \includegraphics[width=\linewidth]{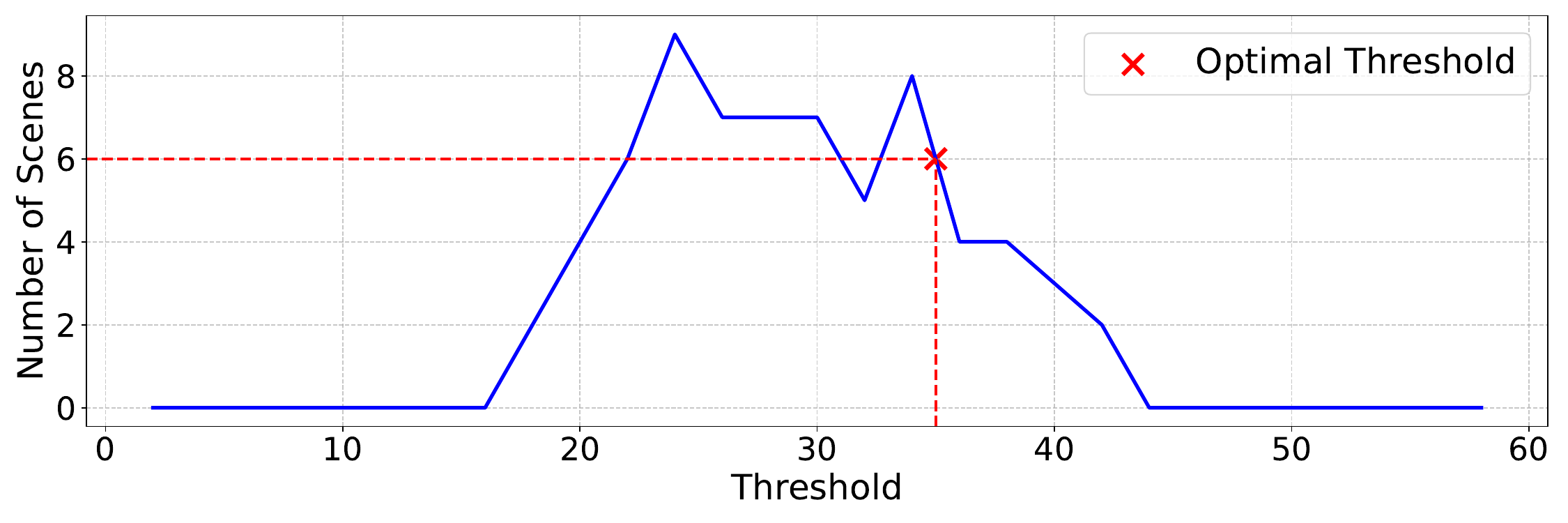}
        \caption[Determining the optimal threshold]{Determining the optimal threshold.}
        \label{fig:Bearpark_climbing_DynamicThresholdSelection}
    \end{subfigure}

    \vspace{0.5em}

    % --- Bottom-left ---
    \begin{subfigure}[t]{0.48\linewidth}
        \centering
        \includegraphics[width=\linewidth]{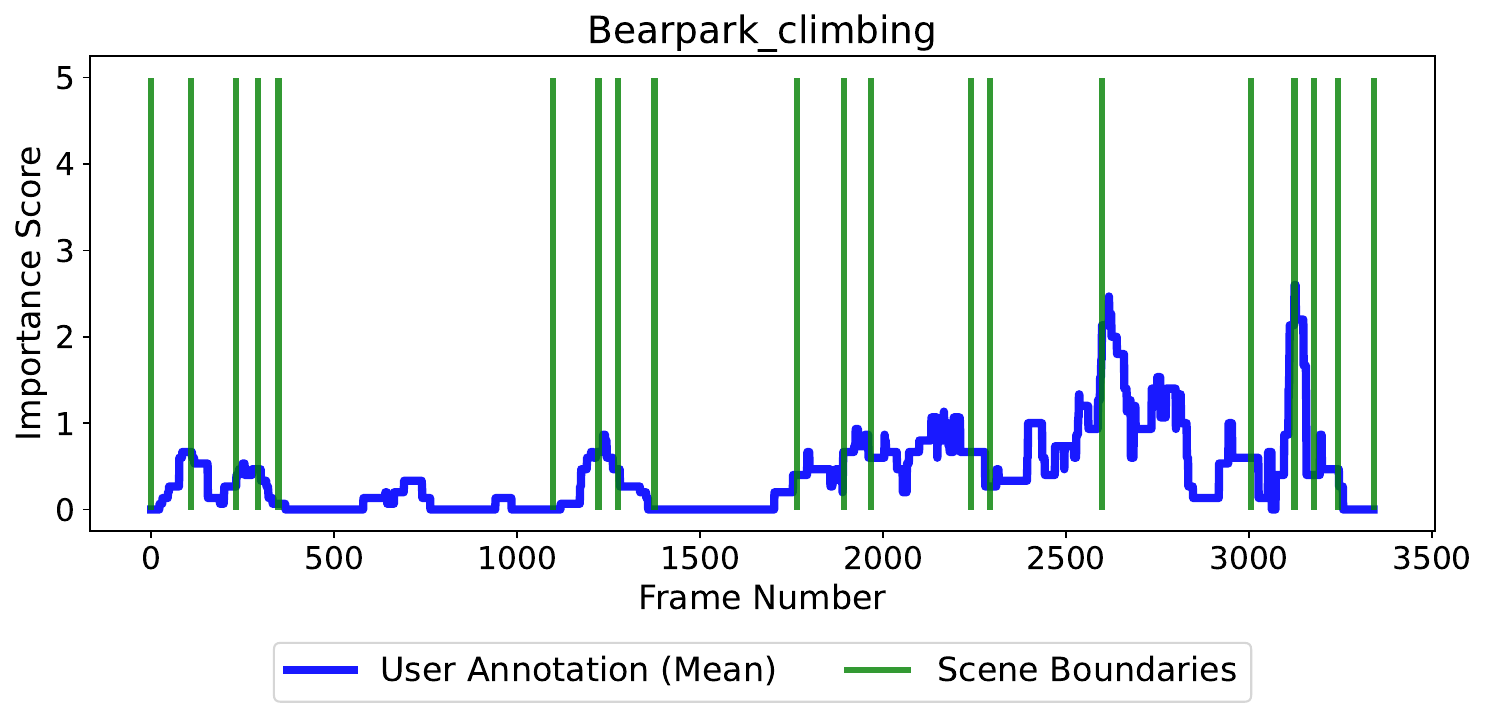}
        \caption[Initial detected scenes based on frame intensity differences]{Initial detected scenes based on frame intensity differences.}
        \label{fig:Bearpark_climbing_detected_scene_b4_ref}
    \end{subfigure}
    \hfill
    % --- Bottom-right ---
    \begin{subfigure}[t]{0.48\linewidth}
        \centering
        \includegraphics[width=\linewidth]{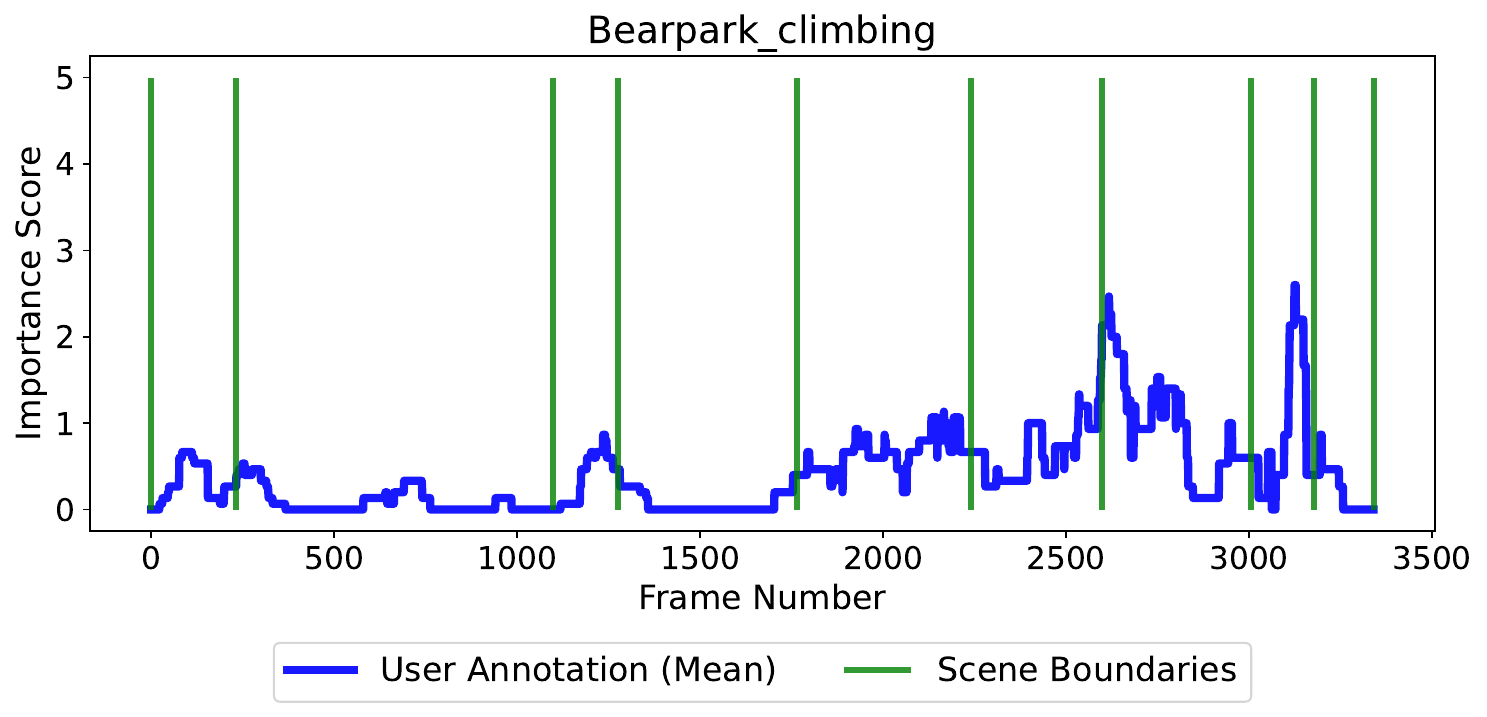}
        \caption[Refined scene boundaries after merging short scenes using cosine similarity over mean scene embeddings]{Refined scene boundaries after merging short scenes using cosine similarity over mean scene embeddings.}
        \label{fig:Bearpark_climbing_detected_scene_a4_ref}
    \end{subfigure}

    \caption[Scene boundary detection and pipeline example]{\textbf{Scene boundary detection and pipeline example for the \textit{Bearpark\_climbing} video (SumMe dataset).} 
    (a) Scene boundary flow diagram. (b) Threshold determination. (c) Initial detection. (d) Refined boundaries.}
    \label{fig:scene_bearpark_pipeline_grid}
\end{figure}

Figure~\ref{fig:scene_bearpark_pipeline_grid} presents an additional pipeline flow example for the \textit{Bearpark\_climbing} video from the SumMe dataset . This example complements the general pipeline in Figure~\ref{fig:general_pipeline_example} in the main paper, and illustrates how the first stages—Selecting optimal threshold, Initial scene detection, and refined detected scene boundaries—is applied in practice. Figure~\ref{fig:scene_detection_flow_diagram} presents the flow procedure for partitioning the input set of frames into scenes based on a given threshold $\tau$.

% ======================================================
% Scene Boundary + Bearpark Pipeline Figures in 2x2 grid
% ======================================================

\subsection{Dataset Details}\label{sec:appendix_datasets}
The \textit{SumMe} dataset comprises 25 videos (1–6 minutes each), covering diverse content such as holidays, events, and sports from both first-person and third-person perspectives. Each video was annotated by 15–18 users who provided key-fragment summaries, resulting in multiple user-generated summaries typically ranging from 5% to 15% of the video duration.

The \textit{TVSum} dataset includes 50 YouTube videos grouped into 10 categories (e.g., changing a tire, making a sandwich, visiting an award-winning shop), with 5 videos per category. Each video was annotated by 20 users who rated frame-level importance on a 1–5 scale~\cite{vo2025integrate}.

The \textit{QFVS} dataset~\cite{sharghi2017query} is built upon the UT Egocentric (UTE) dataset~\cite{lee2012discovering}, consisting of four egocentric videos ranging from 3 to 5 hours.
Each video is segmented into non-overlapping 5-second shots and annotated by three users with textual tags describing the visible content (e.g., \textit{CAR}, \textit{ROAD}, \textit{SKY}, \textit{STREET}).
Each video is also associated with 45–46 user queries and corresponding user-annotated summaries.
The user captions were aggregated per video-query into dense annotations, from which oracle (query-conditioned) summaries were derived using a greedy selection algorithm.
These oracle summaries yield higher inter-user agreement compared to individual annotations.

\subsection{Implementation details}\label{sec:appendix_implementation}
We employ frame value density with a dynamic threshold (Section~\ref{sec:init_scene_detect}) for scene detection, using the \textbf{contentDetector} function from PySceneDetect~\cite{castellano2024pyscenedetect} to identify scene boundaries. Frame embeddings are extracted using the CLIP/ViT-Large-Patch14 model~\cite{radford2021learning}, capturing both visual and contextual information effectively. For description generation, we use the LLAVA/Qwen-2.0-7B Video Language Model~\cite{bai2023qwen, liu2024llavanext} to produce textual representations at both the video and scene levels. \\

\textbf{Memory management details.} 
Experiments were conducted using approximately 12~GB of GPU memory. 
Sampled frames were divided into batches of 80 frames each. 
At a sampling rate of 1~FPS for full videos and 2~FPS for individual scenes, each batch corresponds to roughly 80~seconds and 40~seconds of video content, respectively. 
This configuration ensures detailed and temporally coherent descriptions while keeping memory usage within limits.\\

\textbf{Optimizations for Longer Videos.} For datasets like QFVS, where videos can exceed 5 hours with over 220k frames and multiple queries per video, processing each video–query pair individually is computationally prohibitive. To reduce overhead, we process all queries simultaneously for each scene by passing them together to the LLM, which assigns relevance scores for all queries in a single pass. This strategy can reduce processing time and resource usage by up to \textbf{50×}.

\begin{figure*}[h!]
    \centering

    \begin{subfigure}[t]{0.49\linewidth}
        \centering
        \includegraphics[width=\linewidth]{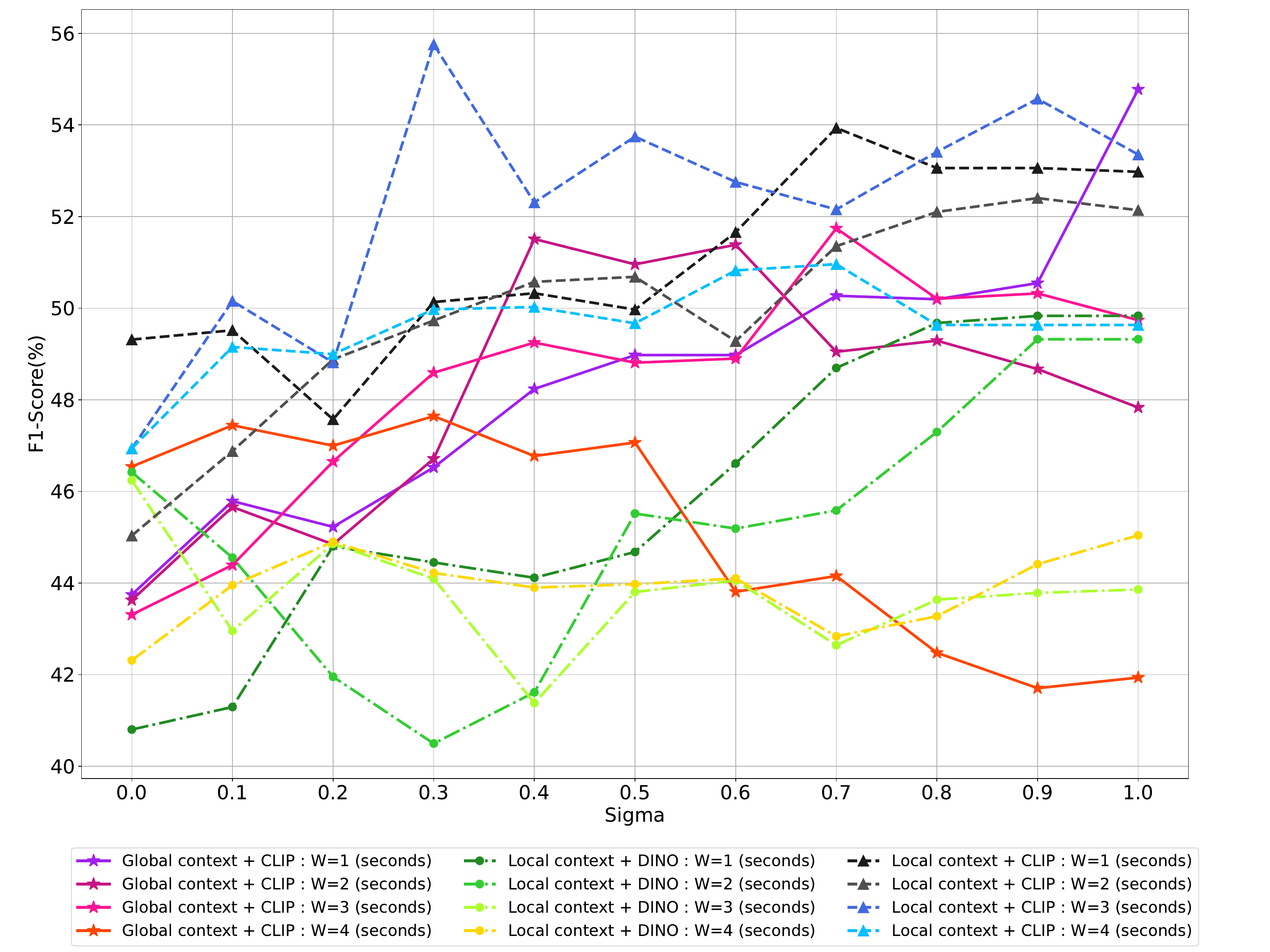}
        \caption{SumMe}
        \label{fig:Exper_ablation_sumMe}
    \end{subfigure}
    \hfill
    \begin{subfigure}[t]{0.49\linewidth}
        \centering
        \includegraphics[width=\linewidth]{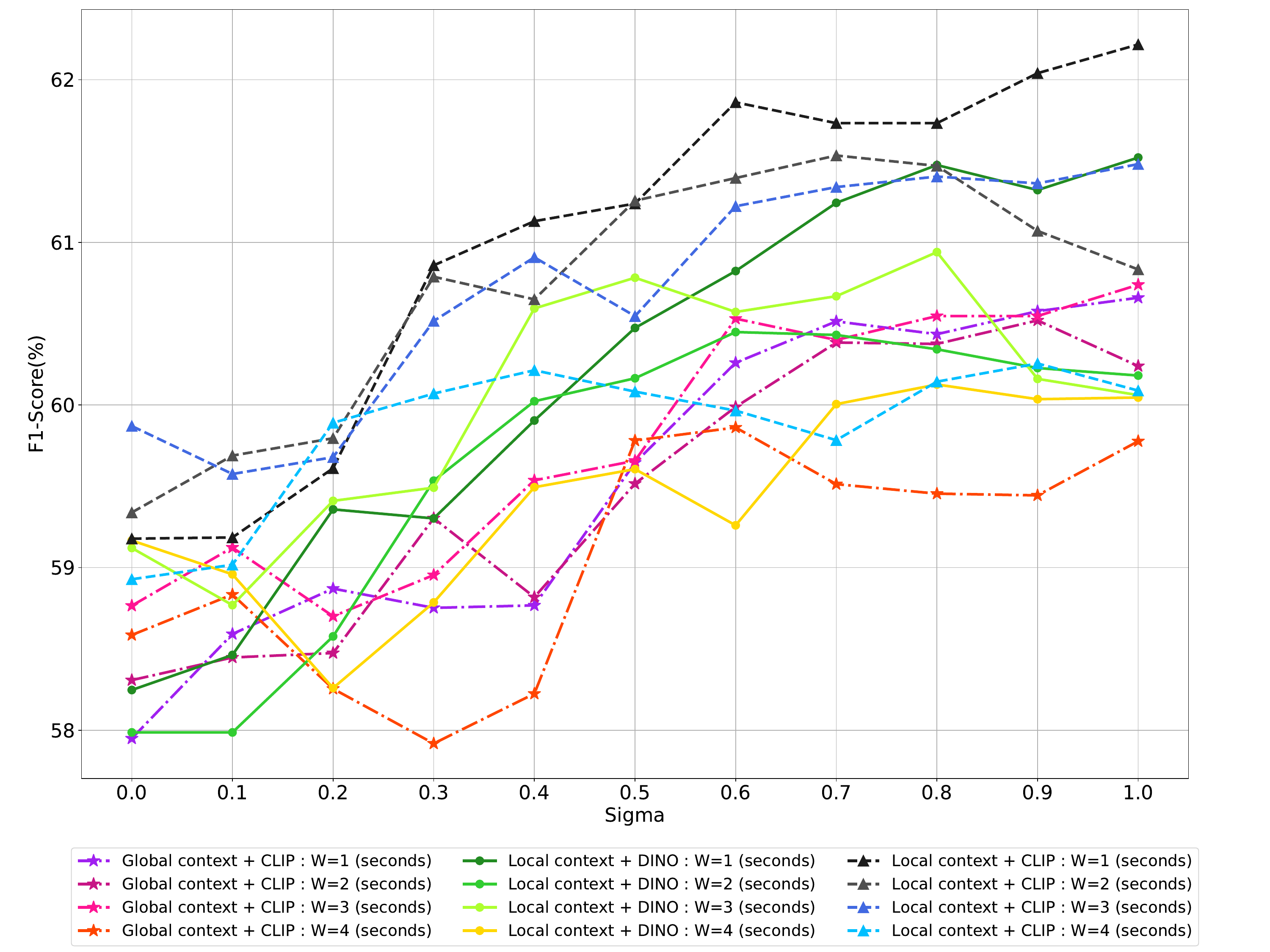}
        \caption{TVSum}
        \label{fig:Exper_ablation_tvSum}
    \end{subfigure}
    \caption[Ablating different Image encoders and Visual context input for description generation]{\textbf{Ablating different Image encoders and Visual context input for description generation.} F1 scores results on the test set of SumMe dataset employing MinMax normalization and TVSum dataset employing Exponential (Exp) normalization across more diverse segment durations and $\sigma$ values ($\sigma \in [0.0, 1.0]$).}
    \label{fig:Exper_ablation}
\end{figure*}

\section{Additional experiments and results}\label{sec:appendix_experiments}
%\addcontentsline{toc}{subsection}{Additional experiments and results}
\subsection{Broader experiments}
We conducted broader experiments to further validate our design choices.
In Figures~\ref{fig:Exper_ablation_sumMe} and~\ref{fig:Exper_ablation_tvSum}, we evaluated a more diverse set of segment durations compared to the experiments in Section ~\ref{sec:ablation}(main paper). Specifically, we tested segment durations $W=[1,2,3,4]$, using different image encoders (DINO and CLIP) and varying the type of visual context input for description generation (Local or Global context).
As shown, the previously identified best configurations (see Section 4.4 (main paper)) remain effective across both datasets. TVSum achieves the best performance with short segment durations and high frame weight values, while SumMe performs best with intermediate segment durations and moderate frame weight values. Notably, as previously mentioned, SumMe also performs competitively under the same configuration that works best for TVSum.\\ \\

\begin{table}[t]
    \centering
    \caption[Impact of video abstraction on summarization performance]{\textbf{Impact of video abstraction on summarization performance.}
    Videos are categorized as \emph{Concrete} or \emph{Abstract} based on the temporal localization and visual saliency of key events.}
    \label{tab:abstract_concrete}

    \begin{tabular}{l c c c}
    \hline
    \textbf{Dataset} & \textbf{Category} & \textbf{\#Videos} & \textbf{Mean F1-score (\%)} \\
    \hline
    SumMe & Concrete & 11/25 & 49.0 \\
          & Abstract & 14/25 & 50.8 \\ 
    \hline
    TVSum & Concrete & 36/50 & 57.5 \\
          & Abstract & 14/50 & 61.2 \\ 
    \hline
    \end{tabular}
\end{table}

\textbf{Video abstraction impact.}
We evaluated videos categorized as \emph{Concrete} or \emph{Abstract} based on the temporal localization of key events and their visual saliency, to assess whether our summarization pipeline is affected by the level of visual concreteness. As shown in Table~\ref{tab:abstract_concrete}, the mean F1-score is comparable—or slightly higher ($\Delta F_1 < 4\%$)—for Abstract videos, indicating that our pipeline is robust even when summarizing videos with more abstract or loosely defined events.\\

\begin{figure*}[h]
    \centering

    \begin{subfigure}[t]{0.49\linewidth}
        \centering
        \includegraphics[width=\linewidth]{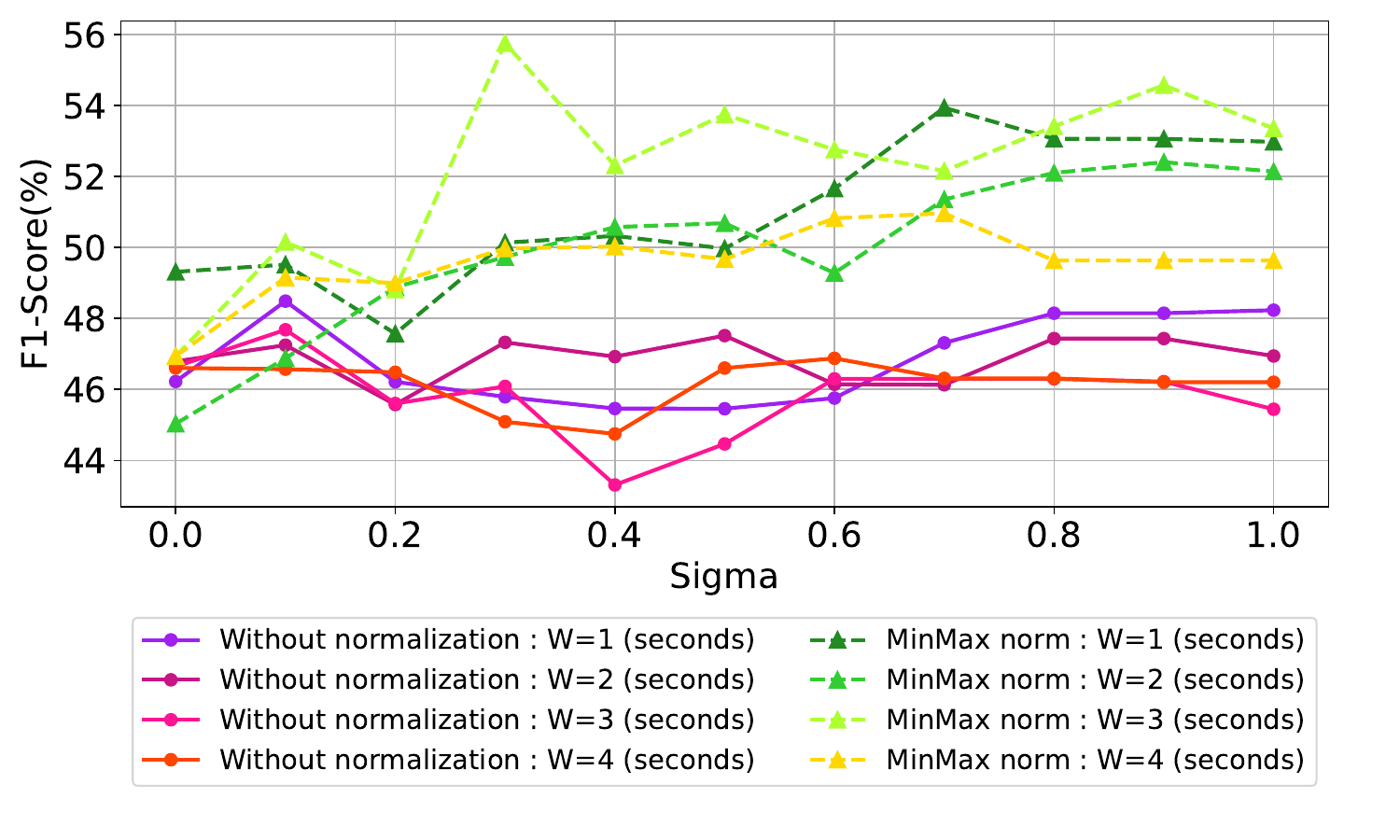}
        \caption{SumMe}
        \label{fig:Exper_ablation_sumMe_norm}
    \end{subfigure}
    \hfill
    \begin{subfigure}[t]{0.49\linewidth}
        \centering
        \includegraphics[width=\linewidth]{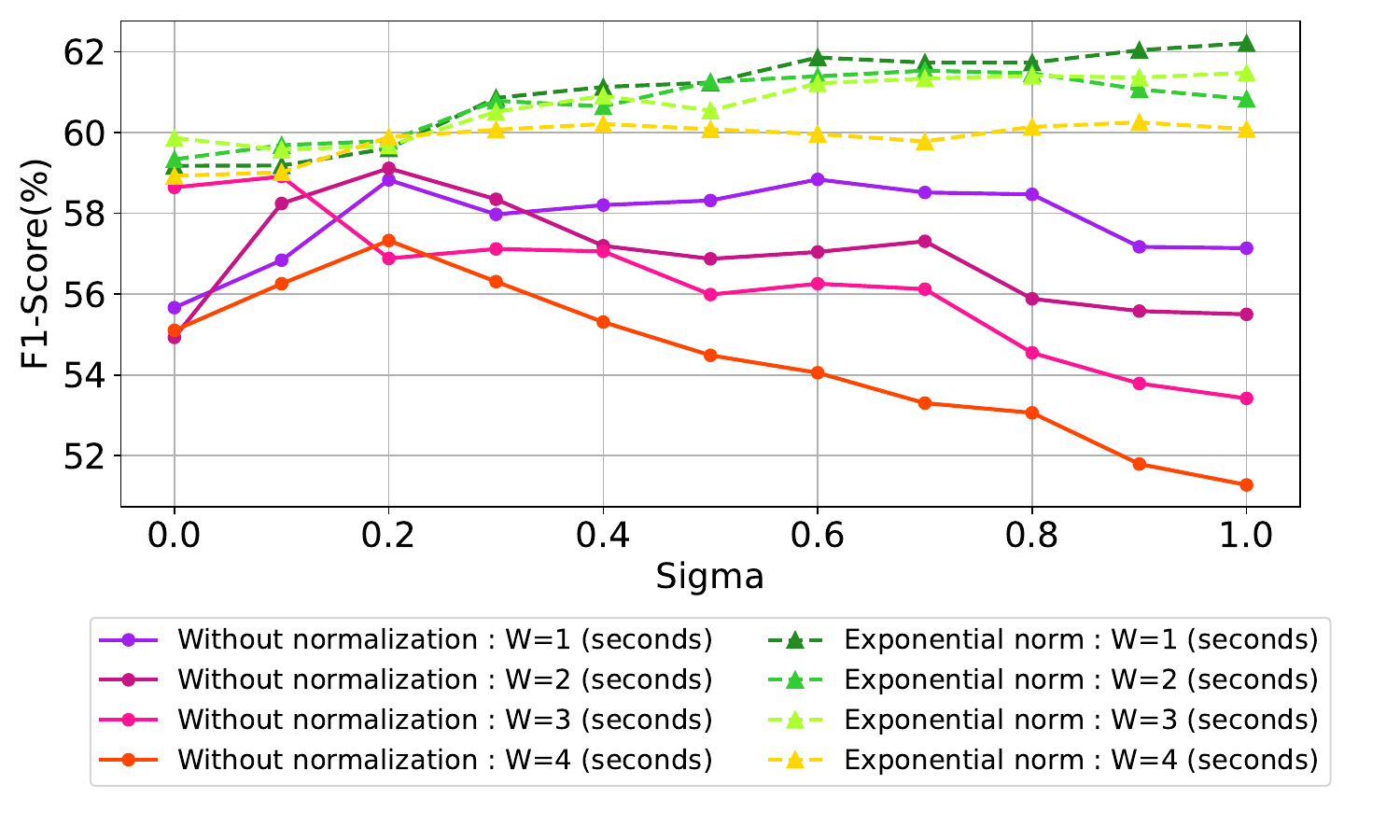}
        \caption{TVSum}
        \label{fig:Exper_ablation_tvSum_norm}
    \end{subfigure}
    \caption[Ablating scene scores and within scene frame normalization]{\textbf{Ablating scene scores and within scene frame normalization.} F1 scores results on the test set of SumMe dataset  and TVSum dataset across more diverse segment durations 
    ($W \in [1,2,3,4]$ seconds) and $\sigma$ values ($\sigma \in [0.0, 1.0]$).}
    \label{fig:Exper_ablation_norm}
\end{figure*}

\textbf{Normalization Impact.} To evaluate the importance of applying normalization functions to the scene scores obtained from the LLM (see Section 3.3 (main paper)) and to the frame scores within each scene (see Section 3.4.3 (main paper)), we conducted experiments across a wider range of segment durations. Specifically, we applied Min-Max normalization for the SumMe dataset and Exponential normalization for the TVSum dataset, and compared performance with and without normalization.

As shown in Figures~\ref{fig:Exper_ablation_sumMe_norm} and~\ref{fig:Exper_ablation_tvSum_norm}, across both datasets and for varying segment durations and frame weight values $\sigma$, applying a normalization function consistently outperforms not using one.\\

\begin{figure*}[h]
    \centering
    \begin{subfigure}[t]{0.49\linewidth}
        \centering
        \includegraphics[width=\linewidth]{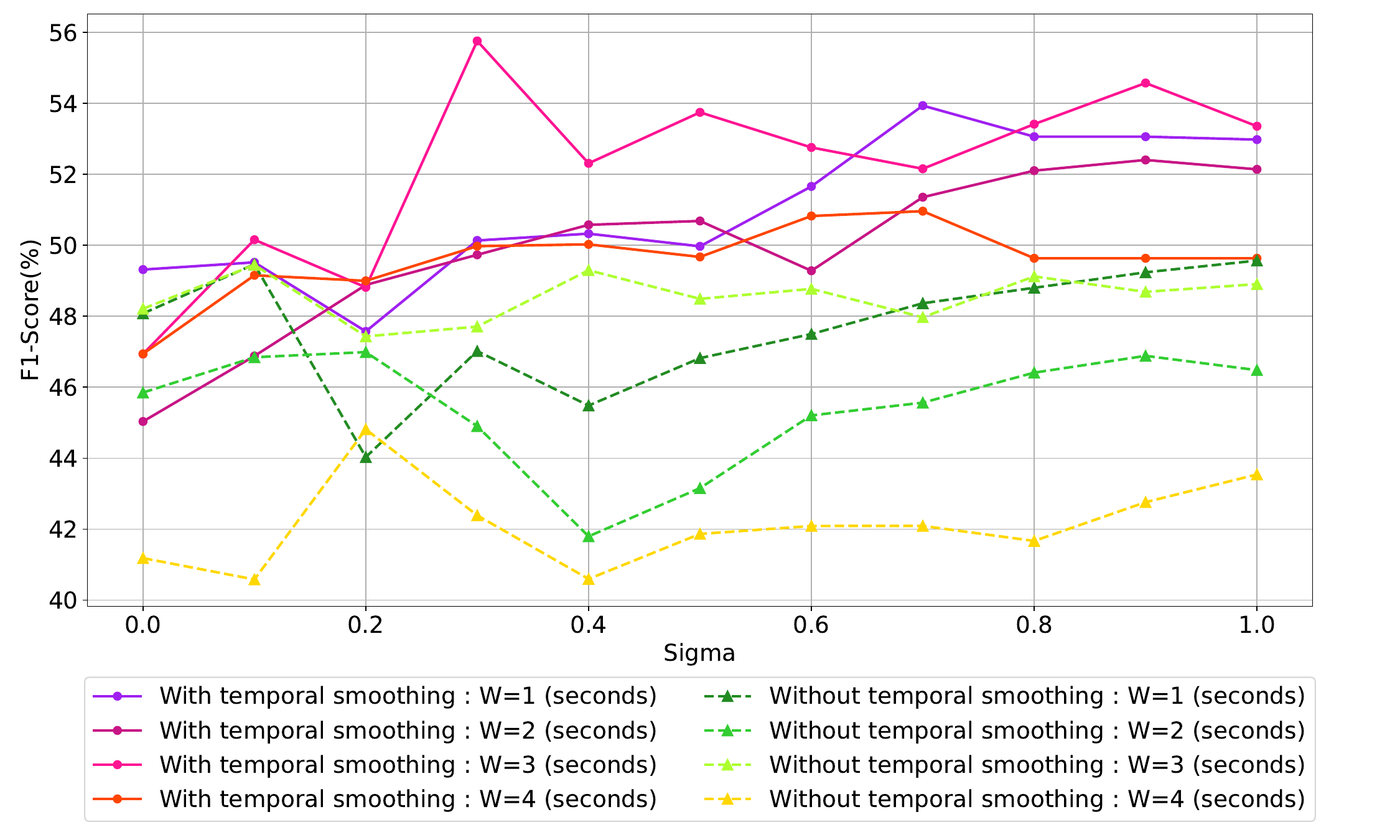}
        \caption{SumMe}
        \label{fig:Exper_ablation_sumMe_smooth}
    \end{subfigure}
    \hfill
    \begin{subfigure}[t]{0.49\linewidth}
        \centering
        \includegraphics[width=\linewidth]{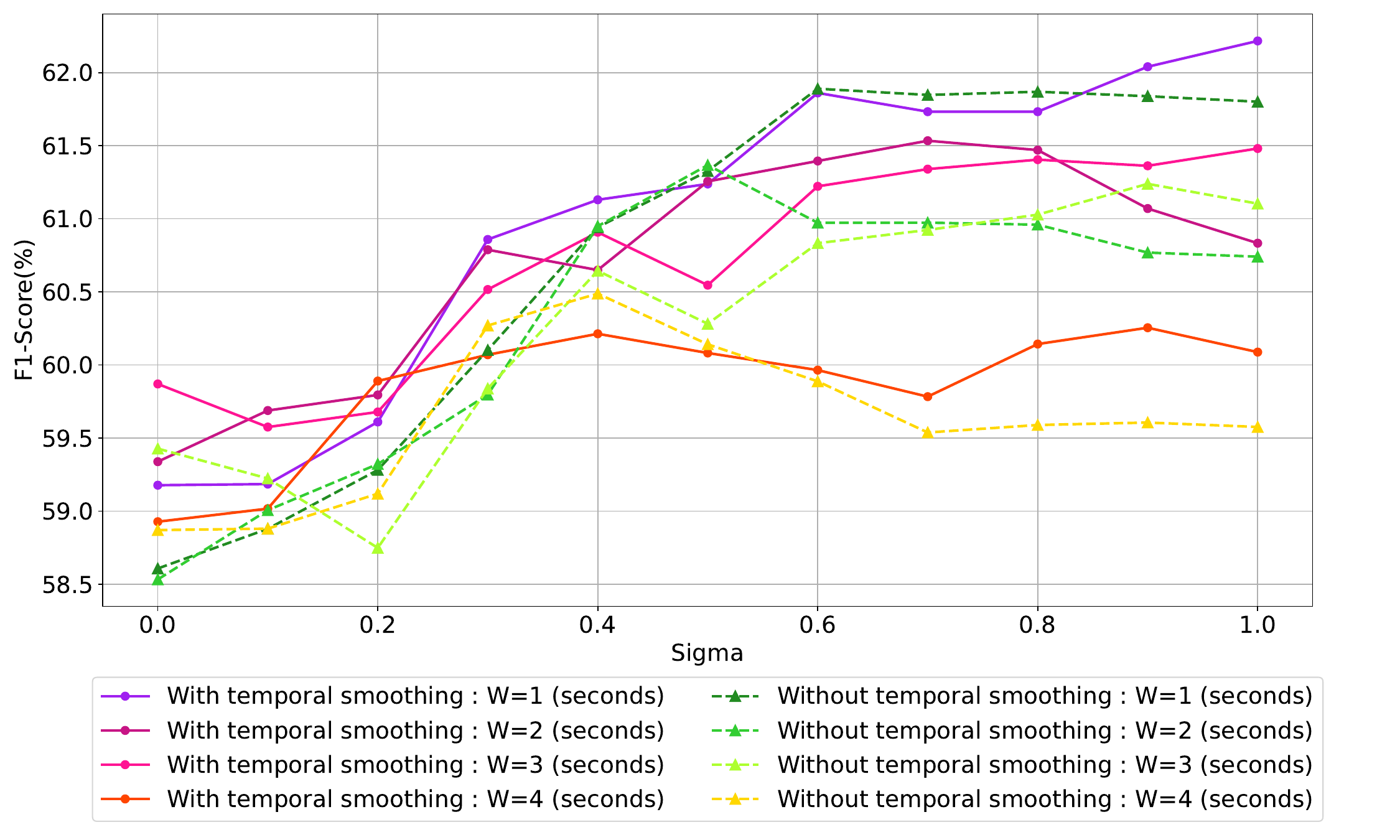}
        \caption{TVSum}
        \label{fig:Exper_ablation_tvSum_smooth}
    \end{subfigure}
    \caption[Ablating temporal smoothing function]{\textbf{Ablating temporal smoothing function.} F1 scores results on the test set of SumMe dataset  and TVSum dataset across more diverse segment durations 
    ($W \in [1,2,3,4]$ seconds) and $\sigma$ values ($\sigma \in [0.0, 1.0]$).}
    \label{fig:Exper_ablation_smooth}
\end{figure*}

\textbf{Temporal smoothing  function affect.} Another key component in our pipeline is the temporal smoothing function. In Figure~\ref{fig:Exper_ablation_smooth}, we analyze the effect of employing this function versus not using it.
It is clear that the smoothing function is critical, as its usage consistently yields better accuracy across all variations. Notably, its impact is more pronounced in the SumMe dataset (see Figure~\ref{fig:Exper_ablation_sumMe_smooth}). We suspect this is due to the misalignment between user-selected key-fragments and the segments produced by KTS, which the smoothing function helps to mitigate.

\subsection{Prompt sensitivity}\label{sec:prompt_sensitivity}
\begin{figure}[t]
    \centering

    \begin{subfigure}[t]{0.49\linewidth}
        \centering
        \includegraphics[width=\linewidth]{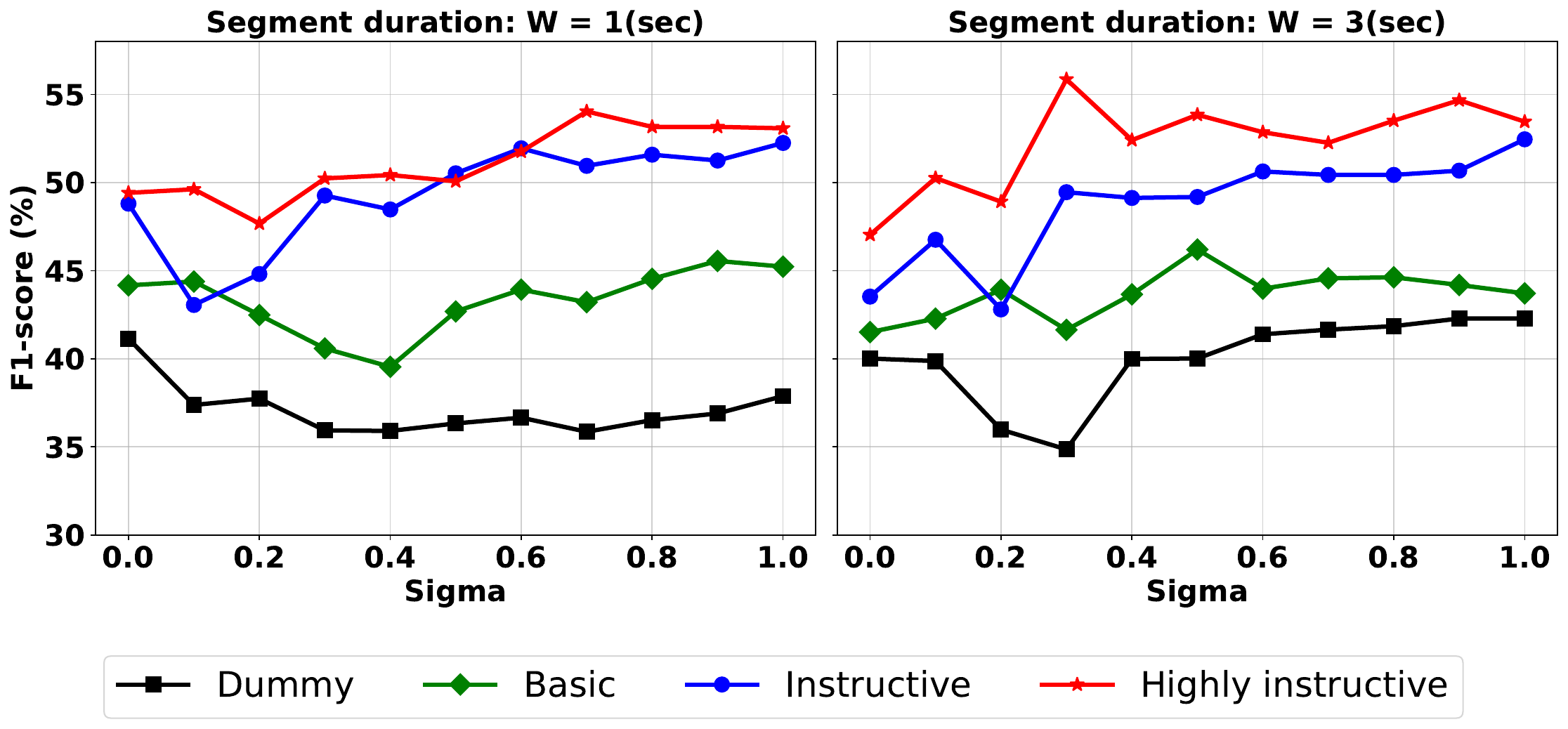}
        \caption{SumMe}
        \label{fig:Exper_ablation_sumMe_prompt}
    \end{subfigure}
    \hfill
    \begin{subfigure}[t]{0.49\linewidth}
        \centering
        \includegraphics[width=\linewidth]{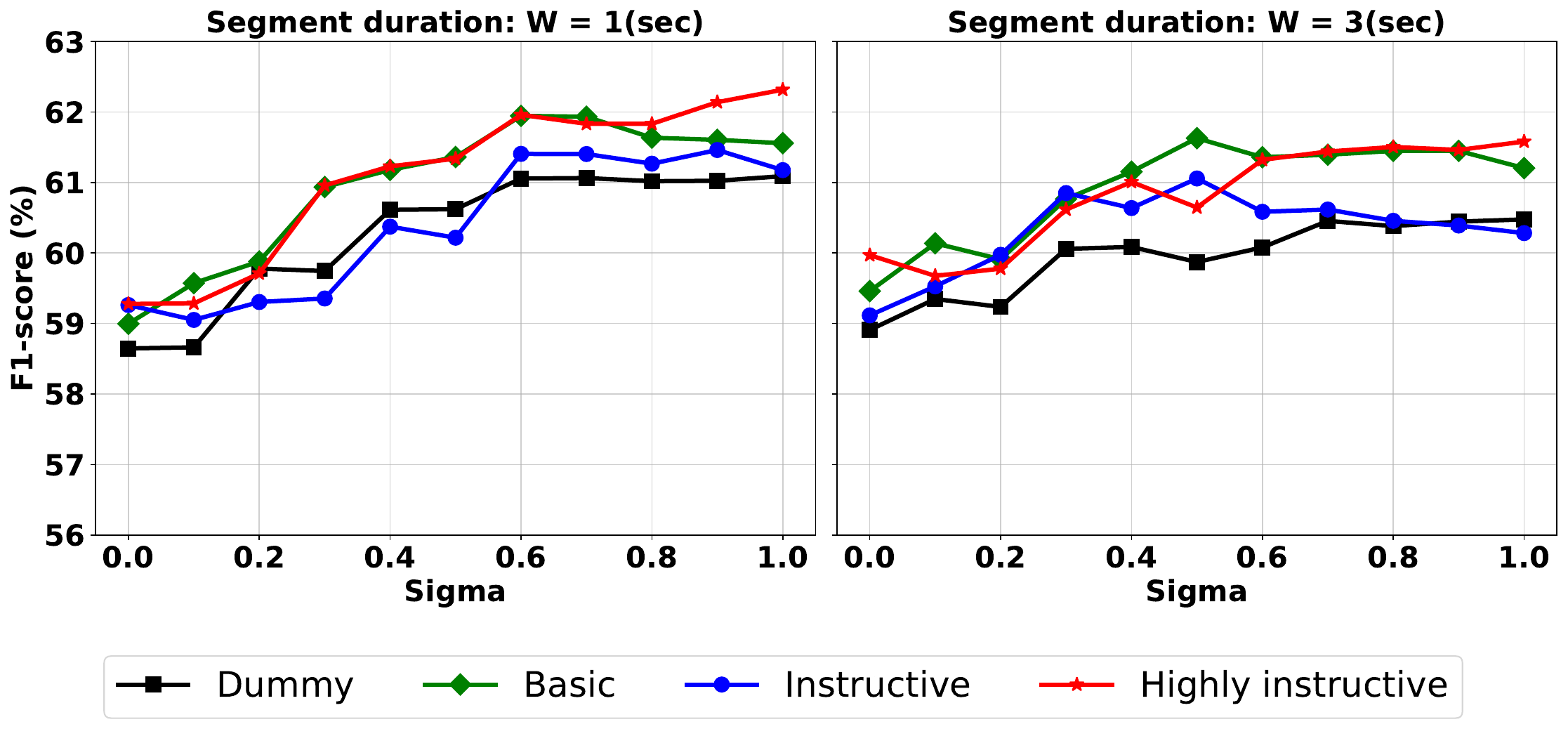}
        \caption{TVSum}
        \label{fig:Exper_ablation_tvSum_prompt}
    \end{subfigure}
    \caption[Ablating prompt sensitivity]{\textbf{Ablating prompt sensitivity.} F1 scores results on the test set of SumMe dataset  and TVSum dataset across diverse segment durations 
    ($W \in [1,3]$ seconds) and $\sigma$ values ($\sigma \in [0.0, 1.0]$).}
    \label{fig:Exper_ablation_prompt}
\end{figure}

Large Language Models (LLMs) are known to be highly sensitive to prompt design. Small variations in phrasing or structure can lead to significantly different outputs, particularly in tasks involving subjective interpretation, such as video summarization. To investigate this sensitivity, we conducted experiments (see Figure~\ref{fig:Exper_ablation_prompt}) using four types of prompts:

\begin{enumerate}
    \item \textbf{Dummy prompt}: A minimal placeholder prompt with no additional instructions or rubric for scoring scene importance.
    
    \item \textbf{\textcolor{mygreen}{Basic prompt}}: Provides the model with simple instructions to score scene importance based on overall relevance. Includes a rubric scale (1–100, with defined ranges for minimal to high importance) to guide scoring. We build upon the dummy prompt by adding the \textcolor{mygreen}{green} parts (see Section~\ref{sec:appendix_LLM_prompt} in the appendix).
    
    \item \textbf{\textcolor{blue}{Instructive prompt}}: Builds upon the Basic prompt by adding more details about each score range and explicitly stating the goal of the task. This helps the model understand how each scene contributes to the overall summary and clarifies what each score represents. We build upon the basic prompt by adding the \textcolor{blue}{blue} parts (see Section~\ref{sec:appendix_LLM_prompt} in the appendix).
    
    \item \textbf{\textcolor{red}{Highly instructive prompt}}: An enhanced prompt containing explicit instructions on how to distribute importance scores across frames. Specifically, it guides the model to avoid a uniform distribution and instead assign higher scores to a small subset of key scenes. It maintains a mean score of approximately 50 out of 100, representing a balanced average importance level. We build upon the instructive prompt by adding the \textcolor{red}{red} parts (see Section~\ref{sec:appendix_LLM_prompt} in the appendix).
\end{enumerate}

We evaluated the four prompt stages—\emph{Dummy}, \emph{Basic}, \emph{Instructive}, and \emph{Highly Instructive}—on the SumMe and TVSum datasets, using GPT-4o as the LLM judge. Experiments were conducted across different segment durations $W \in {1,3}$ and values of $\sigma$, which controls the consistency weighting of frames within each segment. The results, shown in Figure~\ref{fig:Exper_ablation_prompt}, demonstrate a clear performance trend as prompt specificity increases. In particular, the \emph{Highly Instructive} prompt consistently outperforms the simpler prompt variants, indicating that explicitly constraining score distribution—by encouraging high importance scores for only a small subset of critical scenes while maintaining an overall mean near 50/100—enables the model to better approximate human-like importance judgments.

The improvement is more noticeable on the SumMe dataset (see Figure~\ref{fig:Exper_ablation_sumMe_prompt}), which is expected since SumMe represents a more challenging evaluation setting. Specifically, SumMe evaluates performance using the maximum F1 score across multiple annotators, effectively comparing the generated summary to the most similar individual human summary rather than an averaged consensus. This increases sensitivity to individual-level importance judgments and amplifies the impact of prompt design, making refined and structured prompts particularly beneficial. In contrast, TVSum averages annotations before evaluation, reducing variability across annotators and yielding a more stable, consensus-based reference. As a result, prompt-induced performance differences are less pronounced on TVSum dataset (see Figure~\ref{fig:Exper_ablation_tvSum_prompt}).

\begin{table}[h]
\centering
\caption[Prompt sensitivity analysis on SumMe and TVSum datasets]{Prompt sensitivity analysis on SumMe and TVSum datasets.}
\label{tab:prompt_sensitivity}
\resizebox{1.0\linewidth}{!}{
\begin{tabular}{lcccccccc}

                        & \multicolumn{4}{c}{\textbf{\underline{SumMe}}} & \multicolumn{4}{c}{\textbf{\underline{TVSum}}} \\

Prompt Type                       &\textbf{Dummy} & \textbf{\textcolor{mygreen}{Basic}} & \textbf{\textcolor{blue}{Instructive}} & \textbf{\textcolor{red}{Highly instructive}} & 
\textbf{Dummy} &
\textbf{\textcolor{mygreen}{Basic}} & \textbf{\textcolor{blue}{Instructive}} & \textbf{\textcolor{red}{Highly instructive}} \\ \hline
F1-score(\%)(raw)                         & 47.8 & 50.43  & 49.3  & 47.7          & 55.47 &  56.44 &  57.78  & 60.9      \\
F1-score(\%)(w/Norm)                   & 43.7 & 44.5  & 43.3 & 47.84            &  60    & 60.5  & 60.62   & 61.8      \\
F1-score(\%)(final)       & 35.12 & 43.05  & 48.84  & 56.84       & 59.3 & 60.8  & 61.13   & 62.22       \\

\hline 
Scene Score Fluctuation(\%) & Min & Mean & Max & Range     & Min & Mean & Max & Range         \\ 
Mean over scenes                        & 16.92 & 32.46  & 48.12 & 31.2  & 13.22 & 33.13 & 54 & 40.8      \\
\hline
\end{tabular}
}
\end{table}

\noindent
\textbf{Scene score fluctuation under different prompt types.}
Table~\ref{tab:prompt_sensitivity} reports the impact of prompt design on scene importance scoring for the SumMe and TVSum datasets. We compare four prompt types, ranging from minimal instructions (\emph{Dummy}) to a fully \emph{Highly Instructive} prompt that explicitly guides score distribution. The results show that prompt complexity affects both the raw F1-scores and the stability of scene importance assignments. While raw F1-scores may fluctuate across prompt types, performance consistently improves across all stages of our pipeline—from the raw LLM outputs (no normalization or smoothing), through normalization, and finally with normalization and smoothing applied. These gains are most pronounced for the highly instructive prompt, indicating that explicit guidance helps the LLM produce importance scores that better align with the summarization objective.
In addition, the table reports scene score fluctuations, which capture the variability of importance scores assigned within each scene. Specifically, Min, Mean, and Max indicate the average minimum, mean, and maximum differences between scene-level scores across the different prompt designs. The \textbf{Range} represents the average total fluctuation for a scene—that is, on average, the score assigned to a scene could vary up to \textbf{31.2} points on SumMe dataset and up to \textbf{40.8} points on TVSum dataset across its scenes.

\begin{figure}[h!]
    \centering
    \begin{subfigure}[t]{0.46\linewidth}
        \centering
        \includegraphics[width=\linewidth]{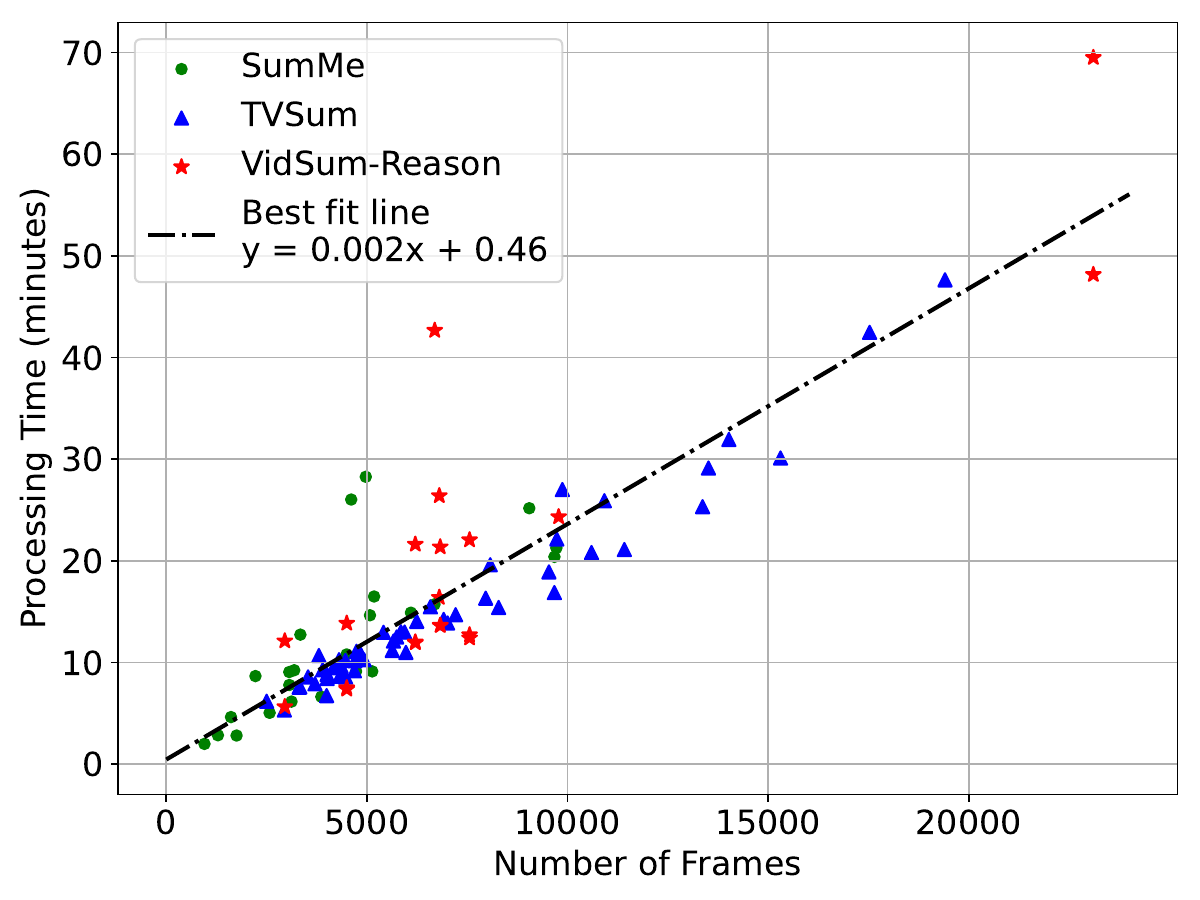}
        \caption[Average processing time for SumMe, TVSum and VidSum-Reason videos as a function of frame count]{\textbf{Average processing time for SumMe, TVSum and VidSum-Reason videos as a function of frame count.} A linear trend is observed, with slope differences explained by varying event distributions across datasets.}
        \label{fig:Exper_tminig_short_videos}
    \end{subfigure}
    \hfill
    \begin{subfigure}[t]{0.48\linewidth}
        \centering
        \hspace{-0.5cm}
        \includegraphics[width=1.03\linewidth]{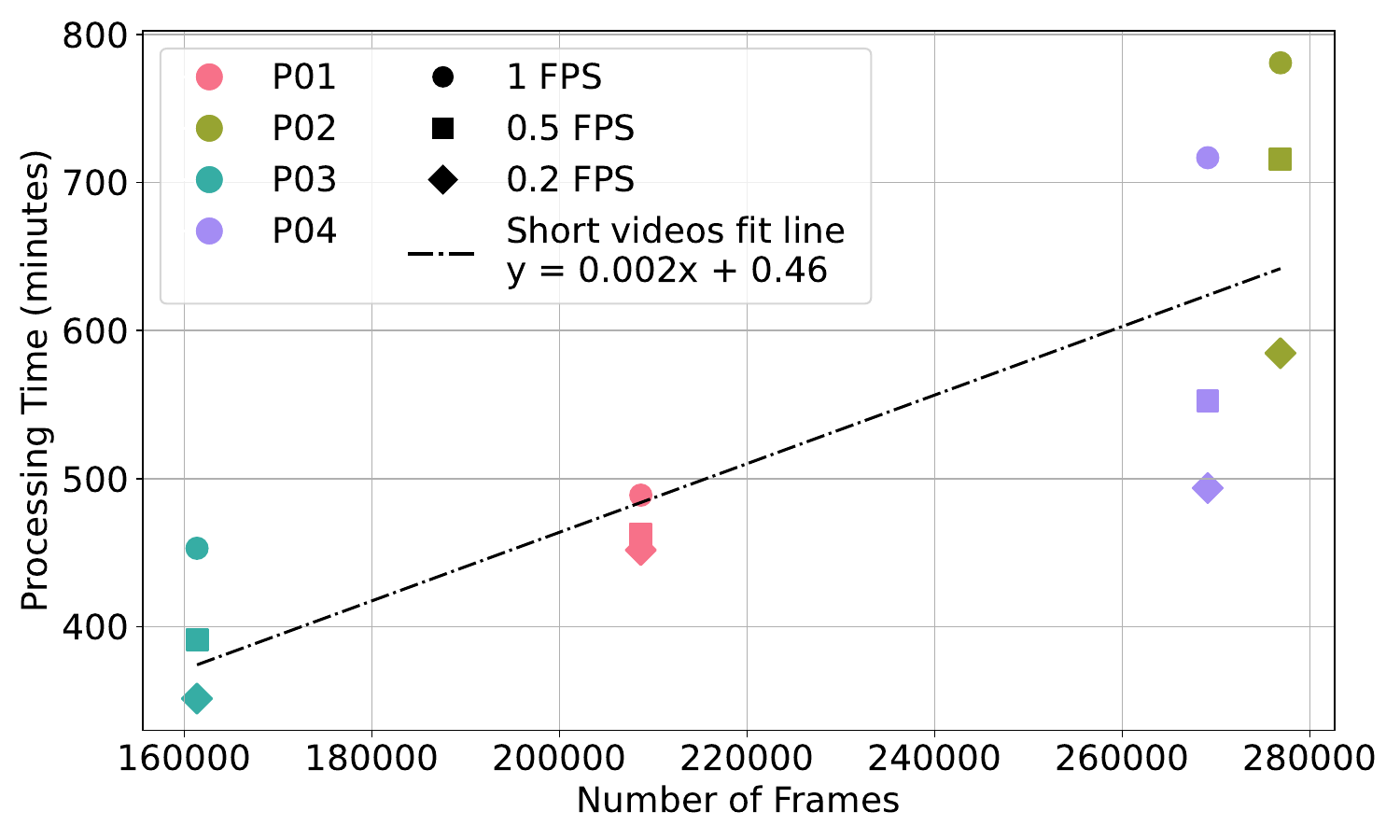}
        \caption[Processing time for QFVS videos at three different sampling rates]{\textbf{Processing time for QFVS videos at three different sampling rates.} Lower FPS yields shorter processing times while preserving dataset coverage, with a near-linear scaling indicated by the fitted line.}
        \label{fig:Exper_tminig_long_videos}
    \end{subfigure}
    \caption[Comparison of processing time trends across datasets and sampling strategies]{\textbf{Comparison of processing time trends across datasets and sampling strategies.}
        Figure~\ref{fig:Exper_tminig_short_videos} presents the average processing time for short videos (SumMe, TVSum and VidSum-Reason) against frame count, showing a clear linear relationship.
        Figure~\ref{fig:Exper_tminig_long_videos} illustrates the processing time for QFVS long videos under different frame sampling rates. Lower FPS significantly reduces computational cost while retaining full content coverage.}
    \label{fig:Exper_ablation_timing}
\end{figure}

\section{Our pipeline's processing time}\label{sec:appendix_timing}
%\addcontentsline{toc}{subsection}{Our pipeline's processing time}

 To evaluate the computational efficiency of our approach, we measured the average processing time for each dataset, alongside the corresponding dataset size and average frame count (Table \ref{tab:videos_timing}). For smaller-scale datasets such as SumMe and TVSum, the complete pipeline (including frame extraction, scene detection, description generation, scene and frame scoring) executes within approximately 12–15 minutes per video on average. Larger datasets, particularly ~\mydata{}, require slightly longer processing times due to their higher frame counts and the exiting of user provided queries.
 
Figure~\ref{fig:Exper_tminig_short_videos} shows the average processing time for videos from SumMe, TVSum and VidSum-Reason as a function of frame count. A clear linear trend is observed, indicating that processing time scales proportionally with the number of frames. Differences in slope across datasets can be attributed to variations in event distribution: SumMe videos are less densely packed with events compared to TVSum and \mydata{}, resulting in differences in scene complexity, scene detection, and description generation times. Despite these variations, the overall relationship remains consistent, highlighting the importance of strategies such as frame-rate reduction to maintain scalability for longer videos.\\

%Figure~\ref{fig:Exper_tminig_long_videos} illustrates processing time per video for the QFVS dataset under three different frame-sampling rates. As expected, lower FPS sampling substantially reduces processing time while retaining the same total video content . The fitted trend line for short videos confirms a near-linear relationship between frame count and processing time, demonstrating that frame sampling effectively reduces computational cost without compromising dataset coverage, and achieves a per-frame processing rate lower than that of short videos.

% In the document:

\textbf{To address scalability challenges when processing longer videos}—particularly the high computational cost of the description generation stage—we adopted a lower frame rate sampling strategy for long-form videos. Specifically, for the QFVS dataset, we sampled videos at reduced FPS rates (0.5 and 0.2) compared to the default 1.0 FPS.  

The fitted trend line for short videos confirms a near-linear relationship between frame count and processing time (see Figure~\ref{fig:Exper_tminig_long_videos}). In Table~\ref{tab:videos_timing}, we report the average processing time for each dataset and the percentage of time spent on each stage of our pipeline.

As expected, long videos (QFVS videos) required significantly more time, with the description generation stage accounting for on average 52\% of the total processing time. To reduce this bottleneck, we evaluated our pipeline at lower FPS rates (0.5 and 0.2). While this reduced the description generation time substantially (from 317 to 164 and 112 minutes for 1.0, 0.5, and 0.2 FPS, respectively), the F1-score decreased by approximately 2 percentage points. This decrease is an expected effect of lower frame sampling, which directly impacts the quality of the generated descriptions and subsequent LLM judgments.

We note that the reported runtime in Table~\ref{tab:videos_timing} corresponds to a sequential batch-wise description generation protocol. This step can be naturally \emph{embarrassingly parallel} across batches (e.g., via multiple concurrent calls), and thus the wall-clock time can be reduced proportionally with the degree of parallelization.

\begin{table}[t]
\centering
\caption[Processing time and frame count across datasets]{\textbf{Processing time and frame count across datasets.} 
For datasets with multiple queries per video (~\mydata{} and QFVS), the average number of queries per video is reported in parentheses. }
\label{tab:videos_timing}
\resizebox{1.0\linewidth}{!}{
\begin{tabular}{l|cccccccc}

\textbf{Dataset} & \textbf{\#Videos (\#Query)} & $\mathbf{\overline{\#Frames}}$ & $\mathbf{\overline{Time\,(min)}}$ & $\mathbf{\overline{Scene\,detection\,(\%)}}$ & $\mathbf{\overline{Description\,generation\,(\%)}}$ & $\mathbf{\overline{LLM\,scoring\,(\%)}}$ & $\mathbf{\overline{Frame\,weighting\,(\%)}}$ & \textbf{F1-score(\%)} \\ 
\hline
SumMe           & 25 (0) & 4,393   & 12    & 0.54 & 0.35 & 0.01 & 0.13 & 56.84 \\
TVSum           & 50 (0) & 7,047   & 15.2  & 0.44 & 0.38 & 0.02 & 0.18 & 62.22 \\
vidSum-Reason   & 9 (2.22) & 7,872   & 20.7  & 0.55 & 0.20 & 0.00 & 0.19 & 43.40 \\ 
QFVS (1.0 fps)  & 4 (45)  & 228,994 & 610   & 0.23 & \underline{0.52} & 0.08 & 0.08 & 53.42 \\ 
QFVS (0.5 fps)  & 4 (45)  & 228,994 & 530   & 0.44 & \underline{0.31} & 0.08 & 0.09 & 52.60 \\ 
QFVS (0.2 fps)  & 4 (45)  & 228,994 & 470   & 0.47 & \underline{0.24} & 0.09 & 0.10 & 51.38 \\ 
\hline
\end{tabular}
}
\end{table}

\section{Additional details about our dataset: \mydata{}}\label{sec:appendix_TGVS_details}
%\addcontentsline{toc}{subsection}{Additional details about our dataset: ~\mydata{}}

\begin{figure}[h]
    \centering
    \begin{minipage}[t]{0.5\textwidth}
        \centering
        \captionof{table}[\textbf{\mydata{} dataset metadata}]{\textbf{\mydata{} dataset metadata.} 
Video-level statistics of the dataset, reporting frame rate, temporal duration, total frame count, and spatial resolution.}
        \resizebox{1.0\linewidth}{!}{
        \begin{tabular}{|c|c|c|c|c|}
        \hline
        \textbf{Video ID} & \textbf{FPS} & \textbf{Duration (s)} & \textbf{\#Frames} & \textbf{Resolution}  \\\hline
        3XA0bB79oGc & 24.00 & 258.71 & 6209  & 1280$\times$720    \\
        5L4DQfVIcdg & 24.00 & 314.96 & 7559  & 1280$\times$536    \\
        5dY1kU\_3fTI & 60.00 & 385.52 & 23108 & 1280$\times$720   \\
        77AA5bWCGlE & 30.00 & 226.83 & 6805  & 1280$\times$720    \\
        9QABNfDhSxs & 24.00 & 123.29 & 2956  & 1280$\times$720    \\
        EXeTwQWrcwY & 30.00 & 150.12 & 4499  & 1280$\times$544    \\
        GIVerZ9mUpU & 30.00 & 223.19 & 6689  & 1920$\times$1080   \\
        GgtUWB405Mo & 25.00 & 273.12 & 6828  & 1280$\times$720    \\
        d7ijM1jGou0 & 30.00 & 120.45 & 3610  & 1280$\times$720    \\
        wvDENCN4i3c & 30.00 & 326.23 & 9777  & 854$\times$480     \\
        \hline
        \textbf{Average} & \textbf{30.70} & \textbf{240.24} & \textbf{7804} & --  \\\hline
        \end{tabular}%
        }
        \label{tab:mydataset_metadata}
    \end{minipage}
    \hfill
    \begin{minipage}[t]{0.48\textwidth}
        \vspace{+0.9cm}
        \centering
        \includegraphics[width=0.85\linewidth]{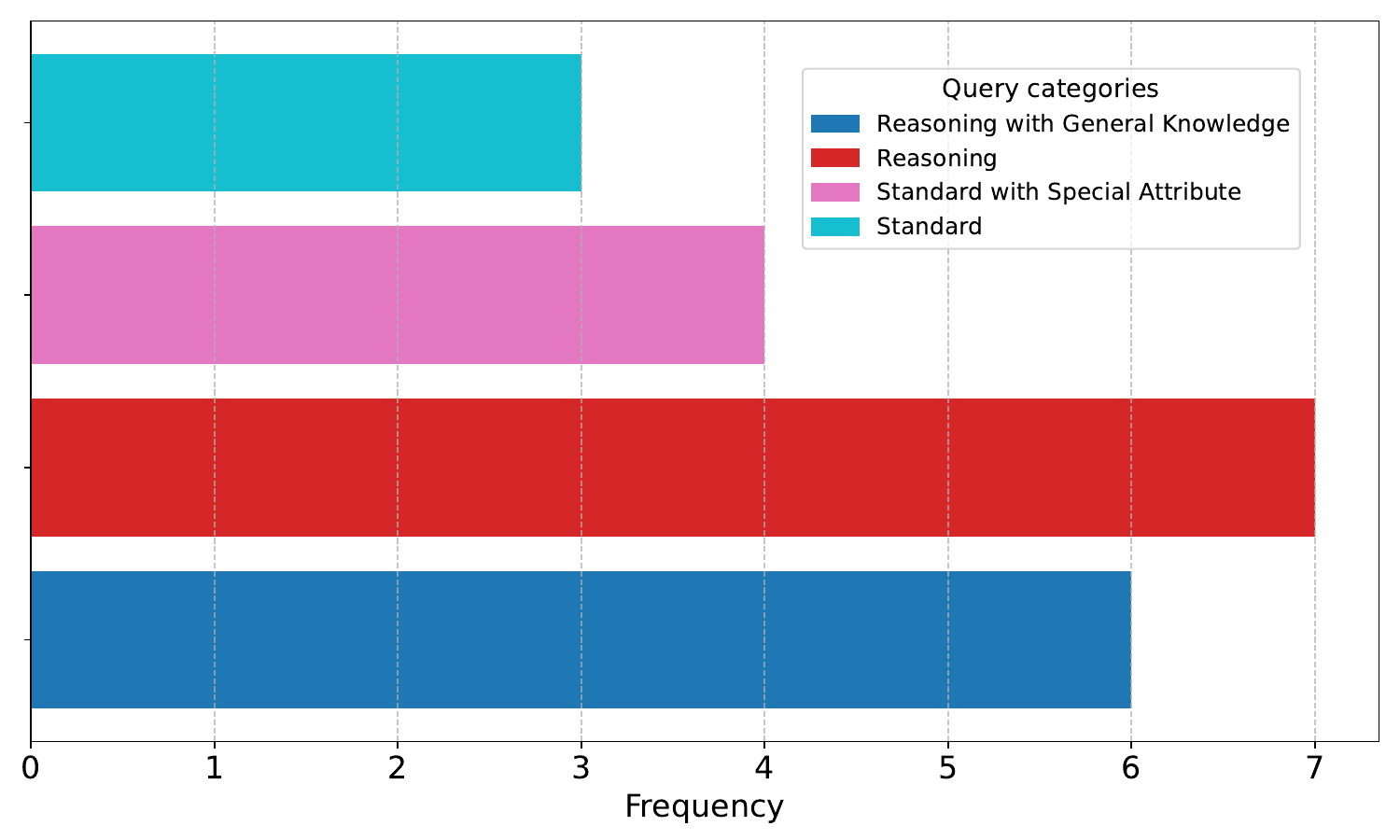}
        \captionof{figure}[\textbf{\mydata{}}: Frequency distribution of query categories]{\textbf{\mydata{}}: Frequency distribution of query categories.}
        \label{fig:TGVS_CategoryFreq}
    \end{minipage}
\end{figure}

In Table~\ref{Table:ourdata}, we provide the full ~\mydata{} dataset. Each video-query pair is accompanied by its query class and the corresponding textual query. Additionally, Figure~\ref{fig:TGVS_CategoryFreq} shows the frequency of each query class in the \mydata{} dataset.

Additionally, we provide metadata for our dataset (see Table ~\ref{tab:mydataset_metadata}), \mydata{}, including FPS, duration, number of frames, and resolution. The original videos can be accessed via their \textit{URL} as follows: \texttt{https://www.youtube.com/watch?v=<video ID>}.

\begin{table*}[h]
\centering
\caption[List of query-annotated videos in the ~\mydata{} dataset]{\textbf{List of query-annotated videos in the ~\mydata{} dataset.} Each row corresponds to a video, showing the video ID, the query class (which defines the type of information requested), and the user-provided query used to guide summarization. This table highlights the diversity of query intents and supports evaluation of personalized video summarization.}\label{Table:ourdata}
\begin{adjustbox}{width=\textwidth}
\begin{tabular}{|c|c|c|}
\hline
\textbf{Video ID} & \textbf{Query Category} & \textbf{Text Query} \\ \hline
GIVerZ9mUpU & Reasoning with General Knowledge & Filter any scenes that involve violence \\ \hline
wvDENCN4i3c & Reasoning & Focus on emotional reactions and scenes \\ \hline
5L4DQfVIcdg & Standard with Special Attribute & Focus on scenes with the coin machine \\ \hline
5L4DQfVIcdg & Reasoning & Highlight scenes showing the passage of time \\ \hline
5L4DQfVIcdg & Reasoning & Prioritize scenes that demonstrate emotions \\ \hline
3XA0bB79oGc & Standard & Focus on scenes with both the boy and the puppy \\ \hline
3XA0bB79oGc & Reasoning & Exclude scenes with negative emotions \\ \hline
3XA0bB79oGc & Reasoning & Prioritize scenes with emotional shifts \\ \hline
77AA5bWCGlE & Reasoning with General Knowledge & Focus on scenes where the score changes \\ \hline
77AA5bWCGlE & Standard with Special Attribute & Highlight the free throws \\ \hline
5dY1kU\_3fTI & Reasoning & Highlight every scene where measurement is involved \\ \hline
5dY1kU\_3fTI & Standard with Special Attribute & Highlight every scene where non-power (hand) tools are used \\ \hline
GgtUWB405Mo & Standard with Special Attribute & Focus on cars with bright colors \\ \hline
GgtUWB405Mo & Reasoning with General Knowledge & Highlight crossover vehicles in the video \\ \hline
GgtUWB405Mo & Reasoning with General Knowledge & Prioritize German cars \\ \hline
EXeTwQWrcwY & Standard & Show scenes with explosions \\ \hline
EXeTwQWrcwY & Reasoning with General Knowledge & Show scenes that feature the Joker or that feel related to him \\\hline
EXeTwQWrcwY & Reasoning & Highlight moments that seem like public spectacle or protest \\ \hline
9QABNfDhSxs & Reasoning with General Knowledge & Exclude scenes that aren’t PG-13 appropriate \\ \hline
9QABNfDhSxs & Standard & Focus on scenes that show blood \\ \hline
\end{tabular}
\end{adjustbox}
\end{table*}

\paragraph{Annotation process.}
\mydata{} was created as a proof of concept to demonstrate that LLM–\vlm{} systems, when properly orchestrated, can perform robust video summarization for long-tailed and reasoning-heavy queries beyond surface-level prompts.

The dataset was annotated by a single primary annotator, followed by a structured verification process conducted by two additional reviewers. Because these reviewers performed quality control rather than independent re-annotation, standard inter-annotator agreement metrics (e.g., Cohen’s Kappa), which require multiple independent labels per item, are not applicable under their conventional definition.

The verification process audited each annotated item according to four criteria:
(i) query clarity and lack of ambiguity,
(ii) summary correctness with respect to the video content,
(iii) evidential grounding in observable events rather than assumptions, and
(iv) schema validity (format and completeness).
Reviewer~1 and Reviewer~2 independently conducted first-pass audits and flagged items requiring correction or clarification. The three annotators then performed a second-pass review focusing on flagged cases (and an additional stratified sample) to validate revisions. Any remaining uncertainty was resolved through adjudication, and the released annotations reflect the adjudicated outcome.

\section{Discussion: Why a Gap to Fully Supervised Methods Can Remain}
\label{sec:sup_discussion_gap_supervised}
%\addcontentsline{toc}{subsection}{Discussion: Why a Gap to Fully Supervised Methods Can Remain}

Although our approach is competitive with strong baselines and can rival the best results on some settings, a residual gap to top-tier \emph{fully supervised} methods may still remain in certain benchmarks. This gap is expected given a fundamental difference in the learning signal and the assumptions available to each paradigm.

\paragraph{Benefit of dataset-specific importance labels.}
Fully supervised methods are trained with \emph{direct access} to dataset-specific importance annotations (e.g., frame/shot-level scores or selected segments). This supervision allows them to learn strong domain- and benchmark-specific priors about what is considered ``important'' under the evaluation protocol. For example, under TVSum style evaluation, supervised models can internalize common annotator preferences (e.g., emphasis on salient actions or visually distinctive events), and can exploit dataset domain uniqueness that repeatedly correlate with higher importance scores.

\noindent
In contrast, our approach operates \emph{without access to ground-truth segment boundaries or importance labels}. Unlike prior methods that rely on pre-segmented videos and in same cases directly predict segment-level scores aligned with the evaluation protocol, we infer importance via LLM-based semantic reasoning over automatically detected scenes, jointly addressing temporal localization and importance estimation.

\paragraph{Calibration and thresholding under the evaluation protocol.}
Supervised training also enables precise calibration of scoring distributions and selection thresholds to match the benchmark's summary-length constraints and evaluation metrics. Concretely, the model can learn how to map features to importance scores that are well aligned with the dataset's annotation scale, and how to tune selection behavior (e.g., coverage vs.\ diversity trade-offs) to maximize the benchmark metric. This calibration advantage is particularly impaction in borderline cases where small shifts in scores change whether a segment is included.

\paragraph{Limitations without labels or even training data.}
In contrast, our method operates \emph{without} dataset-specific importance labels---and, in our setting, even without access to the training data. Consequently, it cannot directly learn benchmark-specific priors. This can yield more conservative selections (e.g., preferring high-confidence salient events) or occasional mismatches in ambiguous scenarios where multiple segments appear similarly relevant, but the benchmark's annotation protocol consistently favors one choice.

\paragraph{Takeaway.}
Overall, the remaining gap should be interpreted as the cost of operating without labels (and without data), rather than a limitation of the core mechanism. Importantly, despite this constraint, our approach remains competitive and, in some settings, effectively rivals the best-performing supervised approaches while requiring substantially weaker assumptions.

\paragraph{Our setting is harder, but also more generic.}
Finally, it is important to emphasize that our evaluation setting is inherently more challenging: we operate without dataset-specific importance labels and (in our case) without access to the benchmark training data, whereas supervised methods can directly optimize for the target protocol. This stronger constraint makes our approach more \emph{generic} by design, as it does not rely on domain- or benchmark-specific priors. As a result, we expect it to be better suited for deployment in \emph{unseen domains} and under distribution shifts, where supervised models may overfit to dataset-specific notions of importance and calibration.

\section{Novelty of This Work}
\label{sec:novelty}
%\addcontentsline{toc}{subsection}{Novelty of This Work}

We propose a \emph{training-free}, \emph{text-guided} video summarization framework that tightly couples pretrained video--language captioning with LLM-based reasoning to produce controllable summaries without task-specific supervision. Our main novelties are:

\begin{itemize}
    \item \textbf{Orchestration of pretrained ~\vlm{} and LLMs.} 
    To our knowledge, we are the first to cast video summarization as \emph{LLM-based importance judging} over structured semantic units---captions generated by pretrained video--language models---and to use these judgments to directly rank and select video content.

    \item \textbf{Scalable caption generation for long-form videos.}
   We introduce a scalable, practical long-video captioning procedure based on a memory-aware batching schedule and a standardized caption style, enabling parsing of entire videos while ensuring efficient processing and preserving cross-segment consistency and comparability of descriptions.

    \item \textbf{Prompt-engineered, reproducible importance scoring.}
    We design the judging step as a controlled, rubric-driven prompting protocol (explicit role, decision criteria, and constraints), yielding more stable importance scores and improving the reproducibility of LLM-driven summarization.

    \item \textbf{Fine-grained frame scoring via principled score propagation.}
    Beyond segment selection, we propose lightweight propagation rules that translate caption/segment-level importance into \emph{frame-level} saliency by combining (i) temporal smoothing and (ii) within-segment weighting based on two measures we define: \emph{consistency} (temporal coherence) and \emph{uniqueness} (embedding-space novelty derived from clustering).

    \item \textbf{Zero-shot, domain-agnostic, and naturally controllable.}
    The entire pipeline operates in a zero-shot setting---without fine-tuning or domain-specific training data---and supports natural language control, making it readily transferable across diverse video domains and user intents.
\end{itemize}


\begin{thebibliography}{10}
\urlstyle{rm}
\expandafter\ifx\csname url\endcsname\relax
  \def\url#1{\texttt{#1}}\fi
\expandafter\ifx\csname urlprefix\endcsname\relax\def\urlprefix{URL }\fi
\expandafter\ifx\csname doiprefix\endcsname\relax\def\doiprefix{DOI: }\fi
\providecommand{\bibinfo}[2]{#2}
\providecommand{\eprint}[2][]{\url{#2}}

\bibitem{ma2002user}
\bibinfo{author}{Ma, Y.-F.}, \bibinfo{author}{Lu, L.}, \bibinfo{author}{Zhang, H.-J.} \& \bibinfo{author}{Li, M.}
\newblock \bibinfo{title}{A user attention model for video summarization}.
\newblock In \emph{\bibinfo{booktitle}{Proceedings of the tenth ACM international conference on Multimedia}}, \bibinfo{pages}{533--542} (\bibinfo{year}{2002}).

\bibitem{money2008video}
\bibinfo{author}{Money, A.~G.} \& \bibinfo{author}{Agius, H.}
\newblock \bibinfo{journal}{\bibinfo{title}{Video summarisation: A conceptual framework and survey of the state of the art}}.
\newblock {\emph{\JournalTitle{Journal of visual communication and image representation}}} \textbf{\bibinfo{volume}{19}}, \bibinfo{pages}{121--143} (\bibinfo{year}{2008}).

\bibitem{truong2007video}
\bibinfo{author}{Truong, B.~T.} \& \bibinfo{author}{Venkatesh, S.}
\newblock \bibinfo{journal}{\bibinfo{title}{Video abstraction: A systematic review and classification}}.
\newblock {\emph{\JournalTitle{ACM transactions on multimedia computing, communications, and applications (TOMM)}}} \textbf{\bibinfo{volume}{3}}, \bibinfo{pages}{3--es} (\bibinfo{year}{2007}).

\bibitem{apostolidis2021video}
\bibinfo{author}{Apostolidis, E.}, \bibinfo{author}{Adamantidou, E.}, \bibinfo{author}{Metsai, A.~I.}, \bibinfo{author}{Mezaris, V.} \& \bibinfo{author}{Patras, I.}
\newblock \bibinfo{journal}{\bibinfo{title}{Video summarization using deep neural networks: A survey}}.
\newblock {\emph{\JournalTitle{Proceedings of the IEEE}}} \textbf{\bibinfo{volume}{109}}, \bibinfo{pages}{1838--1863} (\bibinfo{year}{2021}).

\bibitem{rochan2018video}
\bibinfo{author}{Rochan, M.}, \bibinfo{author}{Ye, L.} \& \bibinfo{author}{Wang, Y.}
\newblock \bibinfo{title}{Video summarization using fully convolutional sequence networks}.
\newblock In \emph{\bibinfo{booktitle}{Proceedings of the European conference on computer vision (ECCV)}}, \bibinfo{pages}{347--363} (\bibinfo{year}{2018}).

\bibitem{zhang2016video}
\bibinfo{author}{Zhang, K.}, \bibinfo{author}{Chao, W.-L.}, \bibinfo{author}{Sha, F.} \& \bibinfo{author}{Grauman, K.}
\newblock \bibinfo{title}{Video summarization with long short-term memory}.
\newblock In \emph{\bibinfo{booktitle}{Computer Vision--ECCV 2016: 14th European Conference, Amsterdam, The Netherlands, October 11--14, 2016, Proceedings, Part VII 14}}, \bibinfo{pages}{766--782} (\bibinfo{organization}{Springer}, \bibinfo{year}{2016}).

\bibitem{hussain2021comprehensive}
\bibinfo{author}{Hussain, T.} \emph{et~al.}
\newblock \bibinfo{journal}{\bibinfo{title}{A comprehensive survey of multi-view video summarization}}.
\newblock {\emph{\JournalTitle{Pattern Recognition}}} \textbf{\bibinfo{volume}{109}}, \bibinfo{pages}{107567} (\bibinfo{year}{2021}).

\bibitem{otani2017video}
\bibinfo{author}{Otani, M.}, \bibinfo{author}{Nakashima, Y.}, \bibinfo{author}{Rahtu, E.}, \bibinfo{author}{Heikkil{\"a}, J.} \& \bibinfo{author}{Yokoya, N.}
\newblock \bibinfo{title}{Video summarization using deep semantic features}.
\newblock In \emph{\bibinfo{booktitle}{Computer Vision--ACCV 2016: 13th Asian Conference on Computer Vision, Taipei, Taiwan, November 20-24, 2016, Revised Selected Papers, Part V 13}}, \bibinfo{pages}{361--377} (\bibinfo{organization}{Springer}, \bibinfo{year}{2017}).

\bibitem{antani2002survey}
\bibinfo{author}{Antani, S.}, \bibinfo{author}{Kasturi, R.} \& \bibinfo{author}{Jain, R.}
\newblock \bibinfo{journal}{\bibinfo{title}{A survey on the use of pattern recognition methods for abstraction, indexing and retrieval of images and video}}.
\newblock {\emph{\JournalTitle{Pattern recognition}}} \textbf{\bibinfo{volume}{35}}, \bibinfo{pages}{945--965} (\bibinfo{year}{2002}).

\bibitem{zhang2016context}
\bibinfo{author}{Zhang, S.}, \bibinfo{author}{Zhu, Y.} \& \bibinfo{author}{Roy-Chowdhury, A.~K.}
\newblock \bibinfo{journal}{\bibinfo{title}{Context-aware surveillance video summarization}}.
\newblock {\emph{\JournalTitle{IEEE Transactions on Image Processing}}} \textbf{\bibinfo{volume}{25}}, \bibinfo{pages}{5469--5478} (\bibinfo{year}{2016}).

\bibitem{mahapatra2018videoken}
\bibinfo{author}{Mahapatra, D.}, \bibinfo{author}{Mariappan, R.}, \bibinfo{author}{Rajan, V.}, \bibinfo{author}{Yadav, K.} \& \bibinfo{author}{Roy, S.}
\newblock \bibinfo{title}{Videoken: Automatic video summarization and course curation to support learning}.
\newblock In \emph{\bibinfo{booktitle}{Companion Proceedings of the The Web Conference 2018}}, \bibinfo{pages}{239--242} (\bibinfo{year}{2018}).

\bibitem{alrumiah2022educational}
\bibinfo{author}{Alrumiah, S.~S.} \& \bibinfo{author}{Al-Shargabi, A.~A.}
\newblock \bibinfo{journal}{\bibinfo{title}{Educational videos subtitles’ summarization using latent dirichlet allocation and length enhancement.}}
\newblock {\emph{\JournalTitle{Computers, Materials \& Continua}}} \textbf{\bibinfo{volume}{70}} (\bibinfo{year}{2022}).

\bibitem{almeida2012vison}
\bibinfo{author}{Almeida, J.}, \bibinfo{author}{Leite, N.~J.} \& \bibinfo{author}{Torres, R. d.~S.}
\newblock \bibinfo{journal}{\bibinfo{title}{Vison: Video summarization for online applications}}.
\newblock {\emph{\JournalTitle{Pattern Recognition Letters}}} \textbf{\bibinfo{volume}{33}}, \bibinfo{pages}{397--409} (\bibinfo{year}{2012}).

\bibitem{muhammad2019deepres}
\bibinfo{author}{Muhammad, K.}, \bibinfo{author}{Hussain, T.}, \bibinfo{author}{Del~Ser, J.}, \bibinfo{author}{Palade, V.} \& \bibinfo{author}{De~Albuquerque, V. H.~C.}
\newblock \bibinfo{journal}{\bibinfo{title}{Deepres: A deep learning-based video summarization strategy for resource-constrained industrial surveillance scenarios}}.
\newblock {\emph{\JournalTitle{IEEE Transactions on Industrial Informatics}}} \textbf{\bibinfo{volume}{16}}, \bibinfo{pages}{5938--5947} (\bibinfo{year}{2019}).

\bibitem{vo2025integrate}
\bibinfo{author}{Vo, B.~Q.} \& \bibinfo{author}{Vo, V.~H.}
\newblock \bibinfo{journal}{\bibinfo{title}{Integrate the temporal scheme for unsupervised video summarization via attention mechanism}}.
\newblock {\emph{\JournalTitle{IEEE Access}}}  (\bibinfo{year}{2025}).

\bibitem{abbasi2023adopting}
\bibinfo{author}{Abbasi, M.} \& \bibinfo{author}{Saeedi, P.}
\newblock \bibinfo{title}{Adopting self-supervised learning into unsupervised video summarization through restorative score.}
\newblock In \emph{\bibinfo{booktitle}{2023 IEEE International Conference on Image Processing (ICIP)}}, \bibinfo{pages}{425--429} (\bibinfo{organization}{IEEE}, \bibinfo{year}{2023}).

\bibitem{apostolidis2020ac}
\bibinfo{author}{Apostolidis, E.}, \bibinfo{author}{Adamantidou, E.}, \bibinfo{author}{Metsai, A.~I.}, \bibinfo{author}{Mezaris, V.} \& \bibinfo{author}{Patras, I.}
\newblock \bibinfo{journal}{\bibinfo{title}{Ac-sum-gan: Connecting actor-critic and generative adversarial networks for unsupervised video summarization}}.
\newblock {\emph{\JournalTitle{IEEE Transactions on Circuits and Systems for Video Technology}}} \textbf{\bibinfo{volume}{31}}, \bibinfo{pages}{3278--3292} (\bibinfo{year}{2020}).

\bibitem{he2023align}
\bibinfo{author}{He, B.} \emph{et~al.}
\newblock \bibinfo{title}{Align and attend: Multimodal summarization with dual contrastive losses}.
\newblock In \emph{\bibinfo{booktitle}{Proceedings of the IEEE/CVF conference on computer vision and pattern recognition}}, \bibinfo{pages}{14867--14878} (\bibinfo{year}{2023}).

\bibitem{jiang2022joint}
\bibinfo{author}{Jiang, H.} \& \bibinfo{author}{Mu, Y.}
\newblock \bibinfo{title}{Joint video summarization and moment localization by cross-task sample transfer}.
\newblock In \emph{\bibinfo{booktitle}{Proceedings of the IEEE/CVF Conference on Computer Vision and Pattern Recognition}}, \bibinfo{pages}{16388--16398} (\bibinfo{year}{2022}).

\bibitem{narasimhan2021clip}
\bibinfo{author}{Narasimhan, M.}, \bibinfo{author}{Rohrbach, A.} \& \bibinfo{author}{Darrell, T.}
\newblock \bibinfo{journal}{\bibinfo{title}{Clip-it! language-guided video summarization}}.
\newblock {\emph{\JournalTitle{Advances in neural information processing systems}}} \textbf{\bibinfo{volume}{34}}, \bibinfo{pages}{13988--14000} (\bibinfo{year}{2021}).

\bibitem{ghauri2021supervised}
\bibinfo{author}{Ghauri, J.~A.}, \bibinfo{author}{Hakimov, S.} \& \bibinfo{author}{Ewerth, R.}
\newblock \bibinfo{title}{Supervised video summarization via multiple feature sets with parallel attention}.
\newblock In \emph{\bibinfo{booktitle}{2021 IEEE International Conference on Multimedia and Expo (ICME)}}, \bibinfo{pages}{1--6s} (\bibinfo{organization}{IEEE}, \bibinfo{year}{2021}).

\bibitem{yuan2019cycle}
\bibinfo{author}{Yuan, L.}, \bibinfo{author}{Tay, F.~E.}, \bibinfo{author}{Li, P.}, \bibinfo{author}{Zhou, L.} \& \bibinfo{author}{Feng, J.}
\newblock \bibinfo{title}{Cycle-sum: Cycle-consistent adversarial lstm networks for unsupervised video summarization}.
\newblock In \emph{\bibinfo{booktitle}{Proceedings of the AAAI Conference on Artificial Intelligence}}, vol.~\bibinfo{volume}{33}, \bibinfo{pages}{9143--9150} (\bibinfo{year}{2019}).

\bibitem{sharghi2016query}
\bibinfo{author}{Sharghi, A.}, \bibinfo{author}{Gong, B.} \& \bibinfo{author}{Shah, M.}
\newblock \bibinfo{title}{Query-focused extractive video summarization}.
\newblock In \emph{\bibinfo{booktitle}{Computer Vision--ECCV 2016: 14th European Conference, Amsterdam, The Netherlands, October 11-14, 2016, Proceedings, Part VIII 14}}, \bibinfo{pages}{3--19} (\bibinfo{organization}{Springer}, \bibinfo{year}{2016}).

\bibitem{gong2014diverse}
\bibinfo{author}{Gong, B.}, \bibinfo{author}{Chao, W.-L.}, \bibinfo{author}{Grauman, K.} \& \bibinfo{author}{Sha, F.}
\newblock \bibinfo{journal}{\bibinfo{title}{Diverse sequential subset selection for supervised video summarization}}.
\newblock {\emph{\JournalTitle{Advances in neural information processing systems}}} \textbf{\bibinfo{volume}{27}} (\bibinfo{year}{2014}).

\bibitem{feng2018extractive}
\bibinfo{author}{Feng, L.}, \bibinfo{author}{Li, Z.}, \bibinfo{author}{Kuang, Z.} \& \bibinfo{author}{Zhang, W.}
\newblock \bibinfo{title}{Extractive video summarizer with memory augmented neural networks}.
\newblock In \emph{\bibinfo{booktitle}{Proceedings of the 26th ACM international conference on Multimedia}}, \bibinfo{pages}{976--983} (\bibinfo{year}{2018}).

\bibitem{apostolidis2021combining}
\bibinfo{author}{Apostolidis, E.}, \bibinfo{author}{Balaouras, G.}, \bibinfo{author}{Mezaris, V.} \& \bibinfo{author}{Patras, I.}
\newblock \bibinfo{title}{Combining global and local attention with positional encoding for video summarization}.
\newblock In \emph{\bibinfo{booktitle}{2021 IEEE international symposium on multimedia (ISM)}}, \bibinfo{pages}{226--234} (\bibinfo{organization}{IEEE}, \bibinfo{year}{2021}).

\bibitem{potapov2014category}
\bibinfo{author}{Potapov, D.}, \bibinfo{author}{Douze, M.}, \bibinfo{author}{Harchaoui, Z.} \& \bibinfo{author}{Schmid, C.}
\newblock \bibinfo{title}{Category-specific video summarization}.
\newblock In \emph{\bibinfo{booktitle}{Computer Vision--ECCV 2014: 13th European Conference, Zurich, Switzerland, September 6-12, 2014, Proceedings, Part VI 13}}, \bibinfo{pages}{540--555} (\bibinfo{organization}{Springer}, \bibinfo{year}{2014}).

\bibitem{pang2024unleashing}
\bibinfo{author}{Pang, Z.}, \bibinfo{author}{Nakashima, Y.}, \bibinfo{author}{Otani, M.} \& \bibinfo{author}{Nagahara, H.}
\newblock \bibinfo{journal}{\bibinfo{title}{Unleashing the power of contrastive learning for zero-shot video summarization}}.
\newblock {\emph{\JournalTitle{Journal of Imaging}}} \textbf{\bibinfo{volume}{10}}, \bibinfo{pages}{229} (\bibinfo{year}{2024}).

\bibitem{li2025spatial}
\bibinfo{author}{Li, Q.}, \bibinfo{author}{Zhan, Z.}, \bibinfo{author}{Li, Y.} \& \bibinfo{author}{Bhanu, B.}
\newblock \bibinfo{journal}{\bibinfo{title}{Spatial--temporal multi-scale interaction for few-shot video summarization}}.
\newblock {\emph{\JournalTitle{Engineering Applications of Artificial Intelligence}}} \textbf{\bibinfo{volume}{142}}, \bibinfo{pages}{109883} (\bibinfo{year}{2025}).

\bibitem{lee2012discovering}
\bibinfo{author}{Lee, Y.~J.}, \bibinfo{author}{Ghosh, J.} \& \bibinfo{author}{Grauman, K.}
\newblock \bibinfo{title}{Discovering important people and objects for egocentric video summarization}.
\newblock In \emph{\bibinfo{booktitle}{2012 IEEE conference on computer vision and pattern recognition}}, \bibinfo{pages}{1346--1353} (\bibinfo{organization}{IEEE}, \bibinfo{year}{2012}).

\bibitem{sharghi2017query}
\bibinfo{author}{Sharghi, A.}, \bibinfo{author}{Laurel, J.~S.} \& \bibinfo{author}{Gong, B.}
\newblock \bibinfo{title}{Query-focused video summarization: Dataset, evaluation, and a memory network based approach}.
\newblock In \emph{\bibinfo{booktitle}{Proceedings of the IEEE conference on computer vision and pattern recognition}}, \bibinfo{pages}{4788--4797} (\bibinfo{year}{2017}).

\bibitem{kanehira2018aware}
\bibinfo{author}{Kanehira, A.}, \bibinfo{author}{Van~Gool, L.}, \bibinfo{author}{Ushiku, Y.} \& \bibinfo{author}{Harada, T.}
\newblock \bibinfo{title}{Aware video summarization}.
\newblock In \emph{\bibinfo{booktitle}{Proceedings of the IEEE Conference on Computer Vision and Pattern Recognition}}, \bibinfo{pages}{7435--7444} (\bibinfo{year}{2018}).

\bibitem{wei2018video}
\bibinfo{author}{Wei, H.} \emph{et~al.}
\newblock \bibinfo{title}{Video summarization via semantic attended networks}.
\newblock In \emph{\bibinfo{booktitle}{Proceedings of the AAAI conference on artificial intelligence}}, vol.~\bibinfo{volume}{32} (\bibinfo{year}{2018}).

\bibitem{liu2022few}
\bibinfo{author}{Liu, D.}, \bibinfo{author}{Zhou, P.}, \bibinfo{author}{Xu, Z.}, \bibinfo{author}{Wang, H.} \& \bibinfo{author}{Li, R.}
\newblock \bibinfo{journal}{\bibinfo{title}{Few-shot temporal sentence grounding via memory-guided semantic learning}}.
\newblock {\emph{\JournalTitle{IEEE Transactions on Circuits and Systems for Video Technology}}} \textbf{\bibinfo{volume}{33}}, \bibinfo{pages}{2491--2505} (\bibinfo{year}{2022}).

\bibitem{liu2023hypotheses}
\bibinfo{author}{Liu, D.} \emph{et~al.}
\newblock \bibinfo{title}{Hypotheses tree building for one-shot temporal sentence localization}.
\newblock In \emph{\bibinfo{booktitle}{Proceedings of the AAAI Conference on Artificial Intelligence}}, vol.~\bibinfo{volume}{37}, \bibinfo{pages}{1640--1648} (\bibinfo{year}{2023}).

\bibitem{liu2023conditional}
\bibinfo{author}{Liu, D.} \emph{et~al.}
\newblock \bibinfo{journal}{\bibinfo{title}{Conditional video diffusion network for fine-grained temporal sentence grounding}}.
\newblock {\emph{\JournalTitle{IEEE Transactions on Multimedia}}} \textbf{\bibinfo{volume}{26}}, \bibinfo{pages}{5461--5476} (\bibinfo{year}{2023}).

\bibitem{castellano2024pyscenedetect}
\bibinfo{author}{Castellano, B.}
\newblock \bibinfo{title}{Pyscenedetect: Video cut detection and analysis tool} (\bibinfo{year}{2024}).

\bibitem{caron2021emerging}
\bibinfo{author}{Caron, M.} \emph{et~al.}
\newblock \bibinfo{title}{Emerging properties in self-supervised vision transformers}.
\newblock In \emph{\bibinfo{booktitle}{Proceedings of the IEEE/CVF international conference on computer vision}}, \bibinfo{pages}{9650--9660} (\bibinfo{year}{2021}).

\bibitem{radford2021learning}
\bibinfo{author}{Radford, A.} \emph{et~al.}
\newblock \bibinfo{title}{Learning transferable visual models from natural language supervision}.
\newblock In \emph{\bibinfo{booktitle}{International conference on machine learning}}, \bibinfo{pages}{8748--8763} (\bibinfo{organization}{PmLR}, \bibinfo{year}{2021}).

\bibitem{lloyd1982least}
\bibinfo{author}{Lloyd, S.}
\newblock \bibinfo{journal}{\bibinfo{title}{Least squares quantization in pcm}}.
\newblock {\emph{\JournalTitle{IEEE transactions on information theory}}} \textbf{\bibinfo{volume}{28}}, \bibinfo{pages}{129--137} (\bibinfo{year}{1982}).

\bibitem{pedregosa2011scikit}
\bibinfo{author}{Pedregosa, F.} \emph{et~al.}
\newblock \bibinfo{journal}{\bibinfo{title}{Scikit-learn: Machine learning in python}}.
\newblock {\emph{\JournalTitle{the Journal of machine Learning research}}} \textbf{\bibinfo{volume}{12}}, \bibinfo{pages}{2825--2830} (\bibinfo{year}{2011}).

\bibitem{bishop2006pattern}
\bibinfo{author}{Bishop, C.~M.} \& \bibinfo{author}{Nasrabadi, N.~M.}
\newblock \emph{\bibinfo{title}{Pattern recognition and machine learning}}, vol.~\bibinfo{volume}{4} (\bibinfo{publisher}{Springer}, \bibinfo{year}{2006}).

\bibitem{thorndike1953belongs}
\bibinfo{author}{Thorndike, R.~L.}
\newblock \bibinfo{journal}{\bibinfo{title}{Who belongs in the family?}}
\newblock {\emph{\JournalTitle{Psychometrika}}} \textbf{\bibinfo{volume}{18}}, \bibinfo{pages}{267--276} (\bibinfo{year}{1953}).

\bibitem{gygli2014creating}
\bibinfo{author}{Gygli, M.}, \bibinfo{author}{Grabner, H.}, \bibinfo{author}{Riemenschneider, H.} \& \bibinfo{author}{Van~Gool, L.}
\newblock \bibinfo{title}{Creating summaries from user videos}.
\newblock In \emph{\bibinfo{booktitle}{Computer Vision--ECCV 2014: 13th European Conference, Zurich, Switzerland, September 6-12, 2014, Proceedings, Part VII 13}}, \bibinfo{pages}{505--520} (\bibinfo{organization}{Springer}, \bibinfo{year}{2014}).

\bibitem{song2015tvsum}
\bibinfo{author}{Song, Y.}, \bibinfo{author}{Vallmitjana, J.}, \bibinfo{author}{Stent, A.} \& \bibinfo{author}{Jaimes, A.}
\newblock \bibinfo{title}{Tvsum: Summarizing web videos using titles}.
\newblock In \emph{\bibinfo{booktitle}{Proceedings of the IEEE conference on computer vision and pattern recognition}}, \bibinfo{pages}{5179--5187} (\bibinfo{year}{2015}).

\bibitem{gygli2015video}
\bibinfo{author}{Gygli, M.}, \bibinfo{author}{Grabner, H.} \& \bibinfo{author}{Van~Gool, L.}
\newblock \bibinfo{title}{Video summarization by learning submodular mixtures of objectives}.
\newblock In \emph{\bibinfo{booktitle}{Proceedings of the IEEE conference on computer vision and pattern recognition}}, \bibinfo{pages}{3090--3098} (\bibinfo{year}{2015}).

\bibitem{he2016deep}
\bibinfo{author}{He, K.}, \bibinfo{author}{Zhang, X.}, \bibinfo{author}{Ren, S.} \& \bibinfo{author}{Sun, J.}
\newblock \bibinfo{title}{Deep residual learning for image recognition}.
\newblock In \emph{\bibinfo{booktitle}{Proceedings of the IEEE conference on computer vision and pattern recognition}}, \bibinfo{pages}{770--778} (\bibinfo{year}{2016}).

\bibitem{xiao2020query}
\bibinfo{author}{Xiao, S.}, \bibinfo{author}{Zhao, Z.}, \bibinfo{author}{Zhang, Z.}, \bibinfo{author}{Guan, Z.} \& \bibinfo{author}{Cai, D.}
\newblock \bibinfo{journal}{\bibinfo{title}{Query-biased self-attentive network for query-focused video summarization}}.
\newblock {\emph{\JournalTitle{IEEE Transactions on Image Processing}}} \textbf{\bibinfo{volume}{29}}, \bibinfo{pages}{5889--5899} (\bibinfo{year}{2020}).

\bibitem{wu2022intentvizor}
\bibinfo{author}{Wu, G.}, \bibinfo{author}{Lin, J.} \& \bibinfo{author}{Silva, C.~T.}
\newblock \bibinfo{title}{Intentvizor: Towards generic query guided interactive video summarization}.
\newblock In \emph{\bibinfo{booktitle}{Proceedings of the IEEE/CVF conference on computer vision and pattern recognition}}, \bibinfo{pages}{10503--10512} (\bibinfo{year}{2022}).

\bibitem{abu2016youtube}
\bibinfo{author}{Abu-El-Haija, S.} \emph{et~al.}
\newblock \bibinfo{journal}{\bibinfo{title}{Youtube-8m: A large-scale video classification benchmark}}.
\newblock {\emph{\JournalTitle{arXiv preprint arXiv:1609.08675}}}  (\bibinfo{year}{2016}).

\bibitem{openai2024gpt4o}
\bibinfo{author}{OpenAI}.
\newblock \bibinfo{title}{Chatgpt (gpt-4o model)}.
\newblock \bibinfo{howpublished}{\url{https://chat.openai.com/}} (\bibinfo{year}{2024}).
\newblock \bibinfo{note}{Version: March 2024. Accessed: March 15, 2025. [Large language model]}.

\bibitem{gemini15pro}
\bibinfo{author}{{Google DeepMind}}.
\newblock \bibinfo{title}{{Gemini 1.5 Pro}}.
\newblock \bibinfo{howpublished}{\url{https://ai.google.dev/gemini-api/docs/models}} (\bibinfo{year}{2025}).
\newblock \bibinfo{note}{Accessed: 2025-08-12}.

\bibitem{claudeSonnet4}
\bibinfo{author}{{Anthropic}}.
\newblock \bibinfo{title}{{Claude Sonnet 4}}.
\newblock \bibinfo{howpublished}{\url{https://www.anthropic.com/claude}} (\bibinfo{year}{2025}).
\newblock \bibinfo{note}{Accessed: 2025-08-12}.

\bibitem{bai2023qwen}
\bibinfo{author}{Bai, J.} \emph{et~al.}
\newblock \bibinfo{journal}{\bibinfo{title}{Qwen technical report}}.
\newblock {\emph{\JournalTitle{arXiv preprint arXiv:2309.16609}}}  (\bibinfo{year}{2023}).

\bibitem{liu2024llavanext}
\bibinfo{author}{Liu, H.} \emph{et~al.}
\newblock \bibinfo{title}{Llava-next: Improved reasoning, ocr, and world knowledge} (\bibinfo{year}{2024}).

\end{thebibliography}
\end{document}